\documentclass[runningheads]{llncs}

\usepackage{eccv}

\usepackage{eccvabbrv}

\usepackage{graphicx}
\usepackage{amsmath}
\usepackage{amssymb}
\usepackage{booktabs}
\usepackage{multirow}
\usepackage{color}
\usepackage{microtype}
\usepackage{makecell}
\usepackage{tikz}
\usepackage{pgfplots}
\usepackage{bbm}
\usepackage{enumitem}
\usepackage{diagbox}
\usepackage{adjustbox}
\usepackage{svg}

\DeclareMathOperator*{\argmin}{arg\,min}
\renewcommand{\vec}[1]{\boldsymbol{#1}}
\newcolumntype{P}[1]{>{\centering\arraybackslash}p{#1}}

\usepackage[accsupp]{axessibility}  % Improves PDF readability for those with disabilities.

\usepackage{hyperref}

\usepackage{orcidlink}

\begin{document}

\title{Robust Incremental Structure-from-Motion with Hybrid Features}

\titlerunning{Robust Incremental Structure-from-Motion with Hybrid Features}

\author{Shaohui Liu\inst{1}$\thanks{Equal contribution}$ \and
Yidan Gao\inst{1*} \and
Tianyi Zhang\inst{1*} \and R\'emi Pautrat\inst{2} \and \\ Johannes L. Sch\"onberger\inst{2} \and Viktor Larsson\inst{3} \and Marc Pollefeys\inst{1,2}}

\authorrunning{S.~Liu et al.}

% % TODO FINAL: Replace with your institution list.
\institute{\ \inst{1}ETH Zurich \quad \quad \inst{2}Microsoft Mixed Reality \& AI Lab Zurich \quad \quad \inst{3}Lund University}

\maketitle

\begin{abstract}
  Structure-from-Motion (SfM) has become a ubiquitous tool for camera calibration and scene reconstruction with many downstream applications in computer vision and beyond. 
While the state-of-the-art SfM pipelines have reached a high level of maturity in well-textured and well-configured scenes over the last decades, they still fall short of robustly solving the SfM problem in challenging scenarios.
In particular, weakly textured scenes and poorly constrained configurations oftentimes cause catastrophic failures or large errors for the primarily keypoint-based pipelines.
In these scenarios, line segments are often abundant and can offer complementary geometric constraints.
Their large spatial extent and typically structured configurations lead to stronger geometric constraints as compared to traditional keypoint-based methods.
In this work, we introduce an incremental SfM system that, in addition to points, leverages lines and their structured geometric relations.
Our technical contributions span the entire pipeline (mapping, triangulation, registration) and we integrate these into a comprehensive end-to-end SfM system that we share as an open-source software with the community.
We also present the first analytical method to propagate uncertainties for 3D optimized lines via sensitivity analysis. 
Experiments show that our system is consistently more robust and accurate compared to the widely used point-based state of the art in SfM -- achieving richer maps and more precise camera registrations, especially under challenging conditions. In addition, our uncertainty-aware localization module alone is able to consistently improve over the state of the art under both point-alone and hybrid setups. 

\end{abstract}

\section{Introduction}

\label{sec::intro}

\begin{figure*}[tb]
\scriptsize
\setlength{\tabcolsep}{1pt}

\begin{tabular}{cccccccccc}
\centering
% First set of images
\begin{minipage}{.485\linewidth}
\centering
\begin{tabular}{cccc}
\centering
    \includegraphics[width=0.24\linewidth, height=15pt, trim={0 50 0 70}, clip]{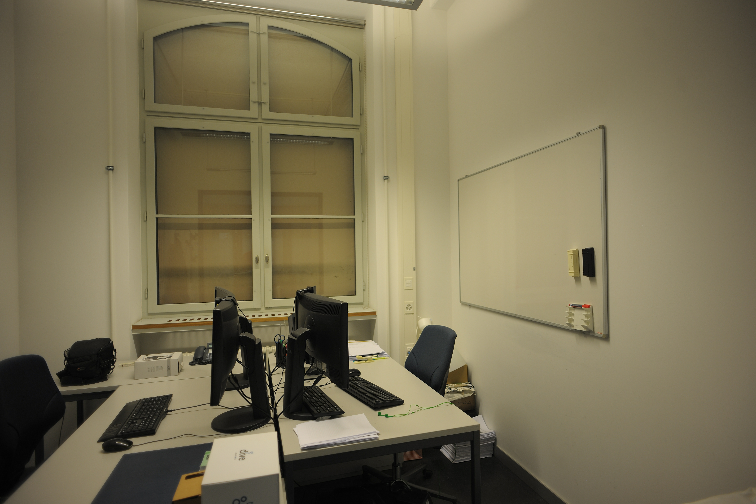} &
    \includegraphics[width=0.24\linewidth, height=15pt, trim={0 60 0 60}, clip]{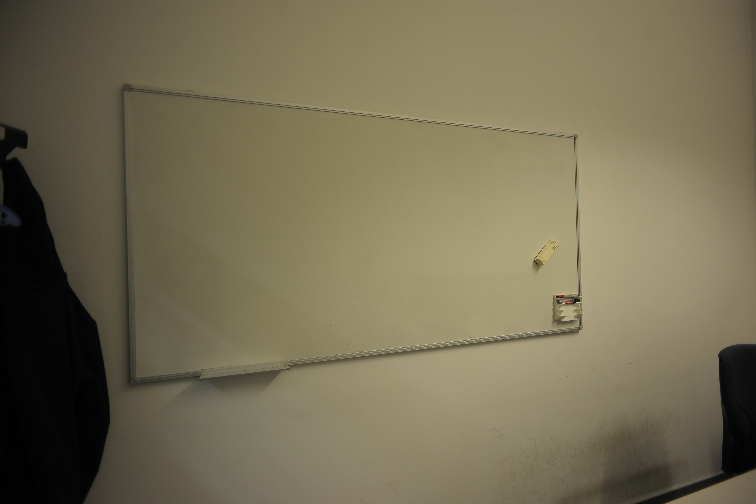} &
    \includegraphics[width=0.24\linewidth, height=15pt, trim={0 10 0 110}, clip]{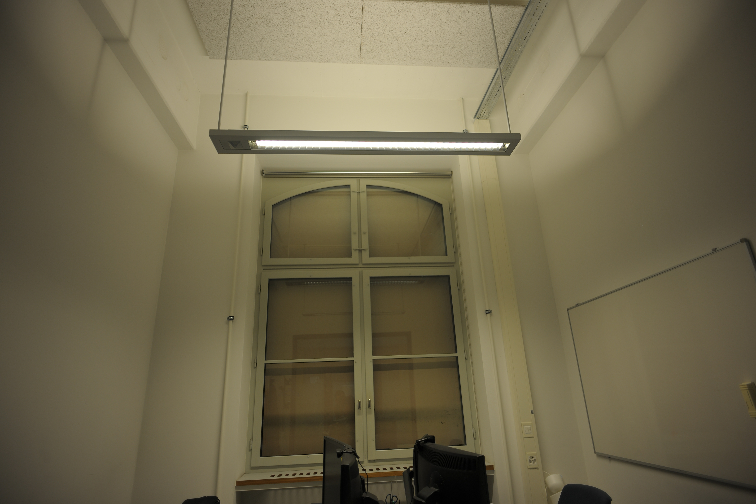} &
    \includegraphics[width=0.24\linewidth, height=15pt, trim={0 70 0 50}, clip]{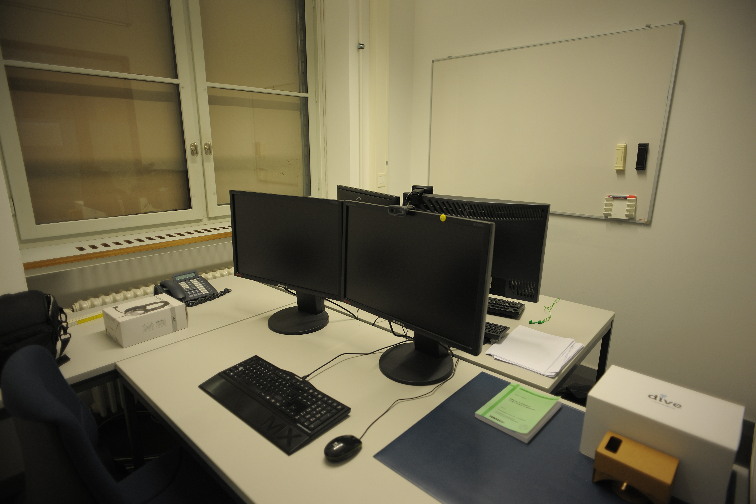}
\end{tabular}
\begin{tabular}{cccccc}
\centering
    \includegraphics[trim={0 50 0 250}, clip, width=0.47\linewidth, height=70pt]{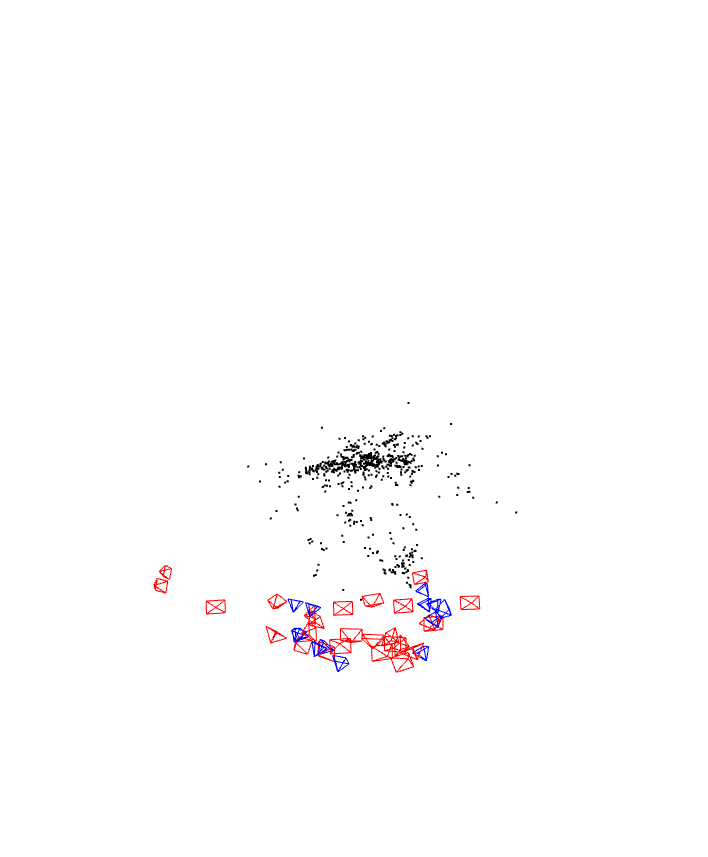} & & & & &
    \includegraphics[trim={0 50 0 250}, clip, width=0.47\linewidth, height=70pt]{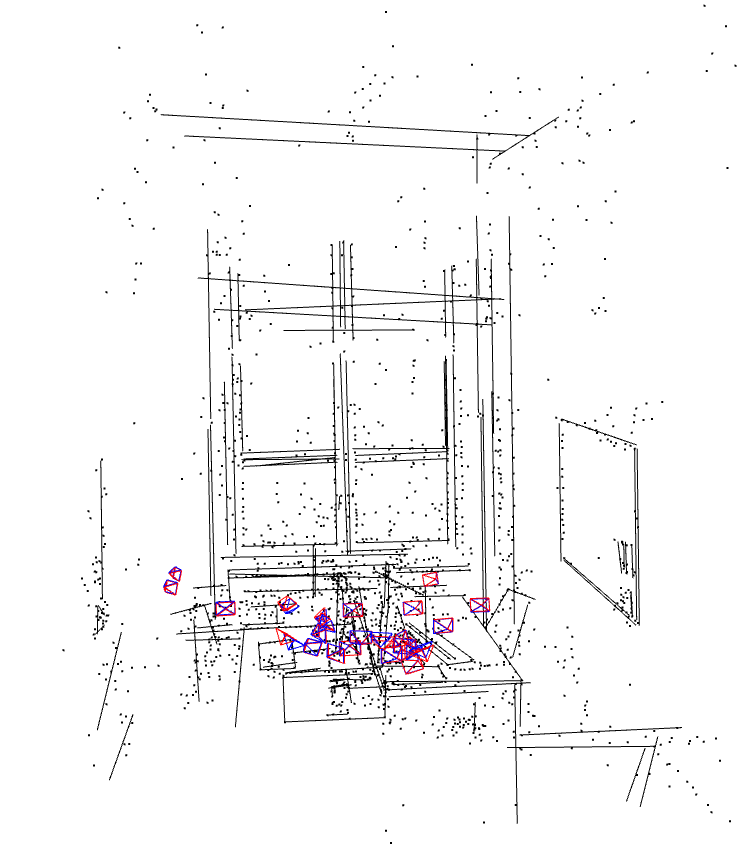}
    \\
    Point (COLMAP) \cite{schonberger2016structure} & & & & & Hybrid (Ours)
    % \\
    % 0 / 26 valid reg. & & & & & 20 / 26 valid reg.
\end{tabular}

\end{minipage}

& & & & & & & & &

% Second set of images
\begin{minipage}{.485\linewidth}
\centering
\begin{tabular}{cccc}
\centering
    \includegraphics[width=0.24\linewidth, height=15pt, trim={0 60 0 60}, clip]{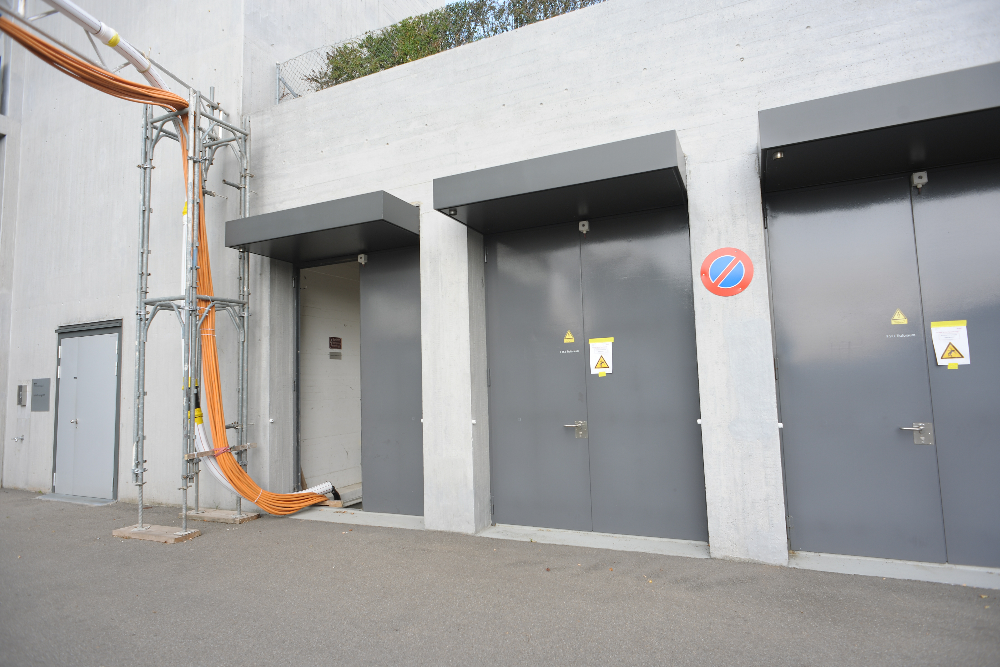} &
    \includegraphics[width=0.24\linewidth, height=15pt, trim={0 60 0 60}, clip]{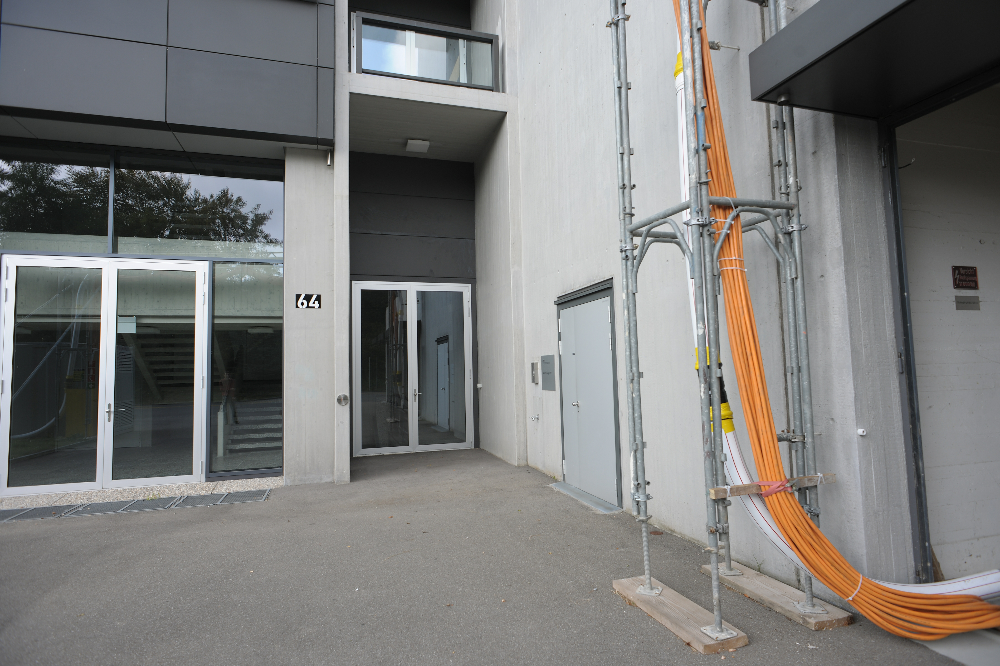} &
    \includegraphics[width=0.24\linewidth, height=15pt, trim={0 60 0 60}, clip]{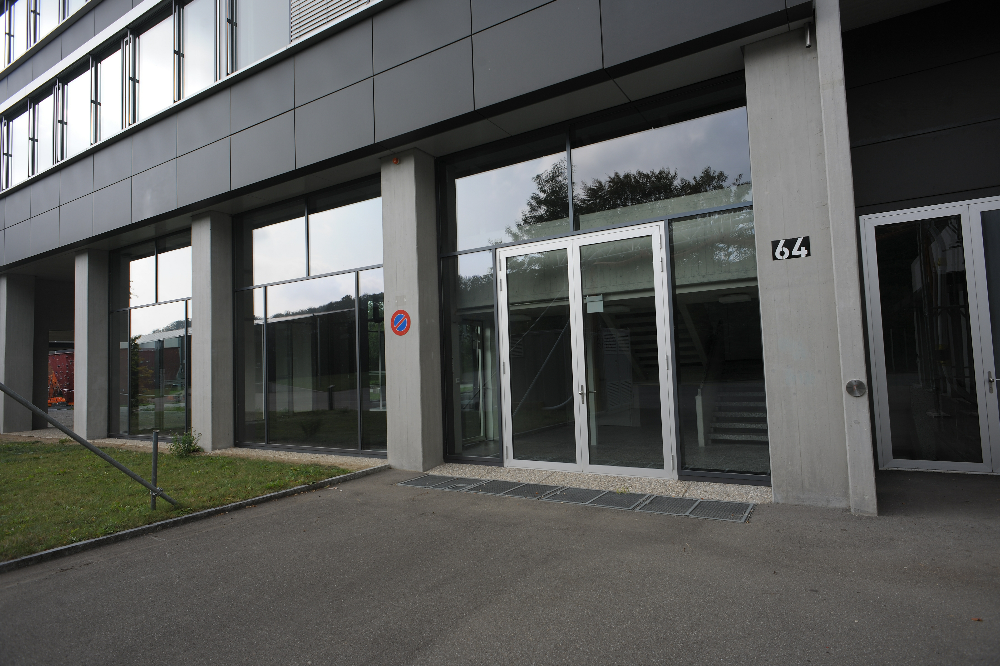} &
    \includegraphics[width=0.24\linewidth, height=15pt, trim={0 60 0 60}, clip]{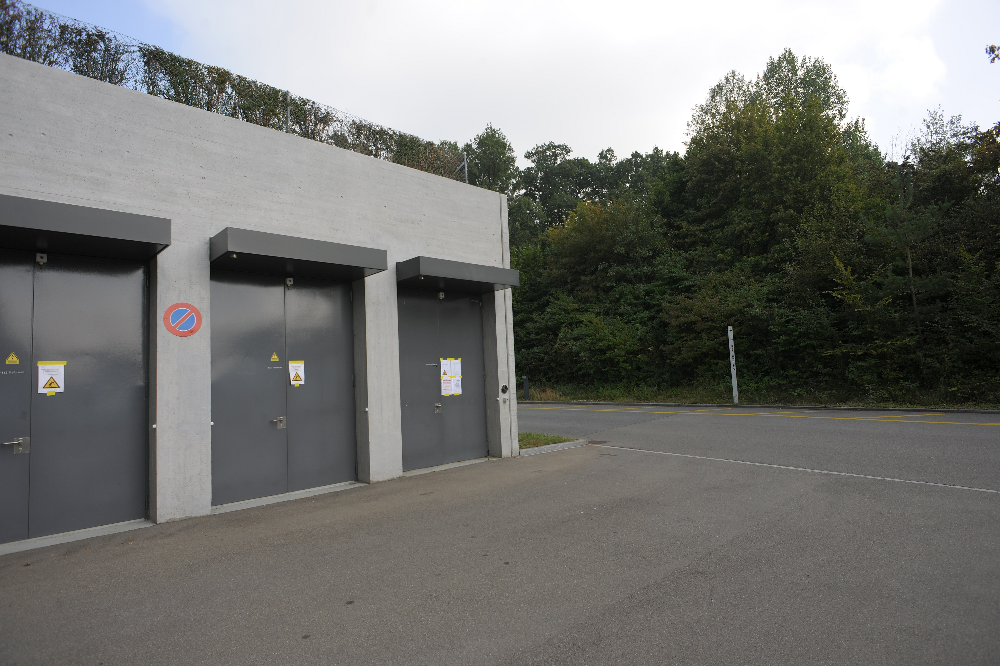}
\end{tabular}
\begin{tabular}{cccccc}
\centering
    \includegraphics[width=0.47\linewidth, height=70pt]{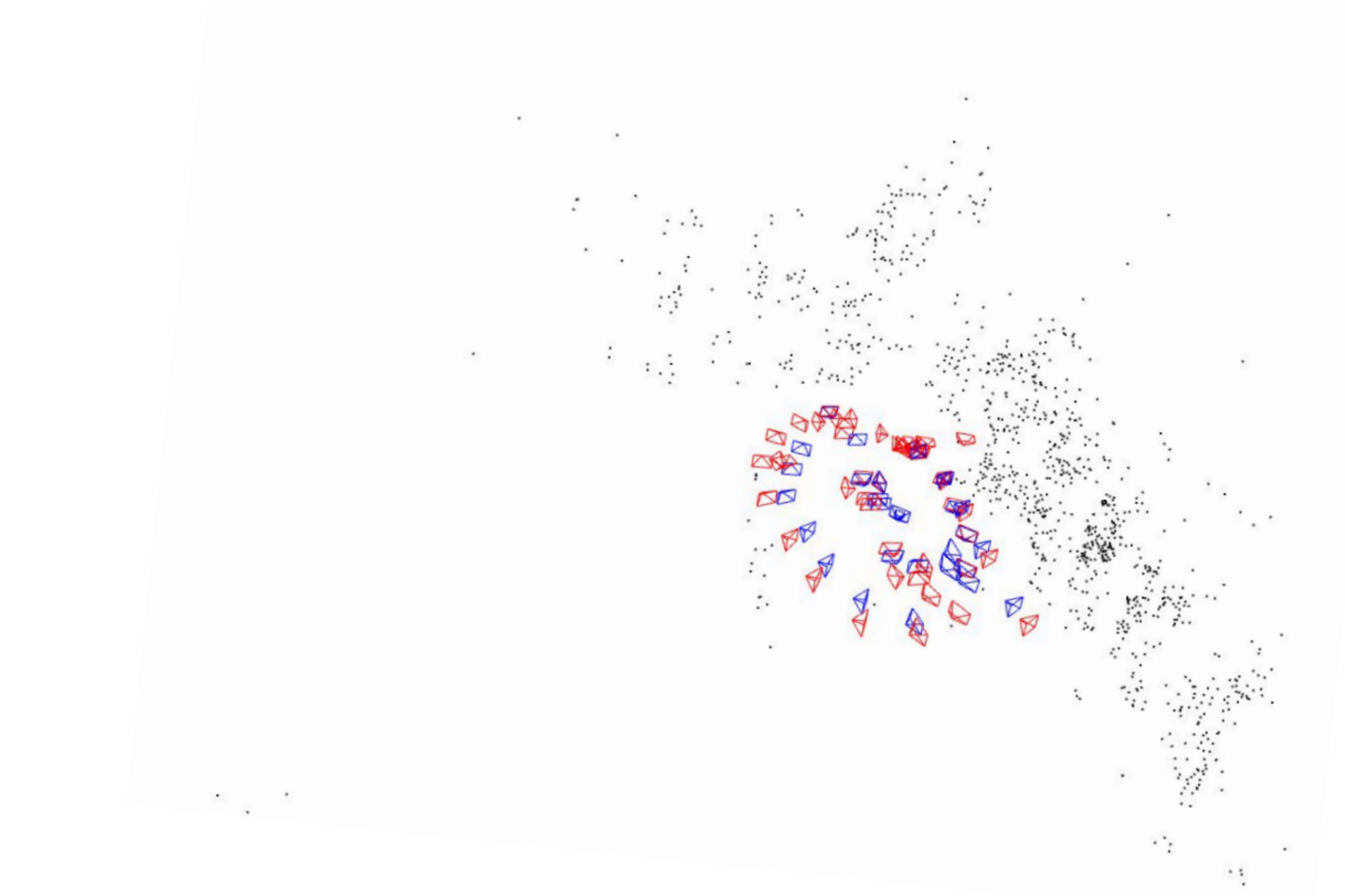} & & & & &
    \includegraphics[width=0.47\linewidth, height=70pt]{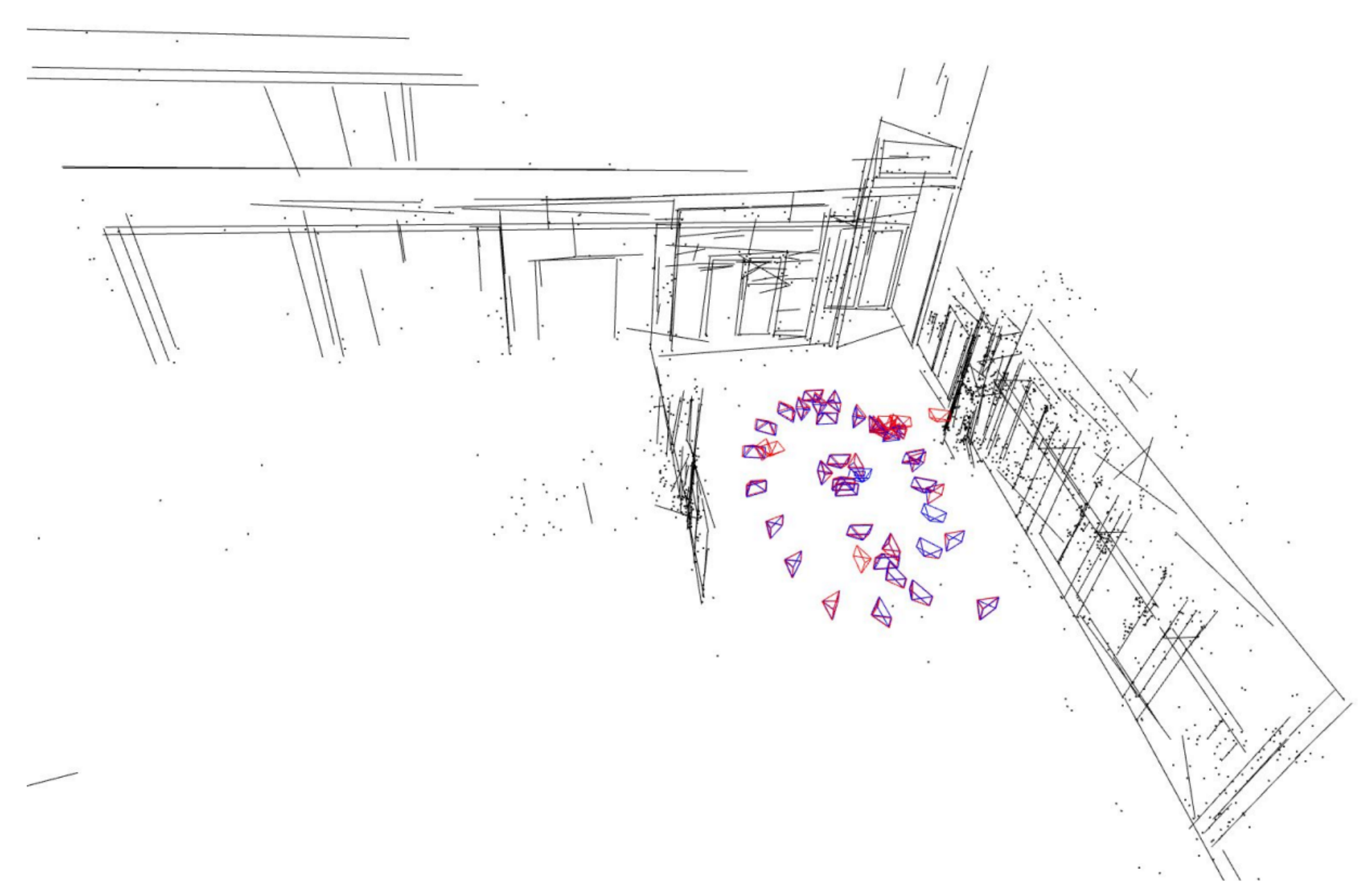}
    \\
    Point (COLMAP) \cite{schonberger2016structure} & & & & & Hybrid (Ours)
    % \\
    % 6 / 45 valid reg. & & & & & 39 / 45 valid reg.
\end{tabular}

\end{minipage}

\end{tabular}
\caption{\textbf{Incremental Structure-from-Motion with points, lines, and vanishing points.} {\color{red} \textbf{Red:}} groundtruths. {\color{blue} \textbf{Blue:}} predictions. We show example indoor scenes where classical point-based Structure-from-Motion fails.
Leveraging additional constraints from line segments, our pipeline can faithfully reconstruct the scene and cameras. }
\label{fig::teaser}
\end{figure*}

Estimating camera parameters and scene geometry from images, also known as Structure-from-Motion (SfM), has enabled a wide variety of applications such as augmented reality \cite{schops2014semi,sarlin2022lamar}, novel view synthesis \cite{nerf,kerbl20233d}, scene reconstruction \cite{yu2022monosdf,li2023neuralangelo}, \etc
For SfM, the incremental paradigm that alternates between updating the map and resectioning cameras is by far the most popular.
This is due to its comparably better accuracy and robustness, as well as having an active open-source community with multiple well-engineered pipelines \cite{snavely2006photo,wu2011visualsfm,schonberger2016structure}, of which COLMAP~\cite{schonberger2016structure} has become the de-facto standard SfM in the recent years.

Most modern SfM systems heavily rely on the presence of stable feature points in the scene, which are detected, matched, and triangulated into sparse 3D point cloud maps followed by repeated bundle adjustments and camera pose estimations.
This, however, regularly prevents these systems from providing robust and accurate results in poorly conditioned scenarios, where the scene has little texture and thus few feature points (\eg, indoor scenes).
% or the camera geometry is weakly constrained (\eg, forward motion).
Compared to points, lines frequently appear in human-made environments where feature points are sparse, they have a large spatial extent, and they often appear in structured configurations offering additional geometric constraints (parallelism, orthogonality, \etc). 
The idea to exploit more structured features, such as straight lines, dates back to the early 2000s~\cite{taylor1995structure,bartoli2004framework,bartoli2005structure,schindler2006line,chandraker2009moving}.
% Also, they generally exhibit higher localization accuracy with less uncertainty in pixels \cite{forstner2016photogrammetric}. Last but not least, line features are especially common in man-made environments such as urban landscapes and indoor scenes, making it complementary to point features with its strong structural cues. 

Despite these clear advantages, line segments are not used in the currently available state-of-the-art SfM pipelines.
This is mainly due to line reconstruction coming with additional challenges compared to the point-based counterparts.
For example, lines are in general more difficult to describe due to often having inconsistent endpoints across views.
In practice, lines also suffer from unstable degenerate configurations during triangulation (as extensively studied in~\cite{Liu_2023_LIMAP}).
Moreover, while there has been great progress in feature point detection and matching in the past decade, line detection and matching has received comparably less attention. 
However, recently, significant progress has been made on line detectors \cite{xue2020holistically,huang2020tp,pautrat2021sold2,Pautrat_2023_DeepLSD} and matchers \cite{abdellali2021l2d2,pautrat2023gluestick} thanks to the advent of deep learning, making it possible to revisit lines as features in SfM.

\iffalse
In the recent work by Liu~\etal~\cite{Liu_2023_LIMAP}, the authors propose the robust and scalable global line mapping framework LIMAP, which tackles the challenges related to the inherent poor conditioning of the line triangulation problem.
The method performs batch processing (global mapping) with known camera poses and is not directly applicable to incremental reconstruction.
The experiments in \cite{Liu_2023_LIMAP} show that joint refinement with both points and lines can improve camera pose accuracy. 
However, since it is only considering mapping with fixed camera poses estimated using classical point-based SfM, it is intrinsically unable to solve the robustness problem
and suffers from the same failure cases.
\fi

In this work, we introduce an end-to-end incremental SfM system that consistently improves the robustness and accuracy over the state of the art by incorporating hybrid features, including points, lines, vanishing points (VPs), and their structured relations.
Our technical contributions span across all the three main steps of incremental SfM: triangulation, refinement, and registration, introducing new robust mechanisms to reliably maintain structural features, and incorporating uncertainty measurements to further improve the robustness. 
Our system is consistently more robust and accurate compared to the widely used SfM pipeline COLMAP \cite{schonberger2016structure}, achieving more precise camera localization, richer sparse maps, more valid registrations, and less catastrophic failure cases (\cf~Fig.~\ref{fig::teaser}). 
By sharing our code as an open-source software, we hope to enable further research on SfM as well as benefit downstream applications in the community. 

Specifically, our technical contributions are listed as follows: 
\begin{itemize}
\item \textbf{System:} We present the first end-to-end incremental SfM system that rigorously integrates points, lines, vanishing points (VPs), and their relations. 

\item \textbf{Incremental Mapping: } We extend incremental triangulation operations initially designed for point features to lines and VPs, leading to an incremental line triangulator with comparable performance to the global triangulator in \cite{Liu_2023_LIMAP}. This removes the need from \cite{Liu_2023_LIMAP} to get all the images posed beforehand.

\item \textbf{Refinement and Bundle Adjustment: }
We propose to explicitly identify reliable/unreliable tracks with uncertainty modeling and apply two-step refinement with cached inactive supports. This prevents prematurely filtering unreliable line tracks 
%at its early stage
in early stages, without sacrificing the pose accuracy. 

\item \textbf{Registration with Hybrid Features: }
We estimate the 6-DoF camera pose with points, lines and vanishing points together in a hybrid RANSAC framework, employing three existing point-line solvers \cite{zhou2018stable} and two extra gravity-based solvers from a VP correspondence.

\item \textbf{Uncertainty Modeling for 3D Maps: }
We perform uncertainty propagation for both 3D points and lines. In particular, this paper introduces the first analytical method to propagate uncertainties for 3D optimized lines.

\item \textbf{Uncertainty Integration for Refinement and Registration: } We successfully integrate uncertainty in both refinement and registration. Our uncertainty-based registration improves upon the state of the art on public localization benchmarks under both point-alone \cite{hloc} and hybrid \cite{Liu_2023_LIMAP} cases.
\end{itemize}

\section{Related Work}
\label{sec::related_work}
\noindent
\textbf{Structure from Motion.}
Incremental methods have traditionally dominated the state of the art in SfM in terms of robustness and accuracy with an active research community and several open-source software packages~\cite{snavely2006photo,wu2011visualsfm,moulon2016openmvg,schonberger2016structure}. Different from global SfM~\cite{crandall2011discrete,sinha2012multi,jiang2013global,wilson2014robust,sweeney2015optimizing,theia-manual,moulon2016openmvg}, incremental methods sequentially register images followed by repeated local and global refinements.
This approach is usually slower but yields more robust and accurate results. The community has made tremendous progress on efficiency and scalability~\cite{steedly2003spectral,crandall2011discrete,agarwal2011building,frahm2010building,kushal2012visibility,wu2013towards,bhowmick2015divide} as well as robustness and accuracy~\cite{dellaert2000structure,chum2003locally,qian2004structure,nister2005preemptive,lebeda2012fixing,schonberger2016structure,dusmanu2021cross,dusmanu2020multiview}. Over the last years, COLMAP~\cite{schonberger2016structure} has emerged as the de facto standard incremental pipeline for SfM, with applications in many downstream tasks in computer vision~\cite{nerf,kerbl20233d} and beyond. Most recently, and orthogonal to our contributions, learning-based pipelines have also been explored~\cite{tang2018ba,xiao2023level,wang2023posediffusion,wang2023dust3r,wang2023visual,zhang2024cameras}, yet still being unable to match the performance of COLMAP on large-scale scenes. Improvements on using pixel-perfect features~\cite{lindenberger2021pixel} and semi-dense matching~\cite{he2023detector} show great potential and could be combined with our work.
In this paper, we introduce a scalable incremental SfM system, that is built upon the success of COLMAP, while improving its robustness and accuracy by carefully incorporating structural features into the entire reconstruction process. 

~\\
\noindent
\textbf{Integration of Lines/Structures in Geometric Pipelines.}
The idea of improving SfM with line features dates back to the early 2000s. Bartoli and Sturm~\cite{bartoli2005structure} pioneered a full SfM system with lines, followed by Schindler~\cite{schindler2006line} who integrated Manhattan assumptions. Chandraker~\etal~\cite{chandraker2009moving} proposed a robust stereo-based system on infinite lines.
Work in the field of SLAM and visual odometry~\cite{marzorati2007integration,gomez2016robust,zuo2017robust,pumarola2017pl,he2018pl,gomez2019pl,wei2019real,mateus2021incremental,lim2021avoiding,lim2022uv,shu2023structure,yan2023plpf} has focused on integrating line features to improve the accuracy, yet only constrained to sequential data and often with strong motion assumption (\eg, from inertial data), while our approach works on general, unstructured input.
In recent years, researchers have also made progress on incorporating lines and vanishing points (VP) in global methods~\cite{micusik2017structure,holynski2020reducing} and exploiting curves in bundle adjustment~\cite{nurutdinova2015towards}.
However, none of the previous works have developed a full end-to-end SfM system that can in practice compete with COLMAP~\cite{schonberger2016structure} in terms of versatility and robustness.
While lines intuitively provide benefits in terms of complementary and rich geometric constraints~\cite{bazin_cvpr_2012,zhang2015,Li_2019_ICCV,Qian2022ARO,Liu_2023_LIMAP,pautrat2023vanishing,forstner2016photogrammetric}, they come with significant practical challenges~\cite{chandraker2009moving,Liu_2023_LIMAP} due to occlusion of endpoints and degenerate configurations as well as inherent difficulties to match them robustly across different views.
The recent breakthrough developments on line detectors~\cite{von2008lsd,xue2020holistically,pautrat2021sold2,Pautrat_2023_DeepLSD} and matchers~\cite{pautrat2021sold2,abdellali2021l2d2,pautrat2023gluestick}, spur a renewed interest in the community to revisit the problem of leveraging lines and their structural configurations~\cite{hofer2015line3d,Liu_2023_LIMAP}, which is further approached with a learning-based solution in~\cite{bui2024representing}.
These prior works have focused on 3D line reconstruction from given camera geometry.
In contrast, we develop a general method that solves the full SfM problem and thus jointly benefits the robustness and accuracy of SfM estimation.

~\\
\noindent
\textbf{Uncertainties in Multi-View Geometry.}
Modeling of uncertainty is a long-standing and important problem in computer vision~\cite{forstner1987fast,aastrom1998statistical,brooks2001value}. Throughout the years, researchers have continuously made progress on modeling the uncertainty of local point features \cite{kanazawa2003we,kanatani2004geometric,steele2005feature,zeisl2009estimation,meidow2009reasoning,detone2018superpoint} and their matches~\cite{zhang2018uncertainty,germain2020s2dnet,muhle2023learning}. There were also attempts on incorporating uncertainty measurements for radar odometry~\cite{burnett2021radar}, 3D benchmark construction~\cite{sarlin2022lamar}, multi-view stereo~\cite{seki2016patch,poggi2016learning,kuhn2020deepc,zhao2021confidence}, \etc.
Despite the breadth and depth of research in this domain, uncertainty modeling remains challenging in practice and has not been embedded in a principled manner in most SfM pipelines.
Our work integrates principled uncertainty modeling into the reconstruction process for improved robustness and accuracy.
This is important for the integration of lines, which, as we show, especially benefit from probabilistic modeling.
Building on top of the Jacobian derivation on the line reprojection error~\cite{bartoli2005structure,zuo2017robust}, we propose the first method to analytically model the uncertainty of 3D optimized lines based on sensitivity analysis~\cite{fiacco1990sensitivity}.
%, and further demonstrate its usability on improving refinement and visual localization (registration).

\section{Methodology}
\begin{figure*}[tb]
\includegraphics[trim={10 5 15 5}, clip, width=0.99\linewidth, height=55pt]{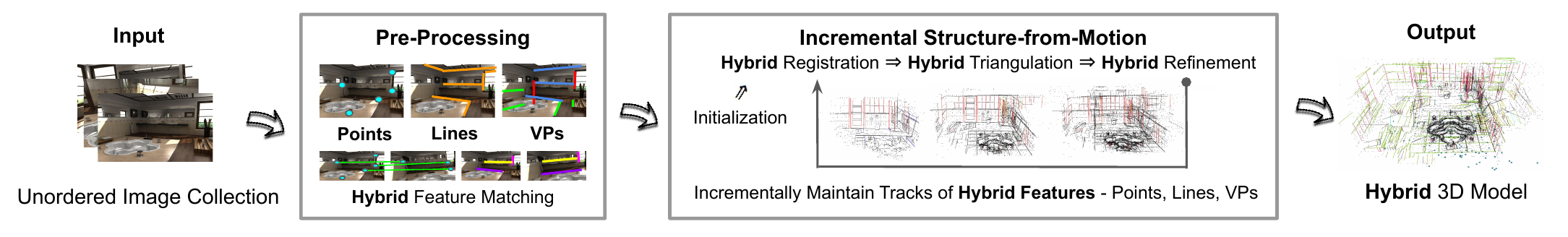}
% lets try to make the height of this figure 80pt
\centering
\caption{\textbf{Our proposed SfM pipeline} exploits hybrid features including points, lines, and vanishing points (VPs). We improve with technical contributions over all three main components: registration, triangulation, and refinement, leading to richer 3D maps (with uncertainty measurements) and more robust camera localization. }
\label{fig::overview}
\end{figure*}

In this section, we present our proposed SfM pipeline. Our method takes an unordered image collection as input, with either calibrated or uncalibrated camera intrinsics. To be able to detect and match straight lines, we require the images to have no radial or tangential distortion.

Fig.~\ref{fig::overview} shows an overview of the proposed pipeline. Our setup and overall system design leverages the same incremental reconstruction approach as the popular point-based SfM software COLMAP~\cite{schonberger2016structure}.
The system first performs local feature detection on the input images and then matches pairs of images with different strategies (exhaustive, sequential, \etc).
Next, the procedure bootstraps the reconstruction process with an initial image pair, followed by progressively adding new images by alternating between camera registration, triangulation of newly observed structures, and iterative local and global refinement using bundle adjustment.
The output of our system is a set of estimated camera parameters and a sparse 3D map with hybrid local features: points, lines, VPs, and their geometric relations. In the following parts, we detail the design of the three main modules: mapping (Sec.~\ref{sec::mapping}), refinement (Sec.~\ref{sec::refinement}), and registration (Sec.~\ref{sec::registration}).

\noindent
\textbf{Detection and Matching on Images.}
Instead of solely relying on point features as in \cite{snavely2006photo,wu2011visualsfm,schonberger2016structure}, we additionally detect and match lines and vanishing points (VPs) across images. The detection and matching of lines can benefit from any existing detectors and matchers. We take the state-of-the-art DeepLSD~\cite{Pautrat_2023_DeepLSD} detector and GlueStick~\cite{pautrat2023gluestick} matcher as our default choices. For VPs, we use JLinkage~\cite{toldo2008robust} as the default detector. The two-view matching of VPs is done through consensus voting from the matches of their associated lines. We consider two VPs a good match if they share at least five line matches. 

\subsection{Incremental Update of Hybrid Maps}
\label{sec::mapping}

\begin{figure}[tb]
\setlength{\tabcolsep}{2pt} % Set the space between columns to zero
\includegraphics[trim={0, 60, 0, 50}, clip, width=0.85\linewidth]{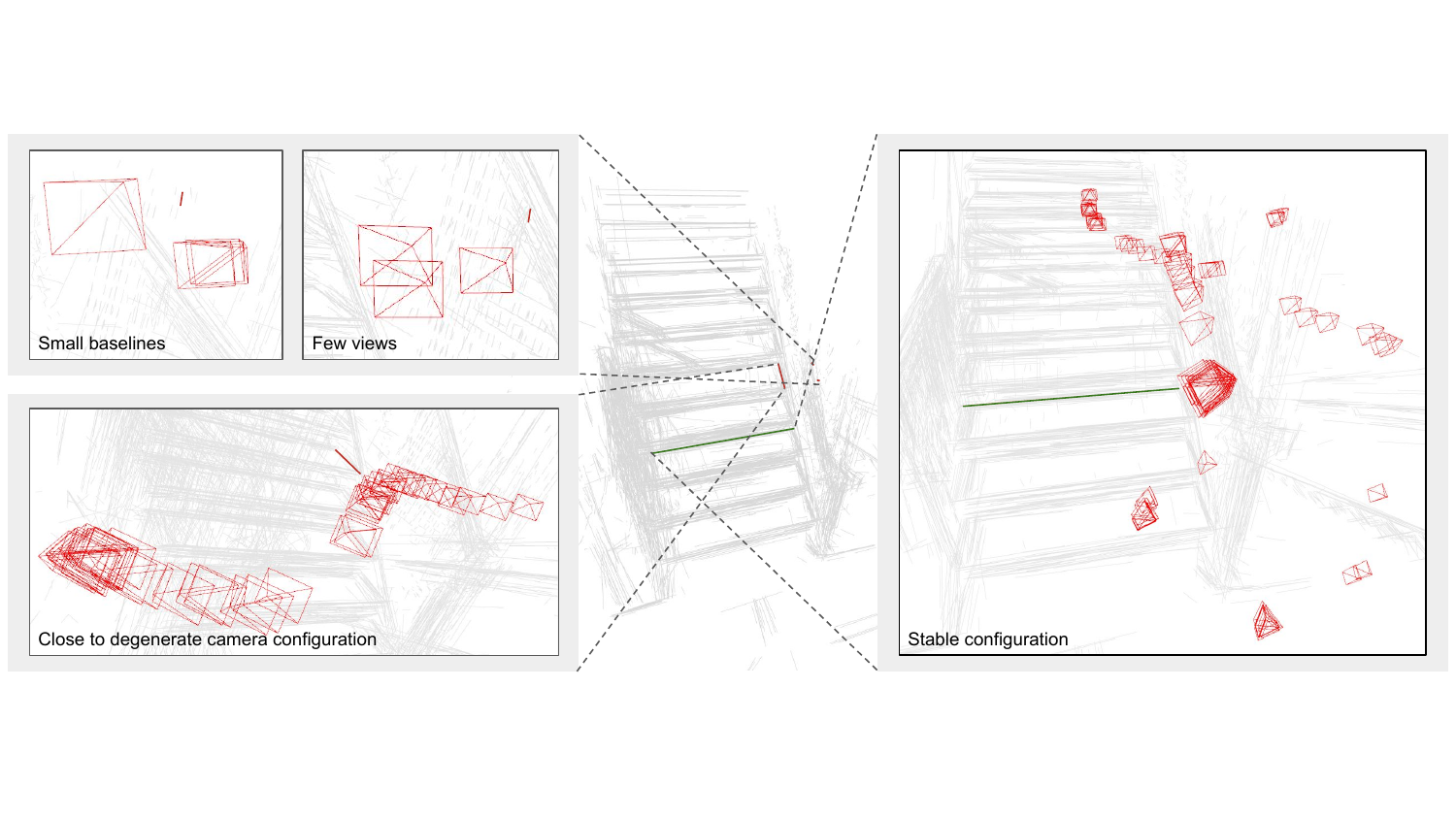}
\centering
\caption{Line triangulation is sensitive to view configurations. \textbf{Left:} unstable tracks from small baselines, few supports, or degenerate patterns. \textbf{Right:} an example stable track. }
\label{fig::motivation}
\end{figure}

While incremental triangulation of points has reached a high level of maturity throughout the years~\cite{hartley1997triangulation,lu2007fast,schonberger2016structure}, the triangulation of lines and VPs has not been studied in great detail.
This is partially due to the natural challenges of line triangulation, including inconsistent endpoints across views and occlusions as well as more frequent unstable and degenerate view configurations (Fig.~\ref{fig::motivation} shows some examples).
While LIMAP~\cite{Liu_2023_LIMAP} recently introduced a global line mapper that can robustly construct line maps from pre-computed posed images, incremental triangulation of lines is drastically harder and requires a sophisticated update mechanism.
Different from points, the verification of a 3D line triangulation requires at least three views, making it much more unstable when only a few views are available.
This happens especially in the early stage of the incremental triangulation process or in scenes with sparse view coverage.

To address these challenges, we combine the techniques from global line triangulation in LIMAP~\cite{Liu_2023_LIMAP} and classic point triangulation \cite{schonberger2016structure} to build a robust and efficient hybrid incremental triangulator for points, lines, and VPs.

\noindent
\textbf{Triangulating Lines from a New Image.} Since line triangulation is inherently unstable and cannot be verified by only two views, we require at least two additional views to triangulate a 3D line. Thus, the incremental triangulation of lines only starts after we register the fourth image. Equivalent to incremental point triangulation, when a new image is registered, we aim to grow the currently triangulated line map with the 2D line features from the new image using the following two operations:
\begin{itemize}\itemsep0pt
    \item \textbf{Continue}: \textit{extends an existing line track.}
    Given a new 2D line detection $l$, we first test if there exists a matched line (in the previously registered images) that is already triangulated in the map and test if the reprojection $\pi(L)$ of the corresponding 3D line $L$ agrees with $l$. Since there can be several such 3D lines, we first compute a score for each of the candidate 3D lines $L$ in a similar fashion as \cite{Liu_2023_LIMAP}. Denoting $d_{perp}$ the 2D perpendicular distance and $d_{ang}$ the 2D angular distance defined in \cite{Liu_2023_LIMAP}, and $\tau_p$ and $\tau_a$ two thresholds (2 pixels and 5 degrees by default), our scoring function is as follows:
    \begin{equation}
        s(L, l) = \min (e^{-(d_{perp}(\pi(L), l) / \tau_p)}, e^{-(d_{ang}(\pi(L), l) / \tau_a)}) .
    \end{equation}
    Similar to \cite{Liu_2023_LIMAP} we additionally check if there is a sufficient overlap. We select the 3D line $L$ with the highest score, and if both errors are below the thresholds, we add the line $l$ into the corresponding line track.
    \item \textbf{Create}:  \textit{triangulates a new line.}
    If we are unable to assign a line segment to any existing track, we try to create a new 3D line with two-view triangulation. We use the same triangulation process as in \cite{Liu_2023_LIMAP}, where a 3D line can be triangulated from either direct algebraic triangulation, or point/VP-guided triangulation. We refer the reader to \cite{Liu_2023_LIMAP} for the details of the triangulation.

    % \item \textbf{Continue}: \textit{extends an existing line track.} We first test if there exists a matched line (in the previously registered images) that is already triangulated in the map and test if the reprojection of the corresponding 3D line agrees with the new line detection. If so, we add the line into the corresponding line track. If there are multiple such 3D lines, we pick the one with the best score (largest agreement) when reprojected in terms of angle and endpoint-to-line distance. We take the lower score between the two measurements, each of which is formulated in the form $s_n = e^{-(r/\tau_r)^2} \in (0, 1]$ following \cite{Liu_2023_LIMAP}, where the scaling factor $\tau_r$ corresponds to the threshold.  
    % \item \textbf{Create}:  \textit{triangulates a new line.}
    % If we are unable to assign a line segment to any existing track, we try to create a new line with two-view triangulation. 
    % This involves a similar triangulation process as in~\cite{Liu_2023_LIMAP}, where the line segment is first triangulated from its matches into several proposals via different strategies (including the naive one and three point-guided methods in \cite{Liu_2023_LIMAP}), and then assigned the best proposal by multi-view scoring. If the best proposal has at least two supports, it is included in the map as a new line track.
    %, with both triangulated lines and its supports included in its track information.
\end{itemize}

We maintain a list of all the tracks that have been updated in the triangulation process and apply non-linear refinement for each updated track.
In the SfM context, this refinement step shares similar functionality with the multi-view point triangulation via singular value decomposition~\cite{schonberger2016structure}. 

To finish the triangulation for the new image, we merge line tracks that are connected in the matching graph.
In addition, we perform a \textit{complete} step as in \cite{schonberger2016structure} to collect potentially missing supports: given a line track, we consider the neighbors of the current 2D supports in the matching graph, and add them to the track if they agree with the reprojection of the 3D line.
% In addition, we perform a \textit{complete} step following \cite{schonberger2016structure}, where we test reprojection agreement on the neighbors of included supports (in the matching graph) to collect potentially missing supports.
This appears to be beneficial to improve the track length under the incremental setup. 

\noindent
\textbf{Recomputation of Endpoints.}
While points are compact in 3D, the spatial extent of the lines changes with their endpoints. Thus, although the infinite line may remain unchanged when the tracks are extended and merged, we need to update its endpoints to be able to correctly merge lines in the future. This is done by unprojecting endpoints onto the infinite line with Pl\"ucker coordinates \cite{bartoli2005structure,Liu_2023_LIMAP}. 

\noindent
\textbf{Retriangulation of Long 2D Lines. }
Long 2D lines are generally more robust than short ones since detection noise is averaged over more pixels. Thus, we force 2D lines that are longer than 100 pixels to \textit{create} new lines through triangulation, even when they are supposed to \textit{continue} on existing ones. This helps build more stable line tracks in the maps.

\noindent
\textbf{Building VP Tracks.} In addition, we also maintain VP tracks to model the parallelism relations among lines. Since a VP in 3D can be mapped from a single view, its maintenance is much easier than lines. Consistency checks only involve measuring the angle of two directions. Please check Sec.~A in supp. for details. 

\subsection{Refinement of Hybrid Structure}
\label{sec::refinement}

After each new image is registered and triangulated, our system performs local and global refinement on both the map and poses. While bundle adjustment of point features is well studied~\cite{triggs2000bundle,agarwal2010bundle}, lines suffer comparably more from outliers and instability after triangulation. On the one hand, during refinement, these incorrect lines can potentially corrupt the entire reconstruction, especially in the early stages of SfM when only a few images are registered. As such, we need to be selective about which set of lines are added to the optimization problem.
%While we aim to prevent the noisy line tracks from affecting the pose optimization, those are often the 3D lines that are at its early stage of creation waiting for more views to support itself.
% While it is important to reduce the impact of the unstable lines on the camera poses, we also want to avoid prematurely filtering line tracks as many of them can become stable once more views are added.
On the other hand, we do not want to prematurely filter line tracks, as many of them can become stable after more views are registered.
In this section, we present several mechanisms for keeping track of the reliability of line tracks and their supports in the refinement process, without having to prematurely remove them. 

\noindent
\textbf{Caching Inactive Supports.} Due to unstable line triangulation from sparse views, a good support can easily be an outlier at an early stage due to pose perturbations in the refinement process.
Deleting those supports will largely slow down the mapping process and also drop a large number of potentially good lines (see supplementary material for visualizations).
Motivated by this fact, we propose to instead attach to each support an \textit{active} label. After each refinement, we check all the supports and set their labels depending on whether it is currently an inlier (active) or an outlier (inactive). We remove the inactive ones only when a track becomes stable (has more than 10 active supports). To avoid the noisy supports to be stuck in the wrong tracks, we include each inactive support at triangulation and refinement. In this way, we keep the option for potential inlier supports to become active later in the optimization. 

\noindent
\textbf{Propagating 3D Uncertainty from 2D Measurements.} 
To be able to keep the noisy tracks without affecting the pose optimization, we need to correctly identify unreliable tracks. While the number of active supports makes a reasonable indicator, lines suffer more from instability due to view configurations (see Fig.~\ref{fig::motivation}). Therefore, we directly model the 3D line uncertainty with covariance propagation. 
% Since we aim to only model the uncertainty coming from the view configuration 
In this paper, we assume the detection of each keypoint and line endpoint follows $\mathcal{N}(\vec{0}, \vec{1})$, while more advanced modeling~\cite{meidow2009reasoning} can be integrated easily as well.

%though this can be further improved with advanced modeling of 2D uncertainty \cite{meidow2009reasoning}.

We start by revisiting the uncertainty modeling of points, where multi-view triangulation can be formulated as a non-linear least-squares problem:
\begin{equation}
    \vec{X}^* 
    = \frac{1}{2}\argmin_{\vec{X}} \sum_{k} \Vert \vec{r_k} \Vert^2, r_k = \Pi_k(\vec{X}) - \vec{x_k},
    \label{eq::opt_point}
\end{equation}
% \begin{equation}
%     \vec{X}^* 
%     = \frac{1}{2}\argmin_{\vec{X}} \sum_{k} \Vert \vec{r_k} \Vert^2
%     = \frac{1}{2}\argmin_{\vec{X}} \sum_{k} \Vert \Pi_k(\vec{X}) - \vec{x_k} \Vert^2
%     \label{eq::opt_point}
% \end{equation}
where the 3D point is optimized across views w.r.t.~its 2D observations $\vec{x_k}$. For such least-squares problems \cite{ceres}, the uncertainty can be propagated from the observation to the optimal 3D point $\vec{X}^*$ using the Jacobian $J$ of the reprojection function. With the assumption of unit covariance, this only involves inverting the approximate Hessian $J^TJ$ (refer to Sec. B.1 in supp. for details). 

For lines, however, the reprojection residual (denoted as $\vec{e_k}$) is generally formulated as the endpoint-to-line distance $d_{perp}(\Pi_k(\vec{L}), \vec{l_k})$, which cannot be written in the least squares form due to the fact that the derivative of the residual over the endpoint observation depends on the optimized 3D line. This makes the 3D uncertainty intractable with the previous formulation. 

% For lines, however, the reprojection residual (denoted as $\vec{e_k}$) is generally formulated as the endpoint-to-line distance, where its derivative over the endpoint observation depends on the optimal 3D line. Thus, the error uncertainty in the optimization not only depends on the observations but also depends on the optimum, making the 3D uncertainty intractable with the previous Jacobian-based propagation. We present more detailed discussions in Sec. B of the supp.

\begin{figure}[tb]
\scriptsize
\setlength\tabcolsep{0pt}
\begin{tabular}{cc}
{\includegraphics[width=0.42\linewidth, height=100pt]{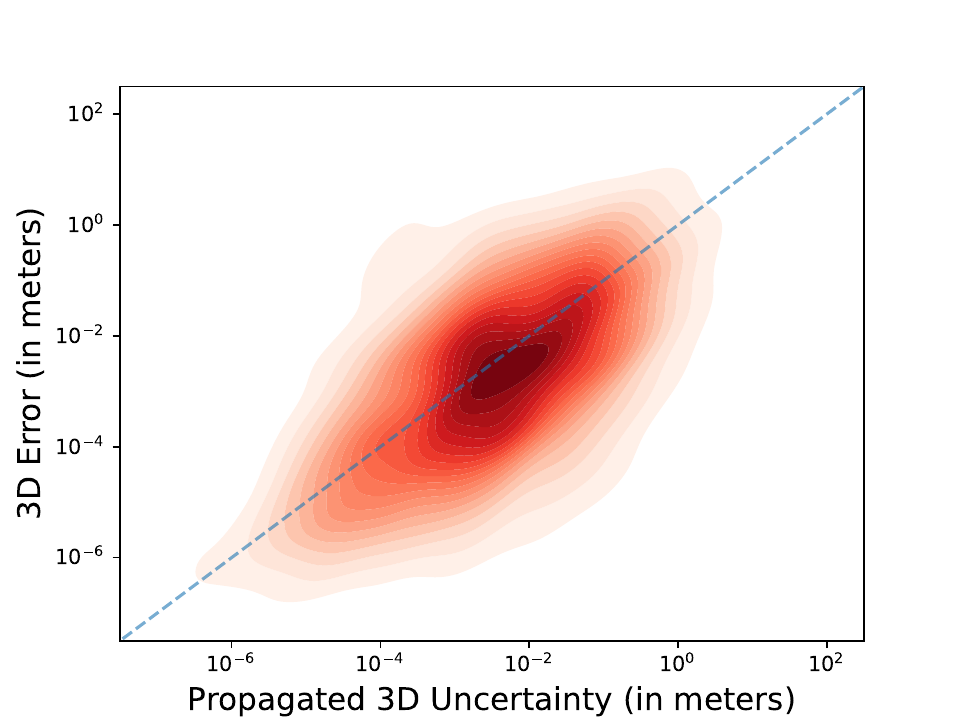}} &
{\includegraphics[width=0.42\linewidth, height=100pt]{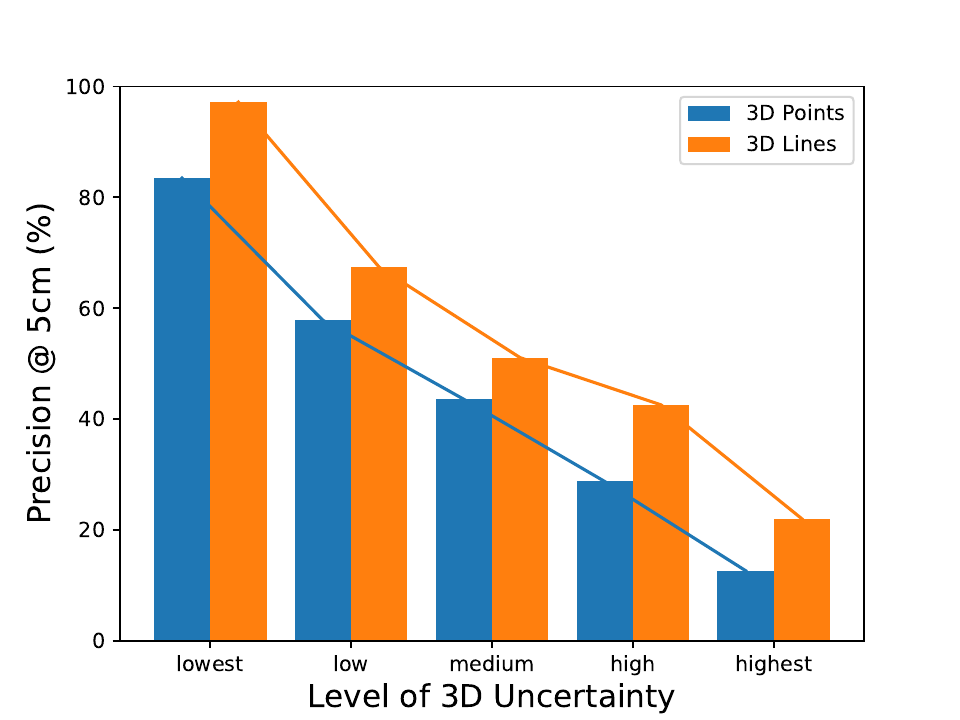}}
% Point tracks & Line tracks 
\end{tabular}
\centering
\caption{\textbf{Relations between the propagated uncertainty for each track and its accuracy on ETH3D \cite{sinha2012multi}.} \textbf{Left:} For each 3D feature, we plot its 3D error and uncertainty both in meters. \textbf{Right:} We report precision over each bin sorted w.r.t.~to the 3D uncertainty. Points and lines with lower uncertainty tend to have higher precision.}
\label{fig::correlation_test}
\end{figure}

We propose to tackle the problem with second-order sensitivity analysis~\cite{fiacco1990sensitivity}, which relies on the fact that the derivatives of the non-linear objective (denoted as $E$) over the optimized variables is always zero at the optimal 3D line $\vec{L^*}$:
\begin{equation}
    \frac{\partial E}{\partial \vec{L}}|_{\vec{L} = \vec{L}^*}
    = \sum_{k} \vec{e_k}^T \frac{\partial \vec{e_k}}{\partial \vec{L}}|_{\vec{L} = \vec{L}^*}
    = \vec{0}
    \label{eq::line_optimal_zero}
\end{equation}
Since the derivative of the left-hand side of Eq.~(\ref{eq::line_optimal_zero}) over the input endpoint observation $\vec{l_k}$ (\ie, $\partial (\partial E / \partial \vec{L}) / \partial \vec{l_k}$) always equals the zero vector at $\vec{L}^*$, we can derive a linear system that solves for the target Jacobian $\partial \vec{L^*} / \partial \vec{l_k}$. This can be used to correctly propagate the uncertainty into optimal 3D line in its Pl\"ucker form (see Sec.~B in supp.~material for detailed derivations). The correctness of our propagated uncertainty is supported by numerical tests with finite differences and correlation tests with map accuracy (Fig.~\ref{fig::correlation_test}).

\noindent
\textbf{Two-step Refinement.}
As previously discussed, we can combine the number of active supports and the level of uncertainty to identify reliable line tracks from unstable ones.
In order to get a scale-invariant metric for the 3D uncertainty, we rescale it into pixels by multiplying it with the median value of ${f}/{d}$ across the supporting images, where $f$ is the focal length and $d$ the depth of the line midpoint. 
% In order to get a scale-invariant metric for the 3D uncertainty, we rescale it into pixels with the median value of depth over focal length across the supporting images. 
Different from conventional practice in bundle adjustment, we propose to perform the refinement in two steps: an initial full hybrid bundle adjustment only including the reliable tracks, followed by a fixed-pose refinement of the unreliable tracks. After each bundle adjustment, we update the \textit{active} label of supports and the reliability for each track. This allows us to decouple the pose optimization from unstable or incorrect line triangulations while keeping the potential for currently unreliable lines to become reliable in the future.

\noindent
\textbf{Integration of Structural Associations.}
Similarly to \cite{Liu_2023_LIMAP}, we also incorporate structural associations between points and lines as well as lines and VPs. While this enables joint optimization with structural constraints, the point-line association residual breaks the block-wise nature of the bundle adjustment problem, leading to slower runtimes. Refer to Sec.~C in the supp.~mat.~for a discussion. 

\subsection{Hybrid Registration}
\label{sec::registration}

With the construction and maintenance of the refined hybrid maps, our system has rich information during camera registration to better pose new images with the existing 3D structure. The additional line and VP correspondences can not only help with more accurate estimation of camera poses but also enable more valid registrations on challenging images with few point correspondences. 

\noindent
\textbf{Integration of Line and VP Correspondences.}
Inspired by~\cite{Liu_2023_LIMAP}, our camera registration uses a hybrid RANSAC framework~\cite{camposeco2018hybrid}. Given a new image to be registered, we collect 2D-3D correspondences of points, lines, and VPs by traversing their matches with the already registered images. We employ six different minimal solvers from the combination of the hybrid correspondences, namely the conventional P3P solver~\cite{persson2018lambda}, the hybrid point-line solvers~\cite{zhou2018stable, PoseLib} (including P2P1LL, P1P2LL, P3LL), and optionally, when a VP correspondence is available, the 2-point and 1-point + 1-line solvers with one VP correspondence. The VP-based solvers are variants of the known-gravity solvers \cite{kukelova2010closed,PoseLib}, with the known direction tilted (check Sec. D in supp.~mat~for details). Following \cite{camposeco2018hybrid}, the sampling probability and termination criteria for each solver depend on the corresponding inlier ratios. Both points and lines are included in the scoring and local optimization with their reprojection errors. After robust estimation, we use both the number of point and line inliers to determine whether the registration is successful. In this way, we relax the requirement of abundant inlier points with the additional line features, which are particularly common in indoor scenes. 

\noindent
\textbf{Integration of Uncertainty Estimation.}
With the 3D uncertainty being propagated to each track, as described in Sec.~\ref{sec::refinement}, we can further model the reliability of different point/line correspondences from the map's perspective. It is worth noting that a 3D point/line with high 3D uncertainty can still be valuable for localization in views where its projection is stable. Thus, during registration, we model the correspondence reliability with the uncertainty of the reprojection error vector rather than the raw 3D global uncertainty. This requires an initial pose, which can be estimated by first running a few iterations of the original uncertainty-free method. After we get the reprojection uncertainty for each correspondence, we can use it as a reweighting factor in both scoring and local optimization in the robust estimation framework, which enables the stable part of the maps to contribute more to the problem. In our experiments, this uncertainty-aware mechanism achieves consistent accuracy improvement in the general localization problem on both point-alone and hybrid cases (Table~\ref{tab:localization}). 

\section{Experiments}
\noindent
\textbf{Implementation Details.}
Our system is implemented in C++ with Python bindings \cite{pybind11}. We use the same hyperparameters for all experiments across datasets. Parameters for points are identical to COLMAP~\cite{schonberger2016structure} for fair comparison. For more details, refer to Sec.~E in the supp.~material.
% The feature extraction and matching were run on a single NVIDIA Geforce RTX 3080.

\begin{table}[tb]
\begin{center}
\scriptsize
\setlength{\tabcolsep}{3pt}
\caption{\textbf{Structure-from-Motion results on Hypersim \cite{roberts:2021} and ETH3D \cite{schops2017multi}}. We report the relative pose AUC and the percentage of valid registration within 5 cm / 5 deg after robust alignment, for both our system (``Hybrid") and COLMAP (``Point") \cite{schonberger2016structure} on SIFT \cite{lowe2004distinctive} + nearest neighbor  and SuperPoint \cite{detone2018superpoint} + SuperGlue \cite{sarlin2020superglue}. }
\begin{tabular}{cccccccccc}
\toprule
\multirow{2}{*}{Dataset} & \multirow{2}{*}{Point Feature} & \multirow{2}{*}{Method} & \multicolumn{4}{c}{\multirow{2}{*}{AUC @ 1°/3°/5°/10° $\uparrow$}} & \multirow{2}{*}{Valid Reg. $\uparrow$} \\
& & & & & \\
\midrule
\multirow{4}{*}{Hypersim} & \multirow{2}{*}{SIFT + NN} & Point & 71.3 & 82.5 & 85.0 & 86.8 & 93.7\% \\
& & Hybrid & \textbf{82.1} & \textbf{86.6} & \textbf{87.6} & \textbf{88.3} & \textbf{93.9\%} \\
\cmidrule(l{5pt}){2-8}
& \multirow{2}{*}{SP + SG} & Point & 80.1 & 89.5 & 91.6 & 93.2 & 96.7\% \\
& & Hybrid & \textbf{87.0} & \textbf{92.1} & \textbf{93.3} & \textbf{94.1} & \textbf{97.0\%} \\
\midrule
\multirow{4}{*}{ETH3D} & \multirow{2}{*}{SIFT + NN} & Point & 16.2 & 26.7 & 28.1 & 32.1 & 46.4\% \\
& & Hybrid & \textbf{24.3} & \textbf{34.8} & \textbf{37.4} & \textbf{40.8} & \textbf{59.4\%} \\
\cmidrule(l{5pt}){2-8}
& \multirow{2}{*}{SP + SG} & Point & 33.0 & 54.7 & 61.1 & 66.4 & 69.8\% \\
& & Hybrid & \textbf{37.3} & \textbf{57.9} & \textbf{63.3} & \textbf{68.8} & \textbf{75.3\%} \\
% \midrule
% \multirow{4}{*}{LaMAR} & \multirow{2}{*}{SIFT + NN} & Point & \ 7.2\% & 0.0 & 0.3 & 0.7 & 1.2 \\
% & & Hybrid & \textbf{\ 8.1\%} & 0.0 & \textbf{0.9} & \textbf{2.7} & \textbf{4.6} \\
% \cmidrule(l{5pt}){2-8}
% & \multirow{2}{*}{SP + SG} & Point & 16.8\% & 0.0 & 2.8 & 7.9 & 14.5\\
% & & Hybrid & \textbf{17.2\%} & 0.0 & \textbf{4.0} & \textbf{10.8} & \textbf{18.5} \\
\bottomrule
\end{tabular}
\label{tab::unknown_intrinsics}
\end{center}
\end{table}

\begin{figure}[tb]
\scriptsize
\setlength\tabcolsep{2pt} % Adjust as needed
\begin{tabular}{cccc}
\includegraphics[width=0.32\linewidth, height=70pt]{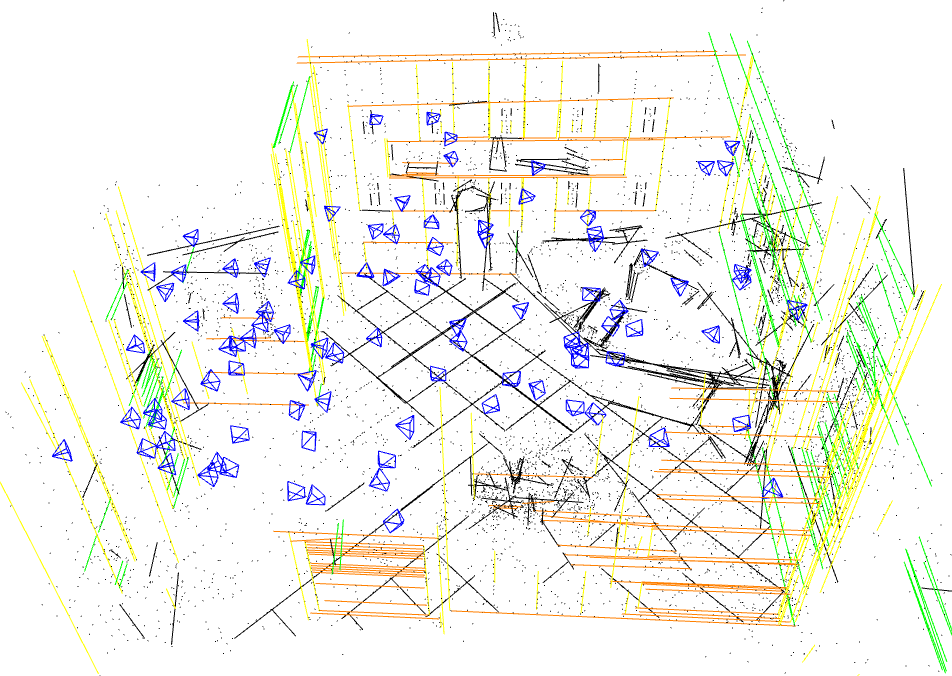} &
\includegraphics[width=0.32\linewidth, height=70pt]{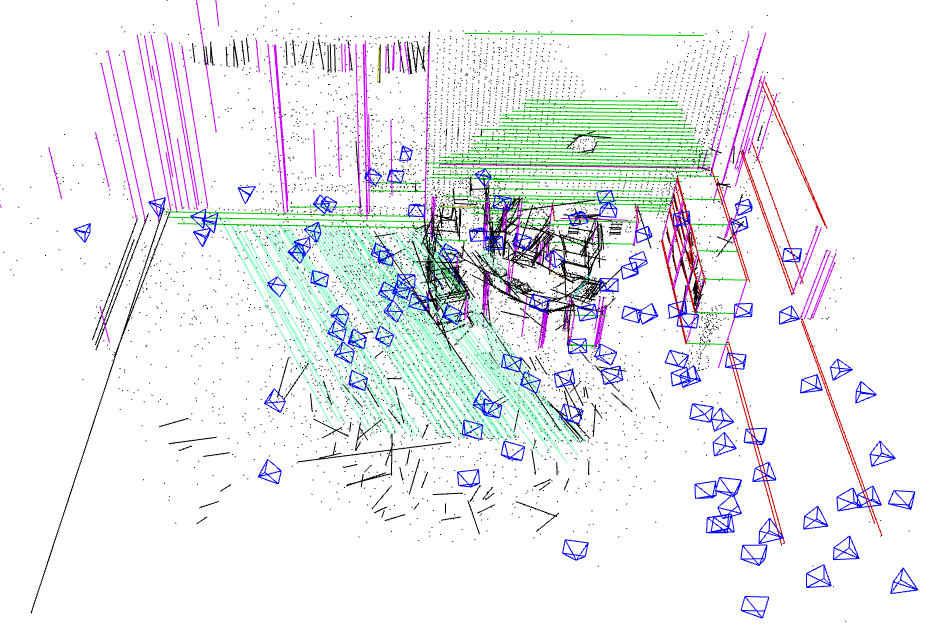} &
\includegraphics[width=0.32\linewidth, height=70pt]{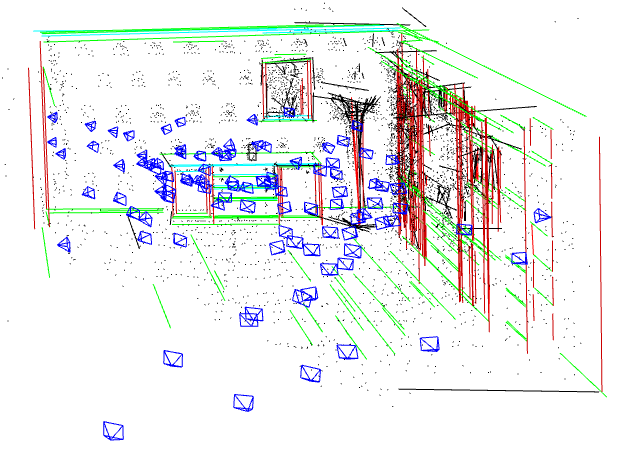}
\end{tabular}
\centering
\caption{Some examples of our hybrid maps on Hypersim \cite{roberts:2021}. Parallel lines from line-VP associations are colored the same.}
\label{fig::exp_hypersim}
\end{figure}

\subsection{Structure-from-Motion}
\noindent
\textbf{Results on Unstructured Data.}
We first evaluate our methods on two public datasets: Hypersim \cite{roberts:2021} and ETH3D \cite{schops2017multi}. For Hypersim, our evaluation runs on the first 8 scenes following~\cite{Liu_2023_LIMAP}. For ETH3D, we use the training set of DSLR images (13 scenes) while resizing it to a maximum image dimension of 756.
% , to be able to run the state-of-the-art learning-based GlueStick line matcher \cite{pautrat2023gluestick}
% For LaMAR, we use the \textit{hetrf} sensor from the HoloLens query validation data (12 sequences) as our visual input. 
Results are shown in Table~\ref{tab::unknown_intrinsics}. Compared to the point-alone baseline COLMAP \cite{schonberger2016structure}, our method largely improves the accuracy on highly structured indoor scenes from Hypersim, and achieves more valid registrations on ETH3D. This holds for different types of point features~\cite{lowe2004distinctive,detone2018superpoint,sarlin2020superglue}. 
% While the LaMAR data remains challenging for visual-based SfM, our method achieves consistent improvements across the metrics. 
% Structural features are particularly helpful in scenes consist of large homogeneous regions, where only few salient keypoints can be detected.
This can be attributed to the strong geometric constraints from structural features.
Fig.~\ref{fig::exp_hypersim} shows that our method is able to incrementally reconstruct richer maps with structural relations. 

To verify that our method does not degrade the performance on scenes without abundant presence of distinctive line features, we test our SfM pipeline on the validation split (3 scenes) from the \textit{Image Matching Benchmark 2020}~\cite{jin2021image}. Table~\ref{tab::phototourism} shows that despite the rich textures in the test set, our method achieves promising improvements on outdoor scenes over COLMAP \cite{schonberger2016structure}, which remains the widely used backend in the benchmark \cite{jin2021image}. Qualitative results in supp. further show that our method achieves reasonable reconstruction from as few as 5 images.

\begin{table}[tb]
    \centering
    \scriptsize
    \caption{\textbf{Structure-from-Motion results on PhotoTourism \cite{snavely2006photo} from Image Matching Challenge 2020 \cite{jin2021image}}. We set the minimum model size to 3 following \cite{jin2021image} for both our system (``Hybrid") and COLMAP (``Point") \cite{schonberger2016structure}.}
    \label{tab::phototourism}
    \setlength{\tabcolsep}{3pt}
    \begin{tabular}{ccccc}
        \toprule
        \multirow{2}{*}{Point Feature} & \multirow{2}{*}{Method} & \multicolumn{3}{c}{AUC @ 1°/3°/5° @ N} \\
        \cmidrule(lr){3-5}
        & & $N = 5$ & $N = 10$ & $N = 25$ \\
        \midrule       
        \multirow{2}{*}{SIFT + NN} & Point & \textbf{11.4} / 24.2 / 29.8 & 29.6 / 49.2 / 56.0 & 59.9 / 80.3 / 85.9 \\
        & Hybrid & 11.3 / \textbf{24.6} / \textbf{30.5} & \textbf{31.3} / \textbf{51.5} / \textbf{58.1} & \textbf{63.3} / \textbf{82.4} / \textbf{87.5} \\
        \midrule
        \multirow{2}{*}{SP + SG} & Point & 49.0 / 75.9 / 83.8 & 62.2 / 84.5 / \textbf{90.3} & 70.6 / 88.2 / 92.6 \\
        & Hybrid & \textbf{49.2} / \textbf{76.2} / \textbf{84.1} & \textbf{62.9} / \textbf{84.6} / 90.2 & \textbf{72.3} / \textbf{89.0} / \textbf{93.1} \\
        \bottomrule
    \end{tabular}
\end{table}

\begin{table}[tb]
\begin{center}
\scriptsize
\setlength{\tabcolsep}{3pt}
\caption{Comparison between our full SfM system and the post-refinement method described in LIMAP \cite{Liu_2023_LIMAP}. }
\begin{tabular}{clccc}
\toprule
Dataset & Method & AUC @ 1°/3°/5°/10°$\uparrow$ & Valid Reg. $\uparrow$ \\
\midrule
\multirow{3}{*}{Hypersim} & COLMAP \cite{schonberger2016structure} & 71.3 / 82.5 / 85.0 / 86.8 & 93.7\% \\
& COLMAP \cite{schonberger2016structure} $\rightarrow$ LIMAP BA \cite{Liu_2023_LIMAP} & 78.6 / 84.2 / 86.5 / 87.3 & 93.8\% \\
& Ours & \textbf{82.1} / \textbf{86.6} / \textbf{87.6} / \textbf{88.3} & \textbf{93.9\%} \\
\midrule
\multirow{3}{*}{ETH3D} & COLMAP \cite{schonberger2016structure} & 16.2 / 26.7 / 28.1 / 32.1 & 46.4\% \\
& COLMAP \cite{schonberger2016structure} $\rightarrow$ LIMAP BA \cite{Liu_2023_LIMAP} & 19.2 / 28.3 / 31.1 / 33.8 & 47.6\%  \\
& Ours & \textbf{24.3} / \textbf{34.8} / \textbf{37.4} / \textbf{40.8} & \textbf{59.4\%} \\
\bottomrule
\end{tabular}
\label{tab::limap_comparison}
\end{center}
\end{table}

\noindent
\textbf{Comparison to Post-Refinement in \cite{Liu_2023_LIMAP}}
To better study the effectiveness of building a full SfM system, we compare our method with the hybrid post-refinement method proposed in \cite{Liu_2023_LIMAP} with global line triangulation. As shown in Table~\ref{tab::limap_comparison}, the post-refinement method falls behind on accuracy even when most images are successfully registered, and cannot recover from the registration failure of COLMAP~\cite{schonberger2016structure} (\eg on ETH3D). Our method, on the contrary, is able to achieve more valid registrations in such challenging scenarios.

\begin{figure}[tb]
\centering

% \includegraphics[width=0.9 \linewidth, height=125pt]{figs/imgs/exp_tum/sampling_ate.png}
% \caption{Comparisons between SfM methods (COLMAP, Ours) and SLAM methods (ORB-SLAM, PLP-SLAM) using different sampling rates, evaluated on TUM RGB-D, fr1\_desk}

\setlength{\tabcolsep}{2pt} % Set the space between columns to zero

\begin{tabular}{*{5}{c}} 
\includegraphics[trim={0 80 0 80}, clip, width=0.18\linewidth, height=12pt]{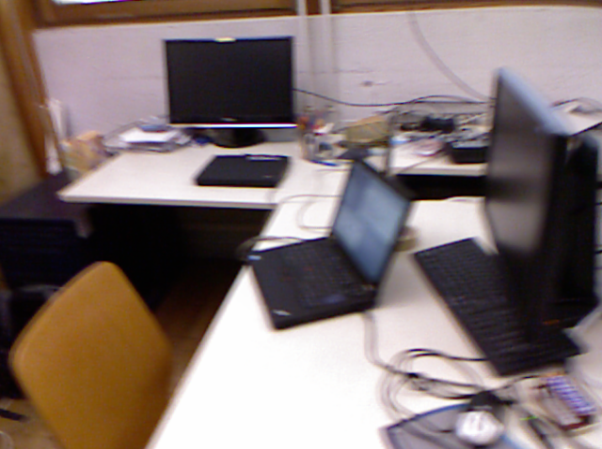} &
\includegraphics[trim={0 80 0 80}, clip, width=0.18\linewidth, height=12pt]{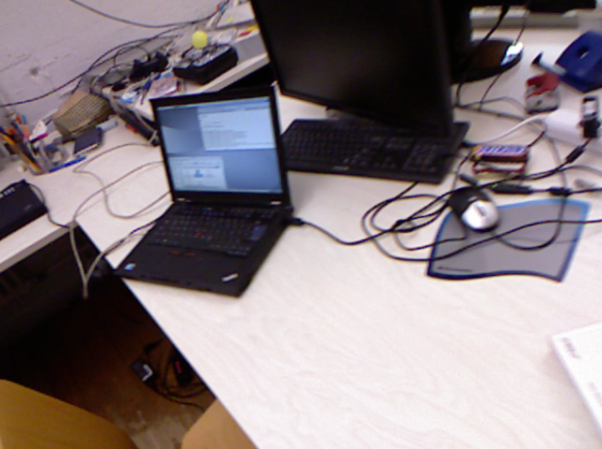} &
\includegraphics[trim={0 80 0 80}, clip, width=0.18\linewidth, height=12pt]{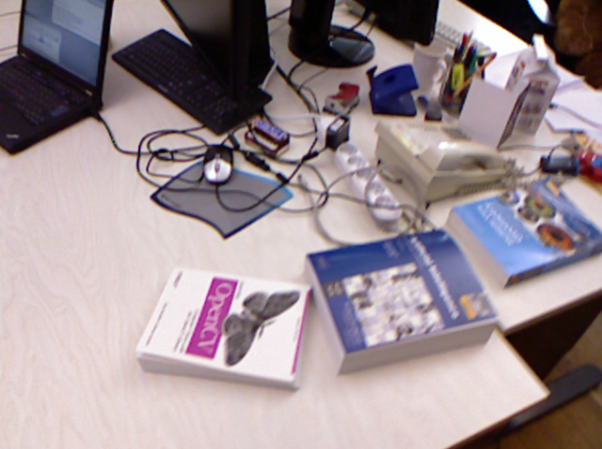} &
\includegraphics[trim={0 80 0 80}, clip, width=0.18\linewidth, height=12pt]{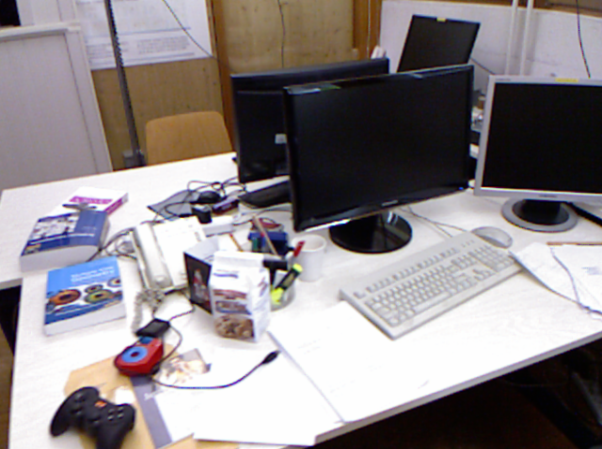} &
\includegraphics[width=0.18\linewidth, height=12pt]{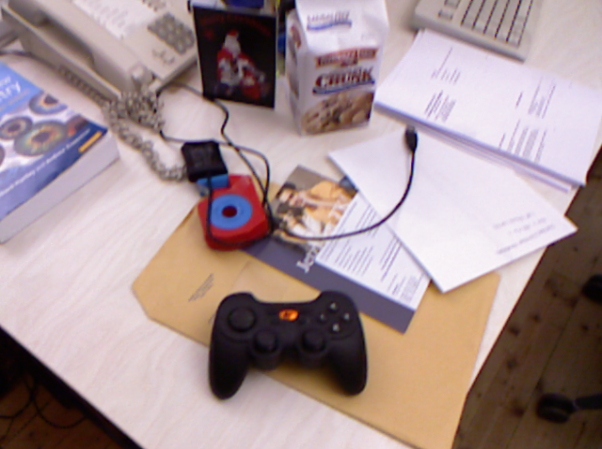}
\end{tabular}

\begin{tabular}{*{4}{c}} 
\includegraphics[width=0.24\linewidth, height=55pt]{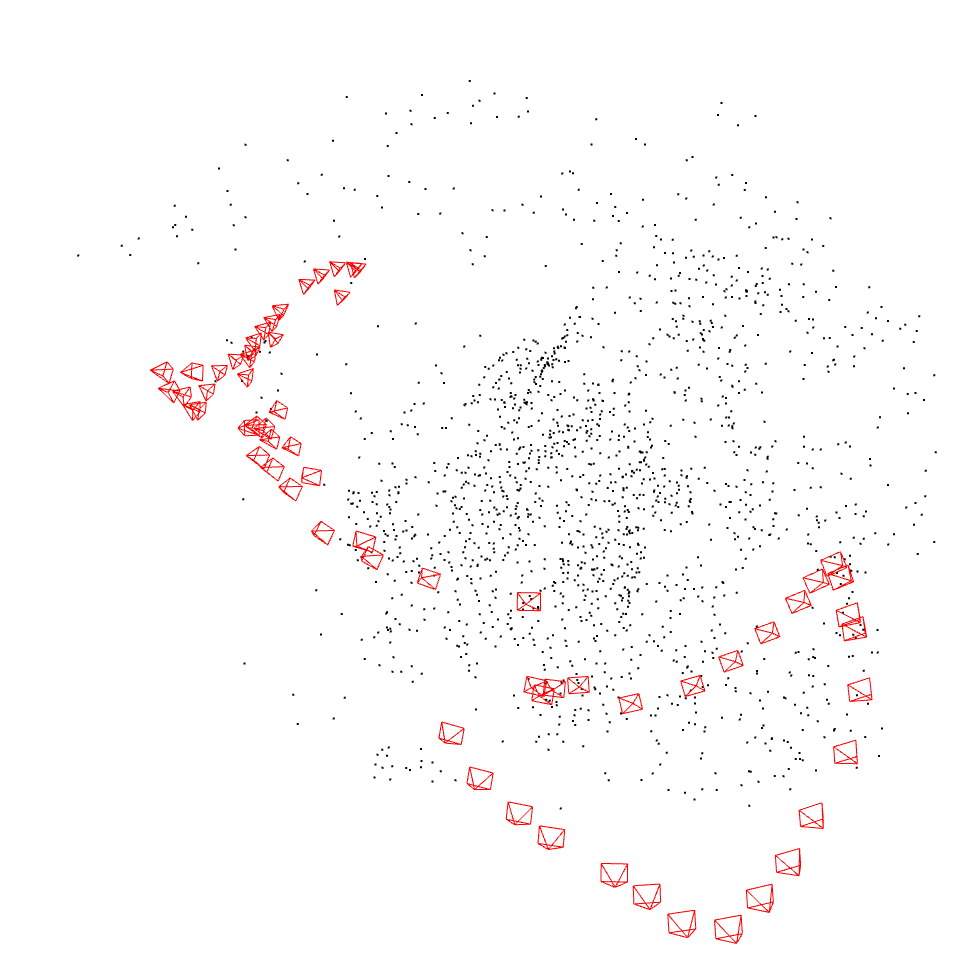} &
\includegraphics[width=0.24\linewidth, height=55pt]{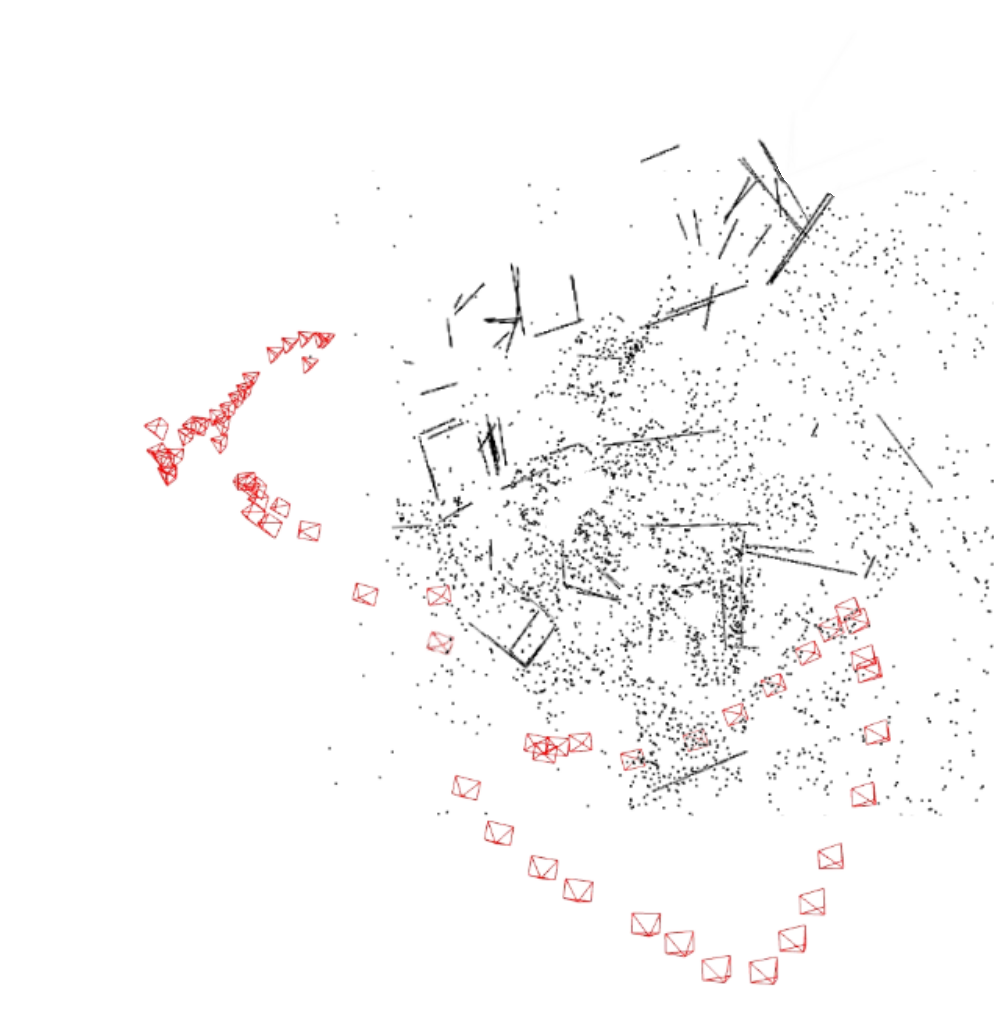} &
\includegraphics[width=0.24\linewidth, height=55pt]{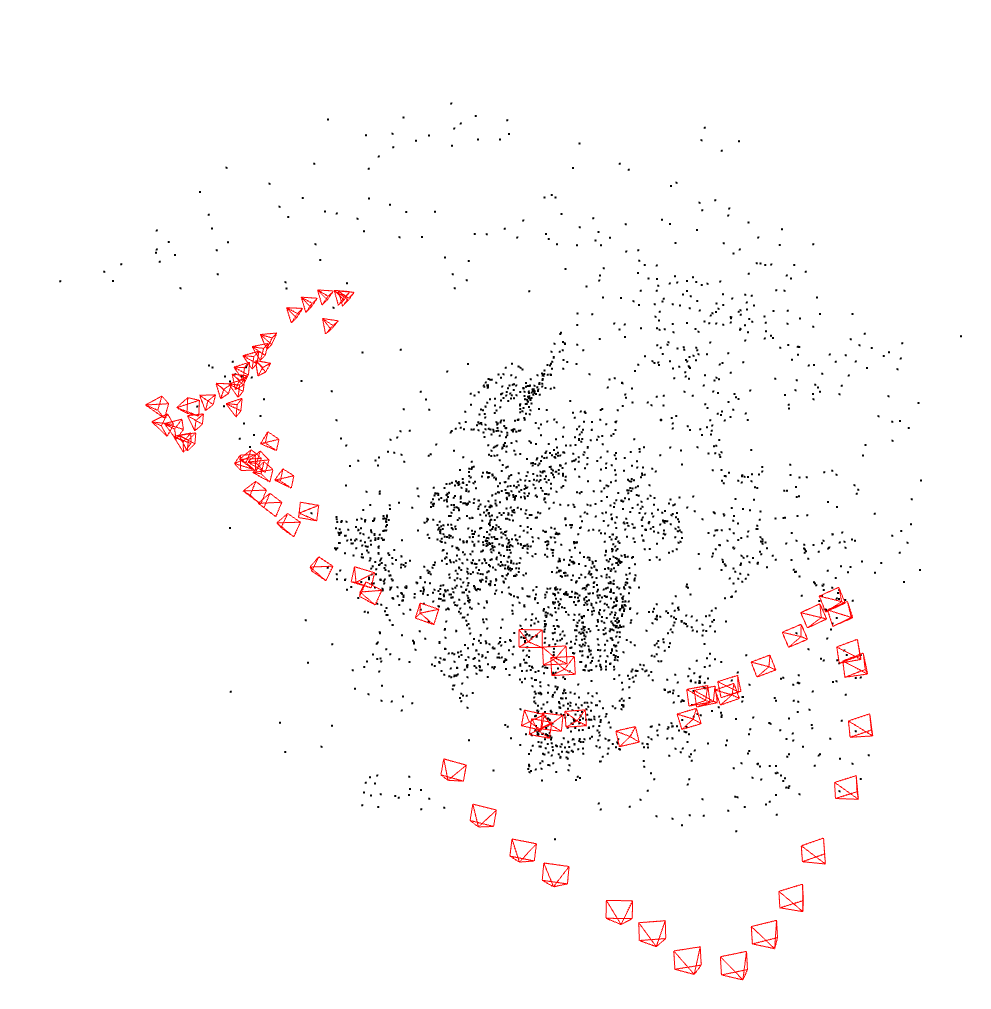} &
\includegraphics[width=0.24\linewidth, height=55pt]{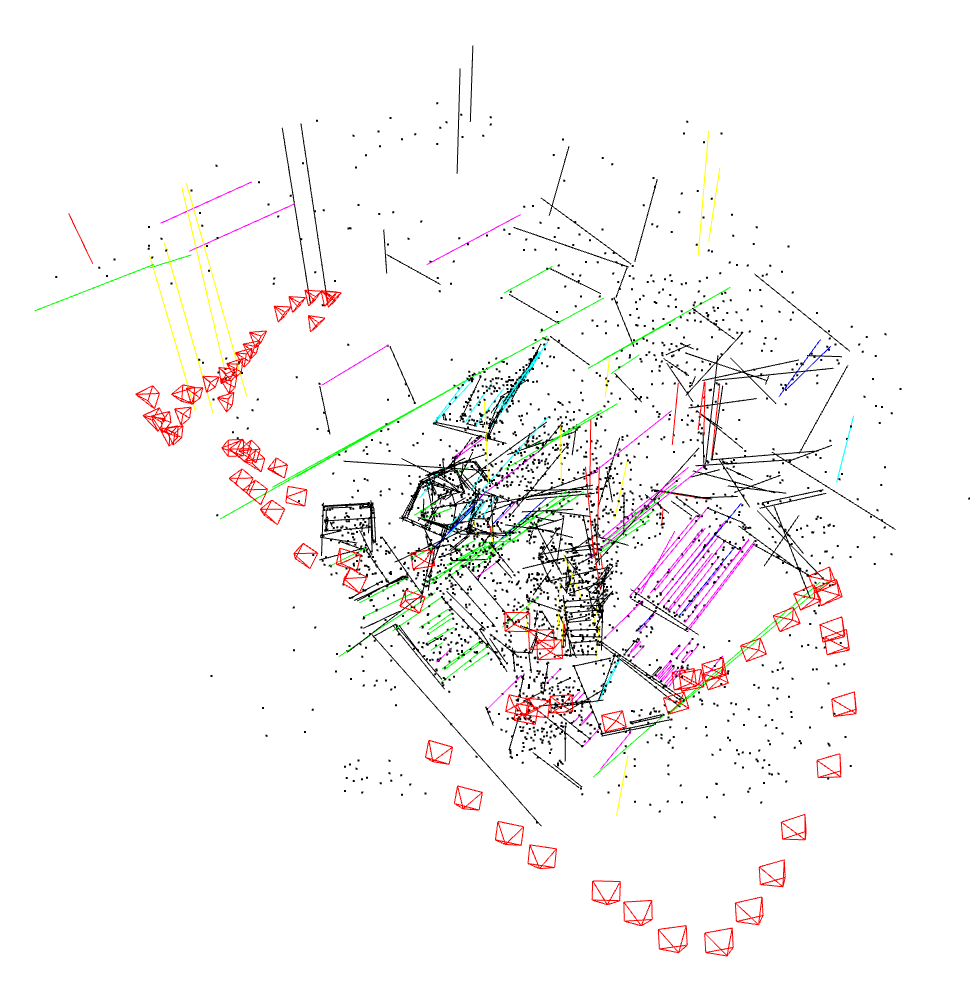} \\
\scriptsize{ORB-SLAM \cite{mur2015orb}} & \scriptsize{S. PLP-SLAM \cite{shu2023structure}} & \scriptsize{COLMAP \cite{schonberger2016structure}} & \scriptsize{Ours}
\end{tabular}

\caption{Map visualization with two SLAM methods \cite{mur2015orb,shu2023structure} on sequential data from \cite{sturm2012benchmark}. Our method is able to reconstruct much richer and more complete 3D maps, especially compared to Structure PLP-SLAM \cite{shu2023structure} that also integrates line features. }
\label{fig::exp_tum}
\end{figure}

\noindent
\textbf{Discussions on Sequential Data.}
We further present some studies with two popular SLAM systems: ORB-SLAM~\cite{mur2015orb} and Structure PLP-SLAM~\cite{shu2023structure}, the latter of which also integrates line features into its pipeline. As shown in Table~\ref{tab::tum}, while SLAM methods can achieve superior results on video sequences, it is constrained to sequential data and cannot deal with sparsely distributed input images. Moreover, in the qualitative results in Fig.~\ref{fig::exp_tum}, we show that our line maps are much richer and more complete compared to the top-performing SLAM counterpart~\cite{shu2023structure}, further highlighting the advantages of our general SfM pipeline.

\begin{table}[tb]
\centering % Center the table
\scriptsize
\caption{Studies on behaviors of SLAM pipelines \cite{mur2015orb,shu2023structure} and our SfM method under different sampling rate on \textit{fr1\_desk} from \cite{sturm2012benchmark}. While SLAM methods achieve superior results on dense frames, they suffer from low frame rates due to strong local assumptions. }
\begin{tabular}{l P{40pt} P{40pt} P{40pt} P{40pt}}
\toprule
\multirow{2}{*}{Method} & \multicolumn{4}{c}{ATE RMSE} \\
\cmidrule{2-5}
& 30 FPS & 15 FPS & 6 FPS & 3 FPS \\
\midrule
ORB-SLAM \cite{mur2015orb} & \textbf{1.27} & 1.51 & 7.42 & N/A \\
Structure PLP-SLAM \cite{shu2023structure} & 1.64 & 2.36 & 3.88 & N/A \\
Ours & 1.39 & \textbf{1.37} & \textbf{1.45} & \textbf{1.80} \\
\bottomrule
\end{tabular}
\label{tab::tum}
\end{table}

\subsection{Ablation Studies and More Insights}

\begin{table}[tb]
\centering % Center the table
\scriptsize
\caption{\textbf{Quantitative results of line reconstruction on Hypersim \cite{roberts:2021}.} Following the same evaluation protocols, our incremental triangulator reconstructs line maps of comparable quality with the global method in \cite{Liu_2023_LIMAP}. By default \cite{Liu_2023_LIMAP} uses top 10 matches. }
\setlength{\tabcolsep}{3pt}
\begin{tabular}{ccccccccccc}
\toprule
\multirow{2}{*}{Mapper} & \multirow{2}{*}{R1} & \multirow{2}{*}{R5} & \multirow{2}{*}{R10} & \multirow{2}{*}{P1} & \multirow{2}{*}{P5} & \multirow{2}{*}{P10} & \multirow{2}{*}{\# lines } & \multirow{2}{*}{\# supports } \\
& & & & & & & & & \\
\midrule
\multicolumn{1}{c}{Global (top 10 matches) \cite{Liu_2023_LIMAP}} & 133.4 & 231.9 & 258.4 & 77.2 & 89.4 & 93.2 & 731.9 & 15.4 / 22.7 \\
\midrule
\multicolumn{1}{c}{Global (raw matches) \cite{Liu_2023_LIMAP}} & 81.2 & 230.1 & 287.3 & \textbf{71.7} & \textbf{84.7} & \textbf{90.2} & 657.3 & \textbf{17.9} / \textbf{22.1} \\
\multicolumn{1}{c}{Incremental (raw matches)} & \textbf{98.8} & \textbf{260.2} & \textbf{345.1} & 68.9 & 80.5 & 85.3 & \textbf{929.2} & 14.4 / 15.6\\
\bottomrule
\end{tabular}
\label{tab::incremental_mapping}
\end{table}

\begin{table}[tb]
\begin{center}
\scriptsize
\setlength{\tabcolsep}{3pt}
\caption{\textbf{Ablation studies on different proposed components from mapping and refinement.} Numbers are reported on Hypersim \cite{roberts:2021}. }
\begin{tabular}{cr@{}lccc}
% \begin{tabular}{c@{\hspace{5pt}}l@{}lccc}
\toprule
& \multicolumn{2}{c}{\multirow{2}{*}{BA Ablations}} & \multirow{2}{*}{AUC @ 1°/3°/5°/10° $\uparrow$} \\
& \multicolumn{2}{c}{} & & \\
\midrule
& & Point (COLMAP) \cite{schonberger2016structure} & 71.3 / 82.5 / 85.0 / 86.8 \\
& & + hybrid point-line BA without caching \cite{Liu_2023_LIMAP} & 75.9 / 83.6 / 86.0 / 87.1 \\
& & + inactive support caching & 79.2 / 84.9 / 86.5 / 87.4 \\
& & + two-step refinement & 80.3 / 85.6 / 86.8 / 87.8 \\
& & + retriangulation & 80.8 / 85.9 / 87.0 / 87.9 \\
& & + VP associations & \textbf{81.2} / \textbf{86.1} / \textbf{87.2} / \textbf{88.0} \\
\bottomrule
\end{tabular} 
\label{tab::ba_ablations}
\end{center}
\end{table}

\begin{table}[tb]
\begin{center}
\scriptsize
\setlength{\tabcolsep}{3pt}
\caption{\textbf{Ablation studies on our registration modules in SfM.} Numbers are reported on both Hypersim \cite{roberts:2021} and ETH3D \cite{schops2017multi}. }
\begin{tabular}{clccc}
\toprule
Dataset & Method & AUC @ 1°/3°/5°/10°$\uparrow$ & Valid Reg. $\uparrow$  \\
\midrule
\multirow{3}{*}{Hypersim} & Point-based registration \cite{gao2003complete,schonberger2016structure} & 80.8 / 85.9 / 87.0 / 87.9 & 93.8\% \\
& + hybrid registration & 81.5 / 86.4 / 87.4 / 88.2 & 93.8\% \\
& + uncertainty for registration & \textbf{81.7} / \textbf{86.5} / \textbf{87.6} / \textbf{88.3} & \textbf{93.9\%} \\
\midrule
\multirow{3}{*}{ETH3D} & Point-based registration \cite{gao2003complete,schonberger2016structure} & 19.2 / 28.0 / 30.6 / 33.0 & 47.6\% \\
& + hybrid registration & 24.2 / 34.4 / 37.0 / 39.2 & 59.0\% \\
& + uncertainty for registration & \textbf{24.3} / \textbf{34.7} / \textbf{37.6} / \textbf{40.1} & \textbf{59.4\%} \\
\bottomrule
\end{tabular}
\label{tab::registration_ablation}
\end{center}
\end{table}

\begin{table}[tb]
    \centering
    \scriptsize
    \setlength{\tabcolsep}{2pt}
    \caption{\textbf{Results of our uncertainty-aware localization module on Cambridge \cite{kendall2015posenet} and 7Scenes \cite{7scenes}} compared with state-of-the-art point-alone \cite{hloc} and hybrid \cite{Liu_2023_LIMAP} methods. We report median errors (cm / deg) and recall on 3cm / 3deg and 5cm / 5deg. }
    \begin{tabular}{llcccc}
        \toprule
        \multirow{2}{*}{Dataset} & \multirow{2}{*}{Method} & \multicolumn{2}{c}{Point} & \multicolumn{2}{c}{Point + Line} \\
        \cmidrule(lr){3-4} \cmidrule(lr){5-6}
        & & Med. error $\downarrow$ & Recall $\uparrow$ & Med. error $\downarrow$ & Recall $\uparrow$ \\
        \midrule
        \multirow{2}{*}{Cambridge} & w/o. uncertainty 
        & 7.1 / 0.13 & 24.3 / 43.1
        & 7.0 / 0.13 & 25.4 / 45.3 
        \\
        & w. uncertainty 
        & \textbf{6.4} / \textbf{0.12} & \textbf{27.4} / \textbf{48.0} 
        & \textbf{6.3} / \textbf{0.12} & \textbf{29.0} / \textbf{48.6}   \\
        \midrule      
        \multirow{2}{*}{7Scenes} & w/o. uncertainty
        & 3.1 / 1.03 & 51.1 / 76.0 
        & 3.1 / 1.01 & 52.7 / 77.7 
        \\
        & w. uncertainty
        & \textbf{2.9} / \textbf{0.95} & \textbf{55.6} / \textbf{79.0}
        & \textbf{2.8} / \textbf{0.95} & \textbf{56.5} / \textbf{79.5}
         \\
        \bottomrule
    \end{tabular}
    \label{tab:localization}
\end{table}

% fire
\begin{figure}[tb]
\centering
\scriptsize
\setlength\tabcolsep{1pt}

\begin{tabular}{ccccc}
% left bottom right top
{\includegraphics[width=0.11\linewidth, height=38pt]{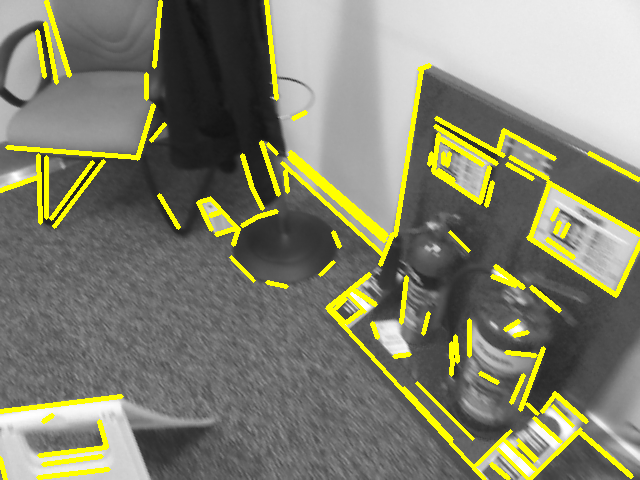}}
&
{\includegraphics[trim={100, 150, 0, 110}, clip, width=0.26\linewidth, height=38pt]{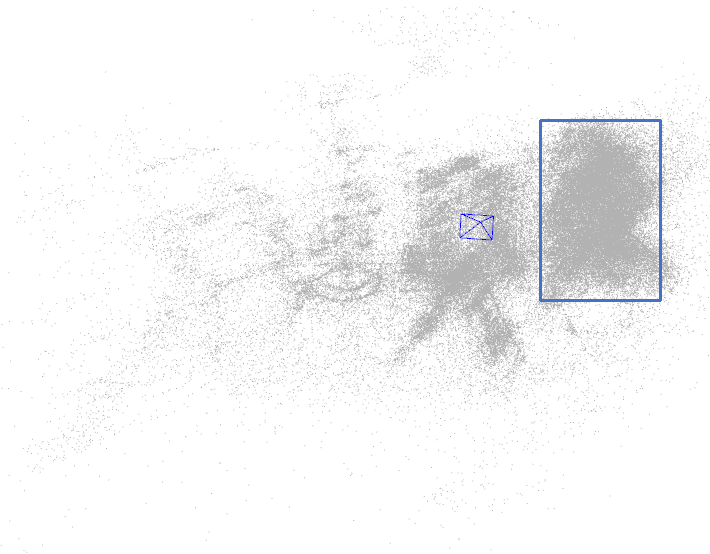}}
 &
 {\includegraphics[trim={100, 150, 0, 110}, clip, width=0.26\linewidth, height=38pt]{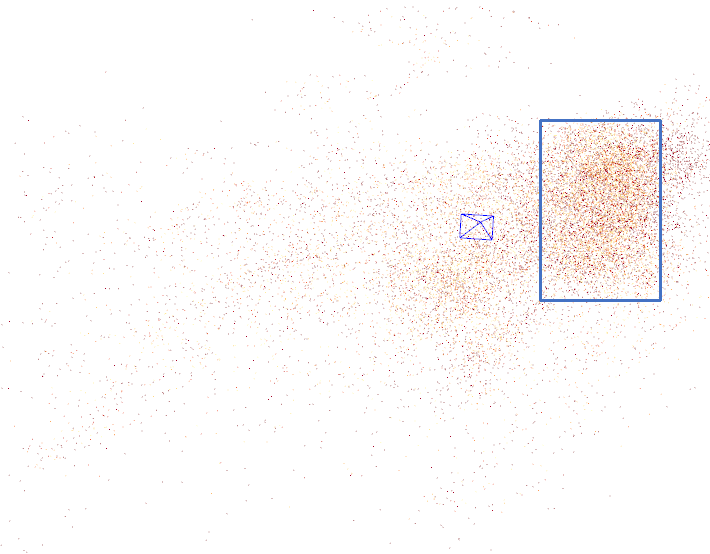}}
 &
{\includegraphics[trim={100, 150, 0, 110}, clip, width=0.26\linewidth, height=38pt]{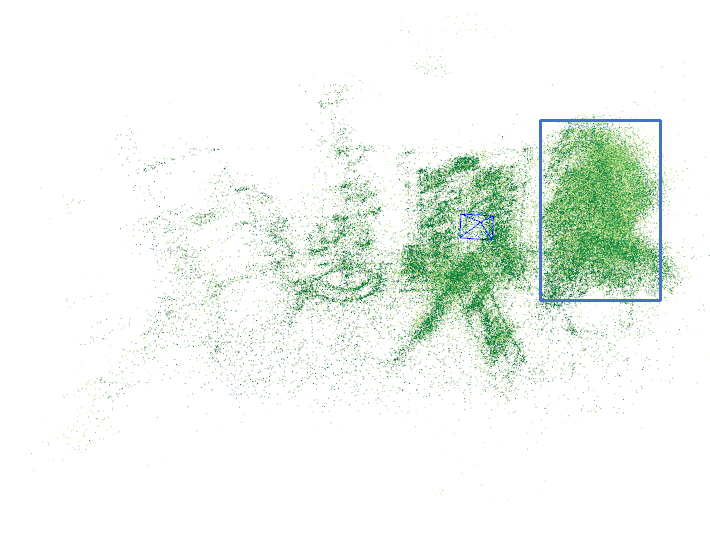}}
\\
{\includegraphics[width=0.11\linewidth, height=38pt]{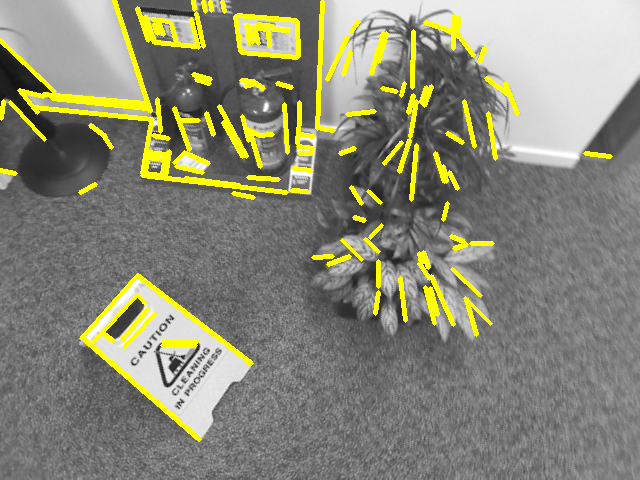}}
&
{\includegraphics[trim={0, 100, 0, 0}, clip, width=0.26\linewidth, height=38pt]{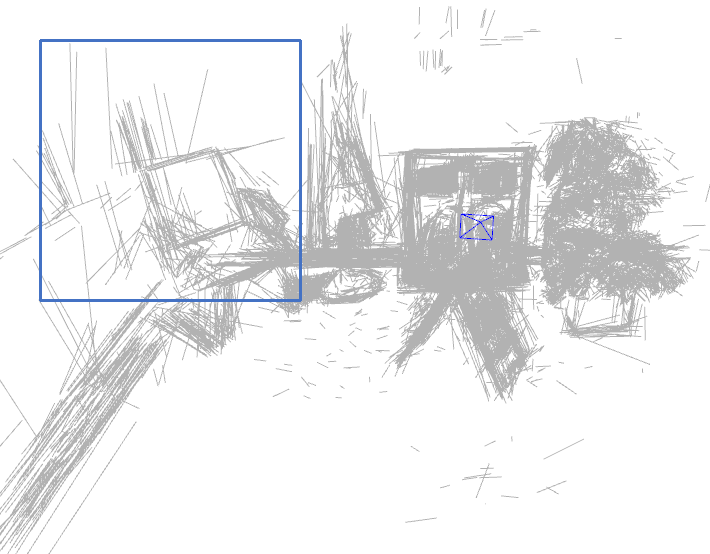}} 
&
{\includegraphics[trim={0, 100, 0, 0}, clip, width=0.26\linewidth, height=38pt]{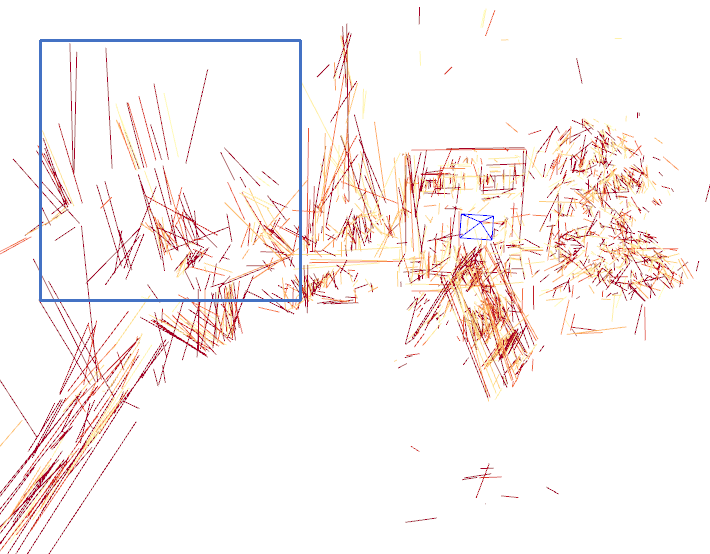}} &
{\includegraphics[trim={0, 100, 0, 0}, clip, width=0.26\linewidth, height=38pt]{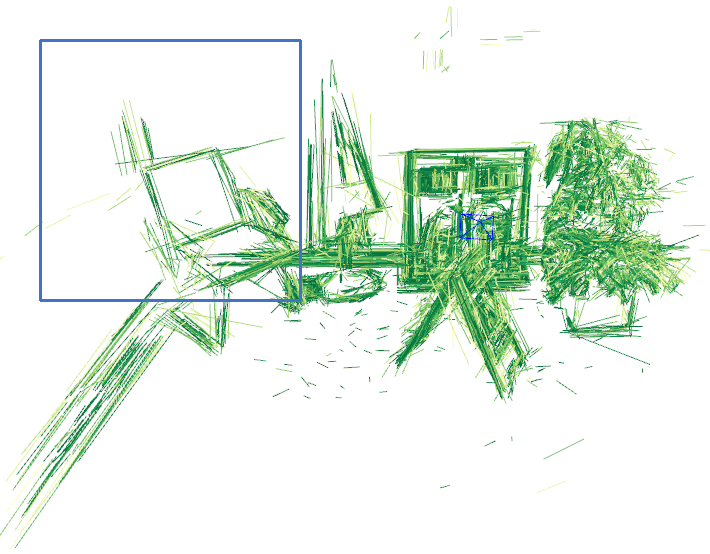}} 
\\
Images & Full map & Filtered map w/ high uncertainty & Reliable features
\end{tabular}
\centering
\caption{\textbf{Visualization of reprojected map uncertainty} on \textit{fire} from 7Scenes \cite{7scenes}. \textbf{First row:} 3D points. \textbf{Second row:} 3D lines. The reprojected uncertainty is a good indicator for identifying unstable reprojection of 3D maps with a given viewpoint.}
\label{fig::localization_covariance}
\end{figure}

\begin{figure}[tb]
    \centering
    \includegraphics[width=0.95\linewidth, height=90pt]{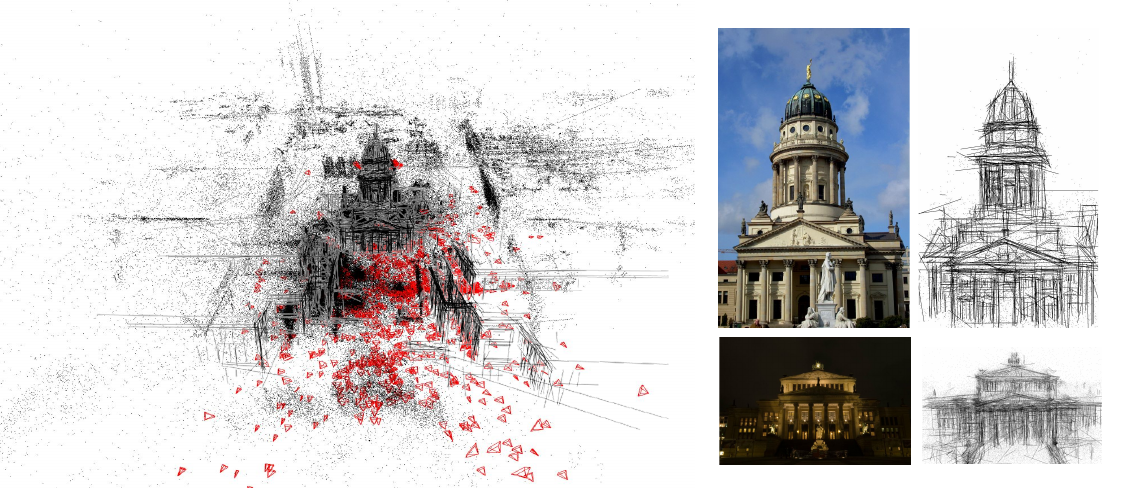}
    \caption{Hybrid reconstruction on Gendarmenmarkt (1,463 images) from \cite{wilson2014robust}. }
    \label{fig:large_scale}
\end{figure}

\noindent
\textbf{Mapping.} We first study our incremental line triangulation module in terms of line reconstruction quality. Specifically, we use the proposed module to progressively triangulate 2D images with ground truth poses on Hypersim~\cite{roberts:2021}, which can be directly compared to the global triangulation described in \cite{Liu_2023_LIMAP}. Table \ref{tab::incremental_mapping} shows the results. Our method achieves comparable completeness and accuracy with the global methods. This is attributed to the carefully designed triangulation and maintenance strategies. In particular, thanks to the ``complete'' strategy inspired by \cite{schonberger2016structure}, our incremental triangulator achieves reasonable track length while removing the need to get all the posed images beforehand. 

\noindent
\textbf{Refinement.} 
We further study the proposed refinement strategy together with map maintenance. We perform ablation studies on different mechanisms with the original point-alone registration from COLMAP \cite{schonberger2016structure}. Results in Table \ref{tab::ba_ablations} show that each component contributes to the improvement. In particular, combining the inactive support caching and the two-step refinement method makes it possible to keep unreliable supports and tracks without corrupting the pose optimization, which speeds up the mapping process by avoiding unnecessary deletion at the early stage of track building. We include visual illustrations in Sec. F of supp. 
% This is partly attributed to the scale-invariant reliability measurement from uncertainty propagation (Fig. \ref{fig::global_covariance}).

\noindent
\textbf{Registration.} 
Lastly, we study the effects of our proposed hybrid registration module with additional line features. Table \ref{tab::registration_ablation} shows that the hybrid robust estimator consistently improves the accuracy and robustness on two different datasets, while uncertainty-aware reweighting can further improve its performance. Moreover, Table \ref{tab:localization} shows results on public localization benchmarks where our uncertainty-aware localization consistently improves upon state-of-the-art practices under both point-alone \cite{hloc} and hybrid \cite{Liu_2023_LIMAP} setup. Being able to identify the noisy features from the map (as in Fig. \ref{fig::localization_covariance}), our method increases the importance of stable 2D-3D correspondences, which can be used as a general plug-in feature in any modern localization system. 

\noindent
\textbf{Scalability.}
Since our method shares a similar design as COLMAP \cite{schonberger2016structure}, it is scalable to large-scale scenes with similar asymptotic complexity when the structural associations are disabled, while exhibiting 1-3x overhead for processing additional line features in both 2D extraction/matching and hybrid bundle adjustment. Fig. \ref{fig:large_scale} shows an example of our reconstruction on 1DSfM dataset~\cite{wilson2014robust}. 

% \noindent
% \textbf{Discussions and Future Work.}
% % \subsection{Discussions and Future Work}
% \input{tex/discussions}

\section{Conclusion}
In this paper, we present a comprehensive SfM system that, in addition to points, leverages lines and their structural relations. We improve over all of the three main steps: triangulation, refinement, and registration. Experiments and ablation studies show that our method is consistently more robust and accurate compared to the widely used point-based pipeline.
Additionally, our analytical uncertainty modeling benefits the localization task, as demonstrated on several public localization benchmarks.
Future improvements include joint (faster) point/line detection and matching, more principled 2D uncertainty modeling, and richer coverage of primitive objects. 
The system will be made open-source on top of LIMAP,  
to enable future research on SfM and downstream applications.

\noindent
\textbf{Acknowledgements.}
We thank Philipp Lindenberger, Daniel Barath, Paul-Edouard Sarlin, Zador Pataki, and Wang Zhao for their helpful feedbacks. This work has been supported by Innosuisse funding (Grant No. 100.567 IP-ICT). Viktor Larsson was supported by ELLIIT and the Swedish Research Council (Grant No. 2023-05424). 

\section*{Appendix}
\appendix
\appendix
This document provides a list of supplementary materials that accompany the main paper. The content of this supplementary material is organized as follows:

\begin{itemize}
    \item In Section \ref{sec::supp_vp_tracks}, we provide details on the construction and maintenance of vanishing point tracks during the reconstruction process.
    \item In Section \ref{sec::supp_covariance}, we present detailed derivations for uncertainty propagation from the 2D observations to the 3D points and lines bundle adjusted across views. In particular, we show how to use sensitivity analysis to derive the uncertainty of the optimized 3D line in its Pl\"ucker form. We further give details on how to use the propagated uncertainty in the geometric pipeline. 
    \item In Section \ref{sec::supp_association}, we present details on integration of point-line and VP-line associations in the hybrid bundle adjustment. We additionally discuss on the challenges on efficiency and some practical solutions on implementation. 
    \item In Section \ref{sec::supp_vp_solvers}, we discuss how to use auxiliary vanishing point correspondences to help improve absolute pose estimation (localization/registration) by providing more combinations of minimal configurations.
    \item In Section \ref{sec::supp_experiments}, we provide more details on implementation, datasets and experimental setup. 
    \item Finally, in Section \ref{sec::supp_more_results}, we provide some additional results to support the content of the main paper. 
\end{itemize}

\section{Maintenance of Vanishing Point Tracks}
\label{sec::supp_vp_tracks}

As mentioned in the main paper, when a new image is registered, we not only triangulate and update the 3D points and lines, but also construct vanishing point (VP) tracks to model the parallelism relations among lines. In practice we find that the VP matching is a relatively easy task such that computing from the consensus of line matches generally gives very reasonable results, thus making VP a good resource to help improve the structures of the 3D line maps. A 2D VP feature is represented with a 3-dimensional homogeneous coordinate, which equals the 3D direction in the local frame left multiplied by the intrinsic matrix. Thus, the cross-view consistency check for VP only involves checking the angle between two 3D directions.

Similar to points and lines, the incremental update of the VP tracks involve all the required operations as follows: 

\begin{itemize}\itemsep0pt
    \item \textbf{Continue}: \textit{extend an existing VP track.} Given a 2D VP feature, we first test if there exists a matched VP (in the previously registered images) that is already triangulated in the map and test if the reprojected 3D direction is within 3 degrees compared to the local one from the 2D VP. If so, we add the VP into the corresponding track.

    \item \textbf{Create}:  \textit{triangulate a new VP.}
    If we are unable to assign a VP to any existing track, we try to create a new 3D VP track. Since a 3D VP is a special point feature that only encodes rotation information, it has only 2 degrees of freedom in total, making the two-view triangulation problem even more overconstrained. In fact, the 3D VP direction can be computed from only one view, and the multi-view triangulation of VPs can be easily achived by taking the average over all the computed directions. Thus, we perform a simple RANSAC loop by iterating over all directions, taking the best one with the most agreement, and computing the average of all the agreed VPs to get the final 3D direction of the newly created VP track.

    \item \textbf{Merge}: \textit{merge two VP tracks into a single one.} We attempt to merge VP tracks after the triangulation of each newly registered image. Though the VP merging can be done solely on the sphere due to its limited DoFs, in our system we require shared visibility for the VP track merging to avoid wrongly chained associations from noisy tracks. Specifically, we test only on pairs of VP tracks that share at least three matches among their supports, and check if their direction is within 3 degrees. If so we merge the two together and recompute the 3D VP using all the supports by taking the averaged direction.

    \item \textbf{Complete}: \textit{recollect supports.} We test reprojection agreement on the neighbors of the included supports in the matching graph to collect potentially missing supports. This is similar to the practices for points and lines as discussed in the main paper. 
\end{itemize}

By iteratively performing those steps we can maintain 3D VP tracks of reasonably good quality, which not only enables richer 3D maps with structural information but also helps on both the refinement and registration. 

At the step of hybrid refinement, the presence of VP can help regularizing the line maps by enforcing structural constraints. We perform the active supports caching and two-step refinement on the VP tracks as discussed in the main paper. Specifically, we only use the VP tracks with at least three active supports at pose optimization, and perform fixed-pose VP optimization and active label update afterwards. The fixed-pose VP optimization only involves a straightforward multi-view triangulation process. As previously discussed, we perform a one-point LO-RANSAC \cite{chum2003locally} by iterating over all the supports, selecting the best one with the most agreement, and taking the average over all the inliers.

\section{Full Derivations on Uncertainty Propagation}
\label{sec::supp_covariance}

In this section, we provide detailed derivations on propagating uncertainty from 2D observations to the optimized 3D line. In the following parts, we first provide some backgrounds on covariance propagation and the representation of 3D line. Then, we present the proposed formulation with the assistance of sensitivity analysis and provide details on the validity tests. Finally, we discuss how to use the acquired 3D uncertainty at refinement and registration (localization).

\subsection{Background}

\subsubsection{Covariance Propagation}
For a standard non-linear least squares problem with the optimized variable $\vec{x}$ and the observations $\vec{y}$:

\begin{equation}
\label{eq::least_squares_problem}
    \vec{x}^* = \argmin_{\vec{x}} \Vert f(\vec{x}) - \vec{y} \Vert^2,
\end{equation}

we can directly propagate the uncertainty from the observation $\vec{y}$ to the optimal solution $\vec{x}^*$ upon the assumption that the noise of the observations $\vec{y}$ follows the distribution $\vec{y} = f(\vec{x}) + N(\vec{0}, \vec{1})$. This can be achieved as:

\begin{equation}
    \vec{\Sigma}_{\vec{x}^*} = (J_f^T(\vec{x}^*)J_f(\vec{x}^*))^{-1},
\end{equation} 

where $J_f$ is the Jacobian of the function $f(\cdot)$. This corresponds to using the approximate Hessian as in \cite{sarlin2022lamar}. 

For a non-uniform distribution $\vec{y} = f(\vec{x}) + N(\vec{0}, \vec{S})$, the clean formulation is provided in \cite{ceres} when the residuals in the least squares problem is rescaled with $\vec{S}^{-1/2}$, which is generally the practice in factor graphs \cite{factor_graphs_for_robot_perception}. With the original formulation, one can also use the first-order approximation \cite{seber2003nonlinear} for propagating covariance from the observation noise in the linear form:

\begin{equation}
    \vec{\Sigma}_{\vec{x}^*} = \frac{\partial \vec{x}^*}{\partial \vec{y}}\vec{\Sigma}_y\frac{\partial \vec{x}^*}{\partial \vec{y}}^T = \frac{\partial \vec{x}^*}{\partial \vec{y}}\vec{S}\frac{\partial \vec{x}^*}{\partial \vec{y}}^T
    \label{eq:jacobian_based_propagation}
\end{equation}
\begin{equation}
    \frac{\partial \vec{x}^*}{\partial \vec{y}} = J_f^{\dagger}(\vec{x}^*) = (J_f^T(\vec{x}^*)J_f(\vec{x}^*))^{-1}J_f^T(\vec{x}^*)
    \label{eq:least_squares_jacobian_based_propagation}
\end{equation}

where $J_f^{\dagger}$ corresponds to the Moore-Penrose Inverse of $J_f$. In our case, the discussion above applies to the 3D point optimized with the reprojection error across multiple views, where the function $f$ corresponds to the perspective reprojection and $J_f$ is its Jacobian. 

Note that this only applies to the case where the error uncertainty can be directly propagated from the observation uncertainty. In other words, the partial derivative of the residual over the observation is not dependent on the optimal $x^*$, which can be formulated in the context of non-linear least squares regression. However, the multi-view line optimization with the endpoint-to-line distance as residuals does not fall into this category, as discussed in Sec. \ref{sec::supp_covariance_line_residuals}. 

\subsubsection{Representation of an Infinite 3D Line}
While in the final reconstruction a 3D line is  represented with its two endpoints, its optimization across multiple views is generally operated on its infinite form due to inconsistent endpoint observations in 2D. This optimization is followed by endpoint unprojection to decide the spatial extent of the 3D line segment. Note that the infinite form is more crucial as it is used in both bundle adjustment and localization, while the endpoints are mainly for correct track merging, robust point-to-line association and final map visualization. Thus, to be able to correctly propagate the 3D uncertainty for a optimized 3D line across views, we need to first study the representation of a 3D infinite line in the optimization. 

An infinite line has 4 degrees of freedom (DOF) and is generally represented in its Pl\"ucker coordinates \cite{hartley2003multiple}. In the optimization \cite{Liu_2023_LIMAP}, the orthonormal representation \cite{bartoli2005structure} of the Pl\"ucker coordinates is generally used to minimally constrain the 3D infinite line.

\noindent
\textbf{Pl\"ucker Coordinates.} 
A 3D line in Pl\"ucker coordinates can be represented by
two vectors, namely $\vec{L} = \begin{bmatrix} \vec{d} & \vec{m} \end{bmatrix}$, where $\vec{d}$ is the direction vector of the line and $\vec{m}$ is the vector normal to the plane containing the origin and the line. 
Given a 3D line segment with its two endpoints $(\vec{x}_s, \vec{x}_e)$, the Pl\"ucker coordinates of its corresponding infinite line is
\begin{equation}
    \tilde{\vec{L}} =
    \begin{bmatrix}
        \tilde{\vec{x}}_s - \tilde{\vec{x}}_e   \\
        \tilde{\vec{x}}_s \times \tilde{\vec{x}}_e  
    \end{bmatrix}
    = \begin{bmatrix}
        \tilde{\vec{d}}   \\
        \tilde{\vec{m}}
    \end{bmatrix}
\end{equation}
where $\tilde{\vec{x}}$ denotes the homogeneous coordinates of the endpoints and $\tilde{\vec{d}}$, $\tilde{\vec{m}}$ refer to the the unnormalized vectors.
Note that Pl\"ucker coordinates are a homogeneous representation, so all pairs $\begin{bmatrix} \alpha \tilde{\vec{d}} & \alpha \tilde{\vec{m}} \end{bmatrix} (\alpha \neq 0)$ represent the same infinite line.

We use here $\tilde{\vec{L}} = \begin{bmatrix} \tilde{\vec{d}} & \tilde{\vec{m}} \end{bmatrix}$ to denote the unnormalized Pl\"ucker coordinates and $\vec{L} = \begin{bmatrix} \vec{d} & \vec{m} \end{bmatrix}$ to denote normalized Pl\"ucker coordinates (displacement $\Vert \vec{d} \Vert = 1$), so that we have 
\begin{equation}
   \vec{L} = 
   \begin{bmatrix}
        \vec{d}   \\
        \vec{m}
    \end{bmatrix} = 
    \frac{1}{\Vert \vec{\tilde{d}} \Vert}
   \begin{bmatrix}
        \tilde{\vec{d}}   \\
        \tilde{\vec{m}}
    \end{bmatrix}
\end{equation}
In this way, the Pl\"ucker coordinates establish a one-to-one correspondence between the 4 DoF infinite lines and points.

\noindent
\textbf{Orthonormal Representation.} 
Since the Pl\"ucker coordinates are over-parameterized, we follow \cite{Liu_2023_LIMAP} to use their orthonormal representation during optimization, which was initially introduced in \cite{bartoli2005structure}. Specifically, the 4-DOF minimal representation of the Pl\"ucker coordinates is formulated as $\vec{\Phi} = [\vec{\theta}, \rho] \in \mathbb{R}^4$, which can be computed by QR decomposition.
\begin{equation}
\begin{array}{lll}
    \vec{L} 
    & = \begin{bmatrix}
            \vec{d} | \vec{m}
        \end{bmatrix} \\ 
    & =\begin{bmatrix}
            \frac{\vec{d}}{||\vec{d}||} & 
            \frac{\vec{m}}{||\vec{m}||} & 
            \frac{\vec{d} \times \vec{m}}{||\vec{d} \times \vec{m}||}
        \end{bmatrix}
        \begin{bmatrix}
            ||\vec{d}|| & 0 \\
            0 & ||\vec{m}|| \\
            0 & 0
        \end{bmatrix}
        \\
    & 
    \propto
    \vec{U}
    \begin{bmatrix}
        w_1 & 0 \\
        0 & w_2 \\
        0 & 0
    \end{bmatrix}
\end{array}
\label{eq:pluker_coor}
\end{equation}

Then, the orthonormal representation of a line can be formulated as 
\begin{equation}
	\vec{L}(\vec{\Phi}) = (\vec{U}(\vec{\theta}), \vec{W}(\rho)) \in SO(3)\times SO(2),
\end{equation}
where
\begin{equation}
    \vec{U} =
    \begin{bmatrix}
        \vec{u}_1 & \vec{u}_2 & \vec{u}_3
    \end{bmatrix},
    \vec{W} =
    \begin{bmatrix}
        w_1 & -w_2 \\
        w_2 & w_1
    \end{bmatrix} = 
    \begin{bmatrix}
        \cos{\rho} & -\sin{\rho} \\
        \sin{\rho} & \cos{\rho}
    \end{bmatrix}
\end{equation}
So the Pl\"ucker coordinates can be represented as 
\begin{equation}
   \begin{bmatrix}
        \tilde{\vec{d}}   \\
        \tilde{\vec{m}}
    \end{bmatrix} = 
    \begin{bmatrix}
        w_1\vec{u}_1 \\
        w_2\vec{u}_2
    \end{bmatrix} = 
    \begin{bmatrix}
        \cos(\rho)\vec{u}_1 \\
        \sin(\rho)\vec{u}_2
    \end{bmatrix}
\label{eq:plucker}
\end{equation}

Note that here we use axis-angle representation at uncertainty derivation for $\vec{U}(\vec{\theta})$, such that $\vec{\theta} = \theta \vec{\eta}$, where $\theta$ is the rotation angle and $\vec{\eta}$ is the rotation axis:
\begin{align}
\label{eq:so3_theta}
    \theta & = \Vert \vec{\theta} \Vert = \arccos(\frac{\text{tr}(\vec{U}) - 1}{2}) \\
    \vec{\eta} &= \frac{\vec{\theta}}{\Vert \vec{\theta} \Vert} = 
\label{eq:so3_eta}
	\frac{1}{2\sin{\Vert{\vec{\theta}}\Vert}}
	\begin{bmatrix}
		\vec{U}_{32} - \vec{U}_{23} \\
		\vec{U}_{13} - \vec{U}_{31} \\
		\vec{U}_{21} - \vec{U}_{12}
	\end{bmatrix}
\end{align}

\noindent
\textbf{Projection Matrix for Lines.} 
\label{sec:refinement_background_projection}
For lines in their Pl\"ucker form, the perspective projection can be done by either constructing the Pl\"ucker matrix \cite{hartley2003multiple,Liu_2023_LIMAP} or constructing the projection matrix for lines \cite{hartley2003multiple}. We use the latter one for the simplicity at derivation which is formulated as:

\begin{equation}
    \tilde{\vec{l}} = \vec{P}_l \vec{L} = 
    \begin{bmatrix}
        l_1 \\ l_2 \\ l_3
    \end{bmatrix},
\end{equation}
where $\tilde{\vec{l}}$ are the unnormalized coordinates of the back-projected 2D line and $\vec{P}_l$ is the line projection matrix. The construction of $\vec{P}_l$ can be done as $\vec{P}_l =     
    \begin{bmatrix}
        \vec{K}_l & \vec{0}
    \end{bmatrix}\vec{H}$, where:

\begin{equation}
    \vec{H} = 
    \begin{bmatrix}
        [\vec{t}]_{\times} \vec{R} & \vec{R} \\
        \vec{R} & \vec{0}
    \end{bmatrix},
    \vec{K}_l = 
    \begin{bmatrix}
        f_v & 0 & 0 \\
        0 & f_u & 0 \\
        -f_v c_u & -f_u c_v & f_u f_v
    \end{bmatrix}.
\end{equation}

Here $f_u$, $f_v$ denote the focal lengths, ($c_u$, $c_v$) denotes the location of the principal point, and $(\vec{R}, \vec{t})$ denote the extrinsic parameters.

\subsection{Covariance Propagation for the Optimal 3D Line}

\subsubsection{Residuals: Endpoint-to-Line Distance}
\label{sec::supp_covariance_line_residuals}
For the refinement of a 3D infinite line across multiple views, we optimize over its minimal parameters $\vec{\Phi} = [\vec{\theta}, \rho]\in \mathbb{R}^4$ with respect to the line reprojection error, which is generally formulated as the perpendicular distance \cite{Liu_2023_LIMAP} from the two endpoints of the 2D line observation. Denoting the two endpoint as $\vec{x}_s$ and $\vec{x}_e$ and the perpendicular distance function as $D(\cdot)$, the line reprojection error can be formulated as:

\begin{equation}
     \label{eq:line_residual}
    \vec{r} =
    \begin{bmatrix}
        r_s \\ r_e
    \end{bmatrix} = 
    \begin{bmatrix}
        D(\vec{x}_s, \vec{l}(\vec{\Phi})) \\
        D(\vec{x}_e, \vec{l}(\vec{\Phi}))
    \end{bmatrix} =
    \begin{bmatrix}
        \frac{\tilde{\vec{x}}_s^T\vec{l}}{\sqrt{l_1^2 + l_2^2}}    \\
        \frac{\tilde{\vec{x}}_e^T\vec{l}}{\sqrt{l_1^2 + l_2^2}}
    \end{bmatrix},
\end{equation}

where $\vec{l}(\vec{\Phi}) = \Pi(\vec{L}(\vec{\Phi}))$ is the reprojected line on the image with perspective projection $\Pi$. Then, the optimization problem across $N_k$ different views can be written into:
\begin{equation}
\label{eq:line_optimization_full}
        \vec{\Phi}^* 
        = \argmin_{\vec{\Phi}}E = \argmin_{\vec{\Phi}}\frac{1}{2}\sum_k^{N_k}\sum_j^{s, e}[D(\vec{x}_j^k, \Pi_k(\vec{L}(\vec{\Phi})]^2)).
\end{equation}
Here $E$ is the optimization objective and $\Pi_k$ is the line reprojection function for the $k$th view. Different from points, the residuals described in Eq. (\ref{eq:line_residual}) cannot be written in the standard from of non-linear least squares problem as in Eq. (\ref{eq::least_squares_problem}). In particular, taking the derivative of the endpoint-to-line perpendicular distance $D(\cdot)$ we have:

\begin{equation}
    \frac{\partial D(\vec{x}, \vec{l})}{\partial \vec{x}} = \frac{\vec{l}}{\sqrt{l_1^2 + l_2^2}}\frac{\partial \tilde{\vec{x}}}{\partial \vec{x}}
\label{eq:jacobian_error_detection}
\end{equation}

\begin{equation}
    \frac{\partial D(\vec{x}, \vec{l})}{\partial \vec{l}} = \frac{1}{\sqrt{l_1^2 + l_2^2}}(\tilde{\vec{x}}^T - \tilde{\vec{x}}^T\vec{l}
    \begin{bmatrix}
		\frac{l_1}{l_1^2 + l_2^2} & \frac{l_2}{l_1^2 + l_2^2} & 0
    \end{bmatrix})
\label{eq:jacobian_error_lproj}
\end{equation}

Note that in Eq. (\ref{eq:jacobian_error_detection}), the derivative of the error over the measured endpoint $\vec{x}$ depends on the reprojection $\vec{l}$, which is dependent on the optimal parameters $\vec{\Phi}^*$. This is a main difference from the point case and prevents the optimization problem to be written in the standard form as in Eq. (\ref{eq::least_squares_problem}). Thus, the uncertainty of the optimal 3D line is intractable with the Jacobian-based propagation in the least squares form with first-order approximation as in Eq. (\ref{eq:least_squares_jacobian_based_propagation}).

\subsubsection{Second-order Sensitivity Analysis}
\label{sec::supp_covariance_sensitivity_analysis}
We propose to perform uncertainty propagation for the optimized 3D line using second-order sensivity analysis \cite{fiacco1990sensitivity}. As mentioned in the main paper, this relies on the fact that:

\begin{equation}
    \frac{\partial E}{\partial \vec{\Phi}}|_{\vec{\Phi} = \vec{\Phi}^*} = \vec{0}
\end{equation}

Note that here $\vec{\Phi}^*$ is an implicit function of the input $\vec{x}_j^k$ ($j\in\{s, e\}$, $k = 0, 1, ..., N_k$). Thus, we have the following property:

\begin{equation}
    \frac{\partial^2E}{\partial \vec{\Phi} \partial \vec{x}_j^k}|_{\vec{\Phi} = \vec{\Phi}^*} = \vec{0}.
    \label{eq:sensitivity_analysis}
\end{equation}

This can be used to derive the target Jacobian $\partial \vec{\Phi}^*/ \partial \vec{x}_j^k$, which can enable uncertainty propagation with Eq. (\ref{eq:jacobian_based_propagation}). Specifically, relating Eq. (\ref{eq:line_optimization_full}) the left-hand side of Eq. (\ref{eq:sensitivity_analysis}) becomes:

\begin{equation}
    \frac{1}{2}\sum_k^{N_k}\sum_j^{s, e}\frac{\partial^2[D(\vec{x}_j^k, \Pi_k(\vec{L}(\vec{\Phi})]^2))}{\partial \vec{\Phi}\partial \vec{x}_j^k}|_{\vec{\Phi} = \vec{\Phi}^*} = \vec{0}
    \label{eq:full_sensitivity_analysis}
\end{equation}

With the denotation $\vec{l}_k(\vec{\Phi}) = \Pi_k(\vec{L}(\vec{\Phi}))$ we have:

\begin{equation}
        \frac{1}{2}\frac{\partial^2 [D(\vec{x}_j^k, \Pi_k(\vec{L}(\vec{\Phi}))]^2}{\partial \vec{\Phi} \partial \vec{x}_j^k}
        = \frac{1}{2}\frac{\partial^2 [D(\vec{x}_j^k, \vec{l}_k(\vec{\Phi}))]^2}{\partial \vec{\Phi} \partial \vec{x}_j^k} 
\end{equation}

Since we have the following property for the first-order derivatives:
\begin{align}
\label{eq:jacobian_loss_minimal}
    \frac{\partial [D(\vec{x}, \vec{l}(\vec{\Phi}))]^2}{\partial \vec{\Phi}}
    & = 2D(\vec{x}, \vec{l}(\vec{\Phi})) \frac{\partial D(\vec{x}, \vec{l}(\vec{\Phi}))}{\partial \vec{\Phi}} \notag \\
    & = 2D(\vec{x}, \vec{l}(\vec{\Phi})) \frac{\partial D(\vec{x}, \vec{l}(\vec{\Phi}))}{\partial \vec{l}(\vec{\Phi})}\frac{\partial \vec{l}(\vec{\Phi})}{\partial \vec{\Phi}}
\end{align}

\begin{align}
\label{eq:jacobian_loss_observations}
    \frac{\partial [D(\vec{x}, \vec{l}(\vec{\Phi}))]^2}{\partial \vec{x}_j^k} 
    & = 2D(\vec{x}, \vec{l}(\vec{\Phi}))\frac{\partial D(\vec{x}, \vec{l}(\vec{\Phi}))}{\partial \vec{x}_j^k} \notag \\
    & = 2D(\vec{x}, \vec{l}(\vec{\Phi}))\bigg(\frac{\partial D(\vec{x}, \vec{l}(\vec{\Phi}))}{\partial \vec{x}}\frac{\partial \vec{x}}{\partial \vec{x}_j^k} +  \notag \\
    & \frac{\partial D(\vec{x}, \vec{l}(\vec{\Phi}))}{\partial \vec{l}(\vec{\Phi})}\frac{\partial \vec{l}(\vec{\Phi})}{\partial \vec{\Phi}}\frac{\partial \vec{\Phi}}{\partial \vec{x}_j^k}\bigg)
\end{align}

Combining Eqs. (\ref{eq:jacobian_loss_minimal}) and (\ref{eq:jacobian_loss_observations}) we can get the full expansion for each term of Eq. (\ref{eq:line_optimization_full}) in Eq. (\ref{eq:full_expansion}).

\begin{table*}[tb]
\begin{align}
\label{eq:full_expansion}
\frac{1}{2}\frac{\partial^2 [D(\vec{x}, \vec{l}(\vec{\Phi}))]^2}{\partial \vec{\Phi} \partial \vec{x}_j^k}
= & \frac{\partial (D(\vec{x}, \vec{l}(\vec{\Phi})) \frac{\partial D(\vec{x}, \vec{l}(\vec{\Phi})))}{\partial \vec{\Phi}}}{\partial \vec{x_j^k}} \notag \\
= & \frac{\partial D(\vec{x}, \vec{l}(\vec{\Phi})))}{\partial \vec{\Phi}}\frac{\partial D(\vec{x}, \vec{l}(\vec{\Phi}))}{\partial \vec{x}_j^k} 
+ D(\vec{x}, \vec{l}(\vec{\Phi}))\frac{\partial^2 D(\vec{x}_j^k, \Pi_k(\vec{L}(\vec{\Phi}))}{\partial \vec{\Phi} \partial \vec{x}_j^k} \notag \\
= & \frac{\partial D(\vec{x}, \vec{l}(\vec{\Phi})))}{\partial \vec{\Phi}}\frac{\partial D(\vec{x}, \vec{l}(\vec{\Phi}))}{\partial \vec{x}_j^k}
+ D(\vec{x}, \vec{l}(\vec{\Phi}))
\bigg(\frac{\partial
(\frac{\partial D(\vec{x}, \vec{l}(\vec{\Phi}))}{\partial \vec{l}(\vec{\Phi})}
\frac{\partial \vec{l}(\vec{\Phi})}{\partial \vec{\Phi}}
)}{\partial \vec{x}_j^k} \bigg) \notag \\
= & \frac{\partial D(\vec{x}, \vec{l}(\vec{\Phi}))}{\partial \vec{l}(\vec{\Phi})}
\frac{\partial \vec{l}(\vec{\Phi})}{\partial \vec{\Phi}}
\bigg(\frac{\partial D(\vec{x}, \vec{l}(\vec{\Phi}))}{\partial \vec{x}}\frac{\partial \vec{x}}{\partial \vec{x}_j^k} + 
\frac{\partial D(\vec{x}, \vec{l}(\vec{\Phi}))}{\partial \vec{l}(\vec{\Phi})}\frac{\partial \vec{l}(\vec{\Phi})}{\partial \vec{\Phi}}\frac{\partial \vec{\Phi}}{\partial \vec{x}_j^k}\bigg) + \notag \\
& D(\vec{x}, \vec{l}(\vec{\Phi}))
\big(\frac{\partial \frac{\partial D(\vec{x}, \vec{l}(\vec{\Phi}))}{\partial \vec{l}(\vec{\Phi})}}{\partial \vec{x}}
\frac{\partial \vec{x}}{\partial \vec{x}_j^k} + 
\frac{\partial \frac{\partial D(\vec{x}, \vec{l}(\vec{\Phi}))}{\partial \vec{l}(\vec{\Phi})}}{\partial \vec{l}(\vec{\Phi})}
\frac{\partial \vec{l}(\vec{\Phi})}{\partial \vec{\Phi}} \frac{\partial \vec{\Phi}}{\partial \vec{x}_j^k}\big) 
\frac{\partial \vec{l}(\vec{\Phi})}{\partial \vec{\Phi}}
+ \notag \\
& D(\vec{x}, \vec{l}(\vec{\Phi}))\frac{\partial D(\vec{x}, \vec{l}(\vec{\Phi}))}{\partial \vec{l}(\vec{\Phi})}
\frac{\partial^2\vec{l}(\vec{\Phi})}{\partial \vec{\Phi}^2}
\frac{\partial \vec{\Phi}}{\partial x_j^k}
\end{align}
\end{table*}

By reorganizing Eq. (\ref{eq:full_expansion}) and measuring at $\vec{\Phi}^*$ we can get a linear equation with respect to the target Jacobian $\partial \vec{\Phi}^*/ \partial \vec{x}_j^k$. Since most terms can be computed in a straightforward manner, we here only provide detailed solutions for computing the first-order derivatives $\partial \vec{l}(\vec{\Phi})/\partial \vec{\Phi}$ and second-order derivatives $\partial^2 \vec{l}(\vec{\Phi})/\partial \vec{\Phi}^2$ of the reprojected line with respect to the minimal parameters $\vec{\Phi}$. 

\noindent
\textbf{Computation of First-order Derivatives $\partial \vec{l}(\vec{\Phi})/\partial \vec{\Phi}$.} According to the chain rule, we have
\begin{equation}
\label{eq:jacobian_error_minimal_first_order_chain_rule}
    \frac{\partial \vec{l}(\vec{\Phi})}{\partial \vec{\Phi}} = \frac{\partial \vec{l}(\vec{\Phi})}{\partial \vec{L}(\vec{\Phi})}
    \frac{\partial \vec{L}(\vec{\Phi})}{\partial \tilde{\vec{L}}(\vec{\Phi})}
    \frac{\partial \tilde{\vec{L}}(\vec{\Phi})}{\partial \vec{\Phi}},
\end{equation}
where the Jacobian from reprojected line to the 3D line in its Pl\"ucker coordinates is:

\begin{equation}
\frac{\partial \vec{l}(\vec{\Phi})}{\partial \vec{L}(\vec{\Phi})} = \vec{P}_l = 
	\begin{bmatrix}
		\vec{K}_l & \boldsymbol{0}
	\end{bmatrix} \vec{H}
\end{equation}

The Jacobian from normalized Pl\"ucker coordinates to unnormalied Pl\"ucker coordinates is: 
\begin{equation}
\frac{\partial \vec{L}(\vec{\Phi})}{\partial \tilde{\vec{L}}(\vec{\Phi})} = \frac{1}{\Vert \tilde{\vec{d}} \Vert}
(\boldsymbol{I}_{6\times6} - \frac{1}{\Vert \tilde{\vec{d}} \Vert^2} \tilde{\vec{L}}
	\begin{bmatrix}
		\tilde{\vec{d}} & \boldsymbol{0}
	\end{bmatrix})
\end{equation}

In the following part, we derive $\partial \tilde{\vec{L}} / \partial \vec{\Phi}$. Although the derivation of this term has already been studied in previous literature \cite{bartoli2005structure,zuo2017robust}, we will provide the solutions for completeness.

As stated in Eq. (\ref{eq:plucker}), the Pl\"ucker coordinates of an infinite line is
\begin{align}
\tilde{\vec{L}}(\vec{\Phi}) = \tilde{\vec{L}}(\vec{\theta}, \rho) = \begin{bmatrix}
        w_1\vec{u}_1 \\
        w_2\vec{u}_2
    \end{bmatrix} = 
    \begin{bmatrix}
        \cos(\rho)\vec{u}_1 \\
        \sin(\rho)\vec{u}_2
    \end{bmatrix}
\end{align}
Since we have $\vec{U} \in SO(3)$, we can compute its derivative in the Lie algebra, which is the space of skew-symmetric matrices

\begin{equation}
	\mathfrak{so}(3) = \{ \vec{\theta}^{\wedge} \in \mathbb{R}^{3\times3} | \vec{\theta} \in \mathbb{R} \} \quad  (\vec{\theta}^{\wedge} = [\vec{\theta}]_{\times})
\end{equation}
According to Rodrigues Formula, the closed-form expression for the exponential map from $\mathfrak{so}(3)$ to $SO(3)$ is 
\begin{equation}
	\vec{U} = \exp(\vec{\theta}^{\wedge}) =
	\boldsymbol{I} + 
	(\frac{\sin\Vert\vec{\theta}\Vert}{\Vert\vec{\theta}\Vert})\vec{\theta}^{\wedge} + 
	(\frac{1-\cos\Vert\vec{\theta}\Vert}{\Vert\vec{\theta}\Vert^2})\vec{\theta}^{\wedge2}
\end{equation}
Combining Eqs. (\ref{eq:so3_theta}) and (\ref{eq:so3_eta}), The logarithm map from $SO(3)$ to $\mathfrak{so}(3)$ is 
\begin{align}
    \vec{\theta}^{\wedge} &= \log\vec{U} = 
	\frac{\Vert\vec{\theta}\Vert}{2\sin\Vert\vec{\theta}\Vert}(\vec{U} - \vec{U}^T)
\end{align}
According to Baker-Campbell-Hausdorff Formulas, the permutation in $\mathfrak{so}(3)$ and $SO(3)$ is related as 
\begin{align}
    \vec{U}(\vec{\theta} + \vec{\delta\theta}) & =
	\exp((\vec{\theta} + \vec{\delta\theta})^{\wedge}) = \exp((\vec{J}_L \vec{\delta\theta})^{\wedge})\exp(\vec{\theta}^{\wedge}) \notag \\ 
 & = (\vec{I} + [\vec{J}_L\boldsymbol{\delta\theta}]_{\times})\vec{U},
\end{align}

where $\vec{J}_L$ is the left Jacobian of $SO(3)$ matrix: 
\begin{equation}
\vec{J}_L = \boldsymbol{I} + (\frac{1 - \cos \Vert\vec{\theta}\Vert}{\Vert\vec{\theta}\Vert^2}) \vec{\theta}^{\wedge} + (\frac{\Vert\vec{\theta}\Vert - \sin \Vert\vec{\theta}\Vert}{\Vert\vec{\theta}\Vert^3}) \vec{\theta}^{\wedge2}.
\label{eq:JL}
\end{equation}

\begin{table*}[tb]
\begin{align}
\frac{\partial \tilde{\vec{L}}(\vec{\theta}, \rho)}{\partial \vec{\theta}} & = \frac{\tilde{\vec{L}}(\vec{\theta} + \vec{\delta\theta}, \rho) - \tilde{\vec{L}}(\vec{\theta}, \rho)}{\vec{\delta\theta}} =
    \begin{bmatrix}
		\frac{\cos(\rho)(\boldsymbol{I} + [\vec{J}_L \boldsymbol{\delta\theta}]_{\times})\vec{u}_1 - 
			\cos(\rho)\vec{u}_1}{\boldsymbol{\delta\theta}} \\
		\frac{\sin(\rho)(\boldsymbol{I} + [\vec{J}_L \boldsymbol{\delta\theta}]_{\times})\vec{u}_2 - 
			\sin(\rho)\vec{u}_2}{\boldsymbol{\delta\theta}}
    \end{bmatrix} 
	= 
	\begin{bmatrix}
		\frac{-\cos(\rho)[\vec{u}_1]_{\times}\vec{J}_L \boldsymbol{\delta\theta}}{\boldsymbol{\delta\theta}} \\
		\frac{-\sin(\rho)[\vec{u}_2]_{\times}\vec{J}_L \boldsymbol{\delta\theta}}{\boldsymbol{\delta\theta}}
	\end{bmatrix} \notag \\
	&= 
	\begin{bmatrix}
		-\cos(\rho)[\vec{u}_1]_{\times} \vec{J}_L \\
		-\sin(\rho)[\vec{u}_2]_{\times} \vec{J}_L
	\end{bmatrix}
	=
	\begin{bmatrix}
		-w_1[\vec{u}_1]_{\times} \vec{J}_L \\
		-w_2[\vec{u}_2]_{\times} \vec{J}_L
	\end{bmatrix} \label{eq:derivative_plucker2minimal_theta} \\
 \frac{\partial \tilde{\vec{L}}(\vec{\theta}, \rho)}{\partial \vec{\rho}} 
 & = \frac{\tilde{\vec{L}}(\vec{\theta}, \rho + \delta\rho) - \tilde{\vec{L}}(\vec{\theta}, \rho)}{\delta\rho} =
	\begin{bmatrix}
		-\sin(\rho)\vec{u}_1 \\
		\cos(\rho)\vec{u}_2
	\end{bmatrix} 
	= 
	\begin{bmatrix}
		-w_2\vec{u}_1 \\
		w_1\vec{u}_2
	\end{bmatrix}
 \label{eq:derivative_plucker2minimal_rho}
\end{align}
\end{table*}

In this way, we can calculate the derivatives as in Eqs. (\ref{eq:derivative_plucker2minimal_theta}) and (\ref{eq:derivative_plucker2minimal_rho}). Combining the two equations we have the final Jacobian:

\begin{equation}
	\frac{\partial \tilde{\vec{L}}(\vec{\Phi})}{\partial \vec{\Phi}} = 
	\begin{bmatrix}
		-w_1[\vec{u}_1]_{\times} \vec{J}_L & -w_2\vec{u}_1 \\
		-w_2[\vec{u}_2]_{\times} \vec{J}_L & w_1\vec{u}_2
	\end{bmatrix}
\label{eq:Jacobian_plucker_minimal}
\end{equation}

Backsubstituting into Eq. (\ref{eq:jacobian_error_minimal_first_order_chain_rule}) we can correctly compute $\partial \vec{l}(\vec{\Phi}) / \partial \vec{\Phi}$ in the end.

\noindent
\textbf{Computation of Second-order Derivatives $\partial^2 \vec{l}(\vec{\Phi})/\partial \vec{\Phi}^2$.}
To be able to extend the derivation of the first-order derivatives, the only missing term from previous derivation is the second order derivatives of the unnormalized Pl\"ucker coordinates over its minimal parameters $\vec{\Phi}$, \ie $\partial^2 \tilde{\vec{L}}/\partial \vec{\Phi}^2$. 

Here we derive $\partial^2 \tilde{\vec{L}}/\partial \vec{\Phi}^2$. From Eq. (\ref{eq:Jacobian_plucker_minimal}) we have:
\begin{align}
\frac{\partial \tilde{\vec{L}}(\vec{\Phi})}{\partial \vec{\Phi}} &= 
	\begin{bmatrix}
		-w_1[\vec{u}_1]_{\times} \vec{J}_L & -w_2\vec{u}_1 \\
		-w_2[\vec{u}_2]_{\times} \vec{J}_L & w_1\vec{u}_2
	\end{bmatrix} \notag \\
 &=
	\begin{bmatrix}
		w_1\frac{\partial u_{11}}{\partial \theta_{1}} & w_1\frac{\partial u_{11}}{\partial \theta_{2}} & w_1\frac{\partial u_{11}}{\partial \theta_{3}} & \frac{\partial w_1}{\partial \rho} u_{11} \\
		w_1\frac{\partial u_{21}}{\partial \theta_{1}} & w_1\frac{\partial u_{21}}{\partial \theta_{2}} & w_1\frac{\partial u_{21}}{\partial \theta_{3}} & \frac{\partial w_1}{\partial \rho} u_{21} \\
		w_1\frac{\partial u_{31}}{\partial \theta_{1}} & w_1\frac{\partial u_{31}}{\partial \theta_{2}} & w_1\frac{\partial u_{31}}{\partial \theta_{3}} & \frac{\partial w_1}{\partial \rho} u_{31} \\
		w_2\frac{\partial u_{12}}{\partial \theta_{1}} & w_2\frac{\partial u_{12}}{\partial \theta_{2}} & w_2\frac{\partial u_{12}}{\partial \theta_{3}} & \frac{\partial w_2}{\partial \rho} u_{12} \\
		w_2\frac{\partial u_{22}}{\partial \theta_{1}} & w_2\frac{\partial u_{22}}{\partial \theta_{2}} & w_2\frac{\partial u_{22}}{\partial \theta_{3}} & \frac{\partial w_2}{\partial \rho} u_{22} \\
		w_2\frac{\partial u_{32}}{\partial \theta_{1}} & w_2\frac{\partial u_{32}}{\partial \theta_{2}} & w_2\frac{\partial u_{32}}{\partial \theta_{3}} & \frac{\partial w_2}{\partial \rho} u_{32} \\
	\end{bmatrix} 
 \label{eq:Jacobian_plucker_minimal_perentry}
\end{align}

Then we have
\begin{align}
	&
	\frac{\partial u_{i1}}{\partial \theta_{j}} = \frac{\frac{\partial \tilde{\vec{L}}}{\partial \vec{\Phi}}_{ij}}{w_1} 
	(1 \leq i \leq 3, 1 \leq j \leq 3) 
	\label{eq:derivative_uij_theta_1}\\
	&
	\frac{\partial u_{i2}}{\partial \theta_{j}} = \frac{\frac{\partial \tilde{\vec{L}}}{\partial \vec{\Phi}}_{(i+3)j}}{w_2}
	(1 \leq i \leq 3, 1 \leq j \leq 3)
	\label{eq:derivative_uij_theta_2}
\end{align}

Since for $-w_i[\vec{u}_i]_{\times} \vec{J}_L (i=1, 2)$ we have: 
\begin{align}
    \frac{\partial (-w_{i}[\vec{u}_{i}]_{\times} \vec{J}_L)}{\partial \vec{\theta}} =& -w_{i} (\frac{\partial [\vec{u}_{i}]_{\times}}{\partial \vec{\theta}} \vec{J}_L + [\vec{u}_{i}]_{\times} \frac{\partial \vec{J}_L}{\partial \vec{\theta}}) \\
    \frac{\partial (-w_{i}[\vec{u}_{i}]_{\times} \vec{J}_L)}{\partial \rho} =& -\frac{\partial w_{i}}{\partial \rho} [\vec{u}_{i}]_{\times} \vec{J}_L
\end{align}
we can use Eqs. (\ref{eq:derivative_uij_theta_1}) and (\ref{eq:derivative_uij_theta_2}) to compute $\partial [\vec{u}_{i}]_{\times} / \partial \vec{\theta}$. Then, we only need to derive $\partial \vec{J}_L / \partial \vec{\theta}$. Furthermore, according to Eq. (\ref{eq:JL}), we have:

% i cannot stand it. Let me define some macro from Tianyi finally
\newcommand{\thetaSkew}{\boldsymbol{\theta}^{\wedge}}
\newcommand{\thetaSkewsq}{\boldsymbol{\theta}^{\wedge2}}
\newcommand{\mintheta}{\vec{\theta}}
\newcommand{\norm}[1]{\lVert#1\rVert}

\begin{equation}
	\vec{J}_L
	= \boldsymbol{I} + f_g \thetaSkew + f_h \boldsymbol{\theta}^{\wedge2},  
\end{equation}
where $f_g$ and $f_h$ are defined as:
\begin{equation}
		f_g = \frac{1 - \cos \norm{\mintheta}}{\norm{\mintheta}^2}, 
		f_h = \frac{\norm{\mintheta} - \sin \norm{\mintheta}}{\norm{\mintheta}^3} 
\end{equation}
Thus, we can compute each component as follows:
\begin{align}
	\frac{\partial \vec{J}_L}{\partial \mintheta} 
	= & \frac{\partial f_g}{\partial \mintheta} \thetaSkew + f_g \frac{\partial \thetaSkew}{\partial \mintheta} + 
	\frac{\partial f_h}{\partial \mintheta} \thetaSkewsq + f_h (\frac{\partial \thetaSkew}{\partial \mintheta} \thetaSkew + \thetaSkew \frac{\partial \thetaSkew}{\partial \mintheta}) 
	\\
	\frac{\partial f_g}{\partial \mintheta} 
	= & (\frac{\sin \norm{\mintheta}}{\norm{\mintheta}^3} + \frac{2(\cos\norm{\mintheta} - 1)}{\norm{\mintheta}^4}) \mintheta^T
	\\
	\frac{\partial f_h}{\partial \mintheta} 
	= & (\frac{1 - \cos \norm{\mintheta}}{\norm{\mintheta}^4} 
	+ \frac{3(\sin \norm{\mintheta} - \norm{\mintheta})}{\norm{\mintheta}^5})\mintheta^T 
 \end{align}
where the partial derivative $\partial \thetaSkew/\partial \mintheta$ can be computed as in Eq. (\ref{eq:jacobian_thetaskew_theta}).

 \begin{table*}[tb]
 \begin{equation}
     \frac{\partial \thetaSkew}{\partial \theta_{1}} 
	=
	\begin{bmatrix}
		0 & 0 & 0 \\
		0 & 0 & -1 \\
		0 & 1 & 0
	\end{bmatrix},
	\frac{\partial \thetaSkew}{\partial \theta_{2}} = 
	\begin{bmatrix}
		0 & 0 & 1 \\
		0 & 0 & 0 \\
		-1 & 0 & 0
	\end{bmatrix},
	\frac{\partial \thetaSkew}{\partial \theta_{3}} = 
	\begin{bmatrix}
		0 & -1 & 0 \\
		1 & 0 & 0 \\
		0 & 0 & 0
	\end{bmatrix}
 \label{eq:jacobian_thetaskew_theta}
   \end{equation}
 \end{table*}

\subsubsection{Extension to Loss Kernels}
In practice the robust loss function is generally used to fight against the potential presence of outliers. In our system we use the Cauchy loss function following \cite{Liu_2023_LIMAP}. These robust loss functions are equivalent to applying a non-linear kernel function on top of the original residuals, which corresponds to the following objective:

\begin{align}
    \label{eq:line_optimization_full_kernel}
        \vec{\Phi}^* &= \argmin_{\vec{\Phi}}E_g \notag \\
        &= \argmin_{\vec{\Phi}}\frac{1}{2}\sum_k^{N_k}\sum_j^{s, e}g([D(\vec{x}_j^k, \Pi_k(\vec{L}(\vec{\Phi})]^2))).
\end{align}

Here without loss of generality we denote $g(\cdot)$ a kernel function that operates on the squared error, with $E_g$ denoting the full optimization objective. Following the same practice as in Eq. (\ref{eq:sensitivity_analysis}), we have:

\begin{equation}
    \frac{\partial^2 E_g}{\partial \vec{\Phi}\partial \vec{x_j^k}} = \vec{0}.
    \label{eq:sensitivity_analysis_kernel}
\end{equation}

Since the first-order derivative over $\vec{\Phi}$ can be written as:

\begin{equation}
    \frac{\partial \bigg(\frac{1}{2} g([D(\vec{x}, \vec{l}_k(\vec{\Phi}))]^2)\bigg)}{\partial \vec{\Phi}} = \frac{1}{2}g'(\cdot)\frac{\partial [D(\vec{x}, \vec{l}_k(\vec{\Phi}))]^2}{\partial \vec{\Phi}},
\end{equation}

\begin{table*}
\begin{align}
\label{eq:full_expansion_kernel}
    \frac{\partial (\frac{1}{2} g'(\cdot) \frac{ \partial [D(\vec{x}, \vec{l}_k(\vec{\Phi}))]^2}{\partial \vec{\Phi}})}{\partial \vec{x}_j^k} 
    =& (\frac{1}{2} \frac{\partial [D(\vec{x}, \vec{l}_k(\vec{\Phi}))]^2}{\partial \vec{\Phi}})^T \frac{\partial g'(\cdot)}{\partial \vec{x}_j^k} + g'(\cdot) \frac{\partial \frac{1}{2} \frac{\partial [D(\vec{x}, \vec{l}_k(\vec{\Phi}))]^2}{\partial \vec{\Phi}}}{\vec{x}_j^k} \notag \\
    =& (\frac{1}{2} \frac{\partial [D(\vec{x}, \vec{l}_k(\vec{\Phi}))]^2}{\partial \vec{\Phi}})^T g''(\cdot)\frac{\partial [D(\vec{x}, \vec{l}_k(\vec{\Phi}))]^2}{\partial \vec{x}_j^k} + \notag \\
    &\frac{1}{2}g'(\cdot) \frac{\partial^2 [D(\vec{x}, \vec{l}_k(\vec{\Phi}))]^2}{\partial \vec{\Phi} \partial \vec{x}_j^k}
\end{align}
\end{table*}

by expanding the LHS of Eq. (\ref{eq:sensitivity_analysis_kernel}) we have the form in Eq. (\ref{eq:full_expansion_kernel}). Note that the three derivative terms in Eq. (\ref{eq:full_expansion_kernel}) have all been discussed in the previous section with the naive loss function. Thus, we can safely extend the derivation from Sec. \ref{sec::supp_covariance_sensitivity_analysis} to complete the follow-up second-order sensitivity analysis to get the target Jacobian $\partial \vec{\Phi}^*/ \partial \vec{x}_j^k$. 

\subsubsection{Validity Test}
Our analytical uncertainty propagation on both points and lines have all been validated with numerical tests. Specifically, we can make small perturbation on each input observation $\vec{x}_j^k$ and perform optimization on top of it, which enables us to compute the Jacobian $\partial \vec{\Phi}^*/ \partial \vec{x}_j^k$ numerically. All the entries in the Jacobian matrix have less than 1\% deviation between the numerical and analytical results.

We further perform a correlation test between the propagated 3D uncertainty and the map precision on the \textit{delivery\_area} scene from ETH3D \cite{schops2017multi}, as shown in Fig. 4 in the main paper. For each point track and line track, we not only compute its analytical 3D uncertainty but also measure its distance (distance from the nearest point for each line) to the groundtruth scans provided in the dataset. For points, we use the squared root of the maximum eigenvalue of the 3x3 point covariance matrix as the scalar-valued point uncertainty. For lines, we first propagate the 3D uncertainty on the optimal infinite line into the two endpoints as discussed in Sec. \ref{sec::supp_covariance_line_endpoints} and take the squared root of the maximum eigenvalue between two endpoints as the scalar-valued line uncertainty. Then, we sort the points and lines into five bins with respect to its 3D uncertainty, and compute the precision for each bin at 1cm / 3cm / 5cm. Results from Fig. 4 in the main paper show clear correlation between the propagated 3D uncertainty and the map precision.

\subsection{Applications of the Propagated Covariance}

\subsubsection{Uncertainty Propagation for 3D Line Segments}
\label{sec::supp_covariance_line_endpoints}
The propagated 3D uncertainty for each optimal line is represented in a 4x4 covariance matrix on its minimal parameters $\vec{\Phi}$. To get a geometrically meaningful uncertainty we can propagate the covariance matrix onto its 3D endpoints using the 3D point-to-line projection in Pl\"ucker form:

\begin{equation}
    \vec{X}_\perp = \vec{X} + \vec{d} \times (\vec{m} + \vec{d} \times \vec{X}),
\end{equation}
where $\vec{X}_\perp$ is the projection of $\vec{X}$ on the 3D line $\vec{L} = \begin{bmatrix} \vec{d} & \vec{m} \end{bmatrix}$ in its Pl\"ucker form. Thus, we can easily compute the covariance of the 3D endpoint $\vec{X}_s$ (without loss of generality we consider the starting endpoint $\vec{X}_s$ here) using the operation of projecting it onto the line:

\begin{align}
    \Sigma_{\vec{X}_{s\perp}} &= \frac{\partial \vec{X}_{s\perp}}{\partial \vec{\Phi}}\Sigma_{\vec{\Phi}}\frac{\partial \vec{X}_{s\perp}}{\partial \vec{\Phi}}^T \\
    \frac{\partial \vec{X}_{s\perp}}{\partial \vec{\Phi}} &= \frac{\partial \vec{X}_{s\perp}}{\partial \vec{L}}\frac{\partial \vec{L}}{\partial \vec{\Phi}}.
\end{align}

Note that here we have $X_{s\perp} = X_s$ since $X_s$ naturally lies on the infinite optimal line. Here we give the solution of the term $\partial \vec{X}_\perp / \partial \vec{L}$ as follows:

\begin{align}
    \frac{\partial \vec{X}_\perp}{\partial \vec{m}} &= [\vec{d}]_\times \\
    \frac{\partial \vec{X}_\perp}{\partial \vec{d}} &= -[\vec{m}]_\times + (\vec{d}^T\vec{X})\vec{I} + \vec{d}\vec{X}^T - 2\vec{X}{\vec{d}}^T
\end{align}

In this way, we can get a 3x3 covariance matrix for each endpoint of the line through the whole uncertainty propagation from the noise of 2D endpoint observations.

\subsubsection{Scale-Invariant 3D Uncertainty}
To be able to identify noisy points and lines, we can use the scalar-valued uncertainty. This can be computed as the squared root of the maximum eigenvalue of the point covariance matrix and endpoint covariance matrix (described in Sec. \ref{sec::supp_covariance_line_endpoints}) respectively. 

However, to make it general, we need an uncertainty measurement that is relatively invariant to the scale of the scene, since the SfM reconstruction naturally exhibits the gauge ambiguity that makes the camera poses optimal only up to a similarity transformation. To deal with this issue, we propose to use the information from the supporting views for each track to rescale the scalar-valued uncertainty. Specifically, for each track, we divide the scalar-valued uncertainty by the median value of "depth over focal length" across its supporting views. In this way, we get a scale-invariance 3D uncertainty that is in the unit of pixels. If the focal length is the same across views, this scale-invariance measurement can be geometrically interpreted as the reprojection uncertainty from the least reliable views at a certain distance. The measurement is used to identify unreliable tracks in the refinement (as shown in Fig. 10 in the main paper).

\subsubsection{Reprojection Uncertainty}

\begin{figure}[tb]
\scriptsize
\setlength\tabcolsep{1pt}
\begin{tabular}{ccc}
{\includegraphics[trim={50 50 50 50}, clip, width=0.32\linewidth]{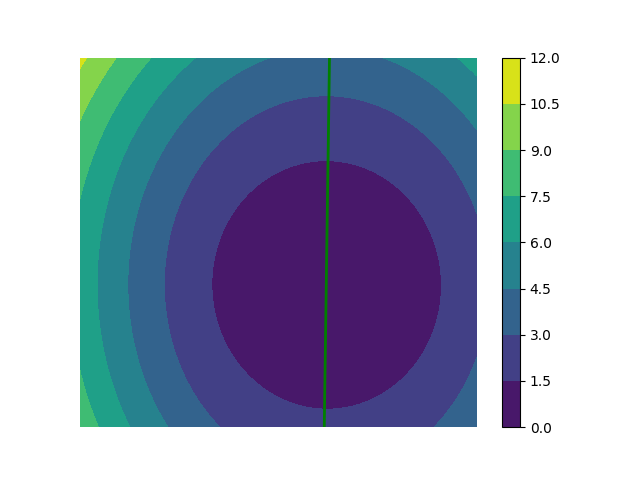}} &
{\includegraphics[trim={50 50 50 50}, clip, width=0.32\linewidth]{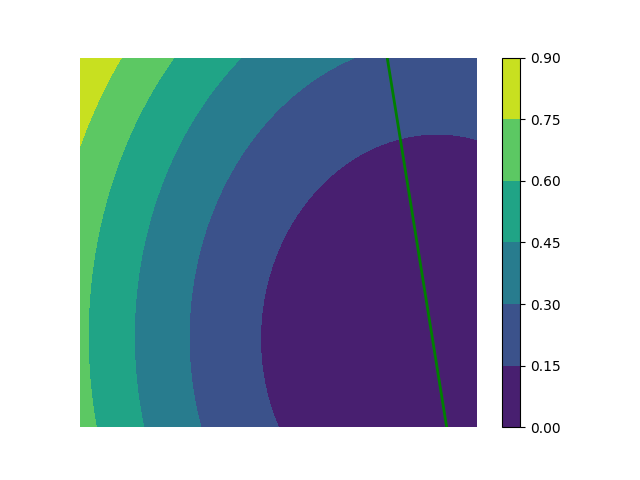}} &
{\includegraphics[trim={50 50 50 50}, clip, width=0.32\linewidth]{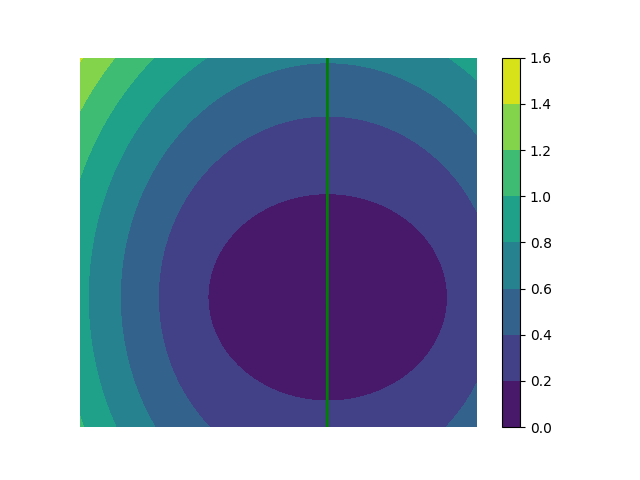}} \\
\end{tabular}
\centering
\caption{Visualization of the uncertainty of reprojection error for 3D lines at different measured locations. Different from points, the uncertainty of the line reprojection error (point-to-line distance) depends on where it is measured. }
\label{fig::ellipse}
\end{figure}

As discussed in the main paper, a point/line track with large 3D uncertainty may still be reliable at certain views for localization. Thus, to reweight correspondences based on the 3D uncertainty from the map perspective, we employ the reprojection uncertainty with respect to an initial pose rather than the global one. This can be achieved by propagating the 3D uncertainty on the infinite line parameterized by $\vec{\Phi}$ to the uncertainty of the reprojection error, which is again formulated as the endpoint-to-line distance (as in Eq. (\ref{eq:line_residual})). 

It is worth noting that different from points, the uncertainty of the reprojection error not only depends on the optimal 3D line, but also depends on the location of the measurement on the image. This is illustrated in Figure \ref{fig::ellipse}. The geometric interpretation is that: with the perturbation on the 3D infinite line with its uncertainty, the corresponding reprojected 2D line will move and rotate in the image plane. This can potentially form a relatively stationary region that has stably small reprojection error. Thus, measuring reprojection error in such regions with small reprojection uncertainty gives more reliable information on the reliability of the absolute pose proposal, therefore improving the accuracy and robustness of the localization results. The visualization of the reprojection uncertainty can be found in Figure \ref{fig::supp_localization_covariance} and Figure 11 in the main paper. We also show that the uncertainty-aware reweighting is able to contribute to consistent improvement under both point-alone and hybrid setup (as in Table 8 in the main paper).

\section{Integration of Structural Associations}

\label{sec::supp_association}

Inspired by LIMAP \cite{Liu_2023_LIMAP}, we integrate structural constraints at both  triangulation and refinement by modeling point-line associations and VP-line associations. Specifically, the 2D point-line association graph for each image can be easily constructed with 2D point-to-line distance, and the 2D VP-line association graph naturally emerges from the VP estimation \cite{toldo2008robust}. 

At triangulation, we directly follow the practice of LIMAP \cite{Liu_2023_LIMAP} to generate proposals from neighboring points and vanishing points to fight against the degeneracy issue of two-view line triangulation. 
At refinement, we can add soft constraints similarly to LIMAP \cite{Liu_2023_LIMAP} between points and lines, lines and VP by counting the connections of the corresponding supports on the 2D association graphs. While this appears to be beneficial on the quality of reconstruction and the pose accuracy, the integration of association residuals unfortunately
breaks the blockwise property of the bundle adjustment Jacobian. 

The general practice for efficient bundle adjustment involves matrix partitioning and the exploitation of Schur complement \cite{triggs2000bundle,agarwal2010bundle}, which largely benefits from the fact that the 3D map (points and lines) and the cameras form a bipartite graph in the optimization objective. However, the association residuals add edges among points and lines, making it unable to reorganize the Jacobian into the form where each point and line makes a block at the map side, inducing more off-diagonal entries in the corresponding submatrix of the Hessian for the map. While we can still perform the Schur complement trick in the larger block with connected component analysis, the relatively dense connections from the current design of using soft residuals can lead to huge blocks that significantly affect the efficiency. Nonetheless, since the Jacobian is still sparse and many soft residuals are not necessary, there is large room for efficiency improvement with more careful implementation. Also, the one without point-line association already achieves very promising results as shown in the main paper.

Note that the VP-line association does not suffer from this problem, since the VPs can be viewed as special cameras and thus moved to the camera side, keeping the graph from the optimization objective a bipartite.

\section{Absolute Pose Estimation with a Vanishing Point Correspondence}
\label{sec::supp_vp_solvers}

Since we additional construct and maintain the vanishing point (VP) tracks in the hybrid map, when registering a new image we have additional 2D-3D vanishing point correspondences from traversing the matching graph. This gives additional constraints on the rotation matrix. Specifically, one 2D-3D vanishing point correspondence gives constraints on 2 degrees of freedom on the absolute rotation, and a second one gives one another constraint on the final degree of freedom, which leaves a non-minimal overconstrained system. In this paper we only focuses on using a single vanishing point correspondence in the minimal estimation inside the hybrid RANSAC framework \cite{camposeco2018hybrid}. 

\subsection{Relation to Gravity-Aligned Solvers}
The presence of a single 2D-3D VP correspondence $(\vec{v}_{2d}, \vec{v}_{3d})$ gives the constraint on the rotation $R$ as follows:

\begin{equation}
    \vec{R}\vec{v}_{3d} = \vec{v}_{2d},
\end{equation}
where $\vec{v}_{2d}$ is the unnormalized VP direction in the local frame. This gives two degrees of freedom for the rotation matrix $\vec{R}$. Note that the constraint is equivalent to the gravity-aligned absolute pose estimation \cite{kukelova2010closed} with a tilted gravity direction. Thus, we can first rotate the system to align the 3D VP direction to the y-axis to employ all advances under the context of gravity-aligned absolute pose estimation. With two DOFs constrained in the rotation matrix, the absolute pose has 4 DOFs left, which can be reduced with 1) two 2D-3D point correspondences; 2) one 2D-3D point correspondence and one 2D-3D line correspondence. Note that two 2D-3D line correspondences do not work in this case due to dependent constraints on the rotation.

\subsection{Gravity-Aligned Absolute Pose with Two Point Correspondences}
The gravity-aligned absolute pose estimation with two point correspondences is initially studied in \cite{kukelova2010closed}. The main idea is to parameterize the final DOF of the rotation matrix in the polynomial form as follows:

\begin{equation} \label{eq:upright_rotation}
    \vec{R}(q) = \frac{1}{1+q^2}\begin{bmatrix}
        1-q^2 & 0 & 2q \\
        0 & 1+q^2 & 0 \\
        -2q & 0 & 1-q^2
    \end{bmatrix}, \quad q \in \mathbb{R}
\end{equation}

From one 2D-3D point correspondence ($\vec{x},\vec{X}$) we have:
\begin{equation}
   [\vec{x}]_\times (\vec{R}\vec{X} + \vec{t}) = 0
\end{equation}
With the availability of two such equations the problem can be finally reduced to a quadratic polynomials in $q$, which can be solved in closed form \cite{kukelova2010closed}. 

\subsection{Gravity-Aligned Absolute Pose with One Point and One Line Correspondences}

Similarly, given one point correspondence ($\vec{x},\vec{X}$) and one line correspondence $(\vec{\ell},\vec{L})$ it is also possible to recover the camera pose under known vertical direction.
Let the 3D line $\vec{L}$ be parameterized as $t \mapsto \vec{X}_L + t \vec{V}_L$, then we get the following constraint on the camera rotation
\begin{equation}
    \vec{\ell}^T \vec{R} \vec{V}_L = 0.
\end{equation}
Using the same parameterization as in Eq. (\ref{eq:upright_rotation}), this yields a quadratic equation in $q$ which can be solved in closed form.
Once the rotation is recovered, we have three linearly independent equations left,
\begin{align}
    [\vec{x}]_\times (\vec{R}\vec{X} + \vec{t}) = 0, \\
    \vec{\ell}^T (\vec{R} \vec{X}_L + \vec{t}) = 0,
\end{align}
from which we can recover the translation $\vec{t}$.

\section{More Experimental Details}
\label{sec::supp_experiments}

\subsection{Implementation Details}
Our system is implemented in C++ with Python bindings. We follow the overall design of COLMAP \cite{schonberger2016structure} and use the same hyperparameters for the point part, which enables fair comparison between ``point" and ``hybrid" setup. For the scoring at line triangulation and hybrid bundle adjustment, we follow LIMAP \cite{Liu_2023_LIMAP} for the parameter choices. The line tracks are optimized with a Cauchy loss with parameter 0.25. For the registration, we use the same weight for points and lines at scoring and local optimization. 

Similar to all existing SfM methods, those hyperparameters in our system can be changed by the users for both practical and research purpose, while using our default parameters at release should already work reasonable well on most in-the-wild cases.

We use the same hyperparameters across all the experiments. For the point feature, we use ``SIFT" \cite{lowe2004distinctive} + ``NN-ratio" and ``superpoint\_max" \cite{detone2018superpoint} + ``superglue outdoor" \cite{sarlin2020superglue} from HLoc \cite{hloc}. For the line feature, vanishing point estimation, and construction of 2D association graph, we follow the official implementation of LIMAP \cite{Liu_2023_LIMAP}. We use exhaustive matching for both the point-alone baseline COLMAP \cite{schonberger2016structure} and our method across all datasets.

\subsection{Datasets and Evaluation}
We test the proposed SfM system on multiple public datasets to verify its effectiveness. We mainly use the following two metrics at the evaluation:
\begin{itemize}\itemsep0pt
    \item \textbf{Valid Registrations}: While the number of registered images is often used in the SfM evaluation, different methods can have different tolerance criteria on rejecting the registrations. This can potentially make the evaluation metric unfair when, for example, one system registers all the images with some of them totally deviating from the groundtruth. Motivated by these observations, we propose to evaluate using the valid registrations. Specifically, we first perform robust model alignment between the SfM predictions and the groundtruth and then count the number of images that are within 5cm / 5deg to the groundtruth. In our implementation we use the interfaces from COLMAP \cite{schonberger2016structure} for robust alignment.

    \item \textbf{Relative pose AUC}: Following the general practice on SfM evaluation \cite{jin2021image}, we compute relative pose errors over all image pairs exhaustively with respect to the groundtruth. For those pairs with one or both images not registered in the SfM reconstruction, we set a maximum relative pose error of 180 degree. AUC at different thresholds (in degrees) are reported on all the errors. 
\end{itemize}

Hypersim \cite{roberts:2021} is a photorealistic synthetic dataset that is used for holistic indoor scene understanding. We follow the practice of LIMAP \cite{Liu_2023_LIMAP} to test on the first eight scenes and resize the image to a maximum dimension of 800. Each of the scene has around 100 images. The average across all the tested scenes are reported for each metric. For the evaluation of line reconstruction in Table 5, we use the evaluation tools in the code release of \cite{Liu_2023_LIMAP}.

ETH3D \cite{schops2017multi} is a real world dataset that includes unordered images in both indoor and outdoor environments. We use the training split of DSLR images, which has a total of 13 scenes. We resize each image to a maximum dimension of 756, which is of the same size of the provided groundtruth depth images. This makes it faster to run our default line matcher GlueStick \cite{pautrat2023gluestick}, while advances on the efficiency of line matching can further reduce the runtime at feature detection and matching. For evaluation we use the same two metrics as discussed with respect to the groundtruth poses provided in the data. 

We further validate our method on the PhotoTourism data \cite{snavely2006photo} from Image Matching Benchmark 2020 \cite{jin2021image}. Specifically we test on the validation split which consists of three scenes in total: \textit{Reichstag}, \textit{Sacre Coeur}, and \textit{Saint Peter's Square}. The official setup \cite{jin2021image} runs COLMAP \cite{schonberger2016structure} on all images as a pseudo groundtruth and evaluates on a collection of small bags. Not only does this induce a small bias towards point-based methods, but the dataset is also with limited presence of structured line features. Nonetheless, we still show consistent improvements over point-alone methods with different types of features. Note that since the official groundtruth construction on this dataset employs a radial distortion model which does not favor line detection, we first perform undistortion on all the images and run SfM for both our system and the baseline COLMAP \cite{schonberger2016structure} with known intrinsics on the undistorted images.

\subsection{Details on Visual Localization}
To verify the effectiveness of our uncertainty-aware visual localization module, we perform visual localization experiments on two public datasets: Cambridge \cite{kendall2015posenet} and 7Scenes \cite{7scenes}. We follow the same experimental setup of HLoc \cite{hloc} and LIMAP \cite{Liu_2023_LIMAP} to ensure fair comparison. Specifically, we use the triangulated point map from COLMAP \cite{schonberger2016structure} and line map from LIMAP \cite{Liu_2023_LIMAP}. We use NetVLAD \cite{arandjelovic2016netvlad} for image retrieval and SuperPoint \cite{detone2018superpoint} + SuperGlue \cite{sarlin2020superglue} as the point feature for both datasets. For line features, following LIMAP \cite{Liu_2023_LIMAP} we use LSD \cite{von2008lsd} on Cambridge and SOLD2 \cite{pautrat2021sold2} on 7Scenes \cite{7scenes}.

\section{Additional Results}
\label{sec::supp_more_results}

\begin{figure}[tb]
\scriptsize
\setlength\tabcolsep{1pt}
\begin{tabular}{cccc}
{\includegraphics[width=0.25\linewidth, height=48pt]{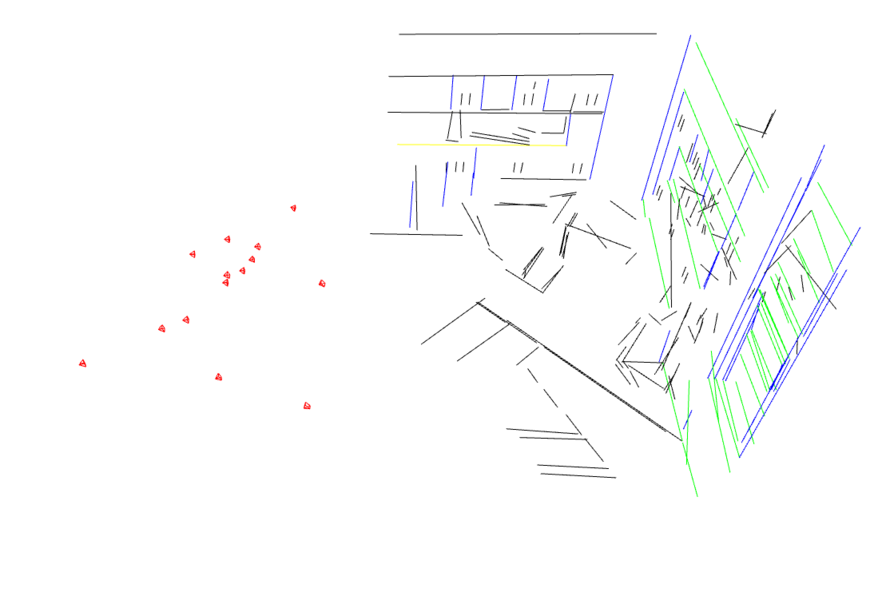}} &
{\includegraphics[width=0.25\linewidth, height=48pt]{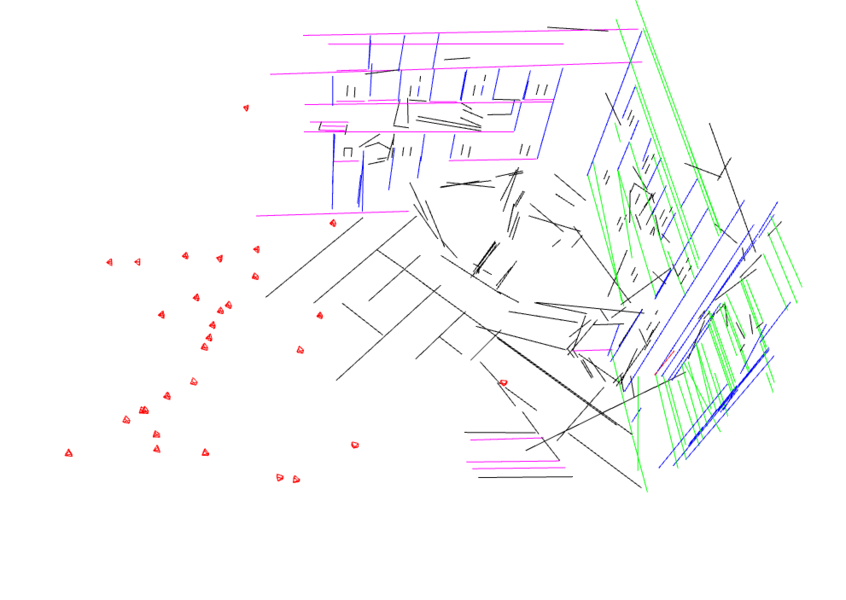}} &
{\includegraphics[width=0.24\linewidth, height=48pt]{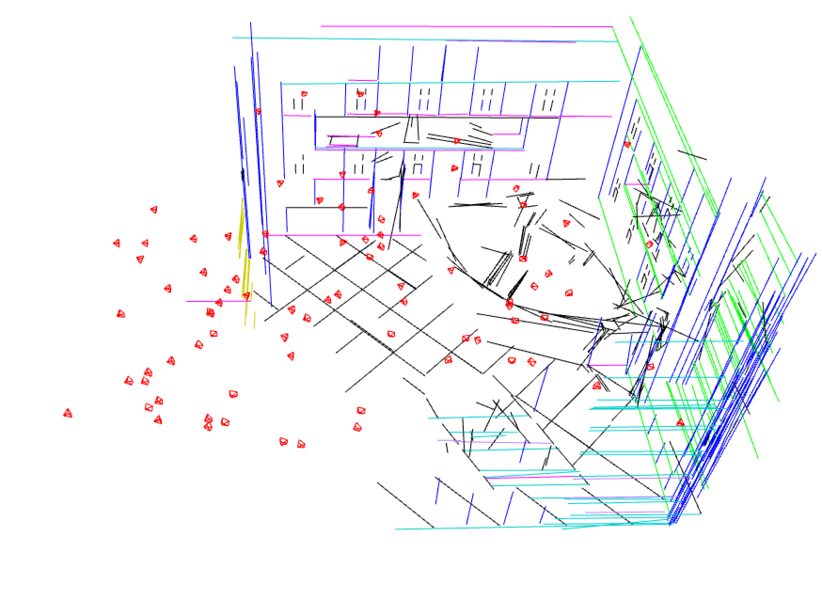}} &
{\includegraphics[width=0.24\linewidth, height=48pt]{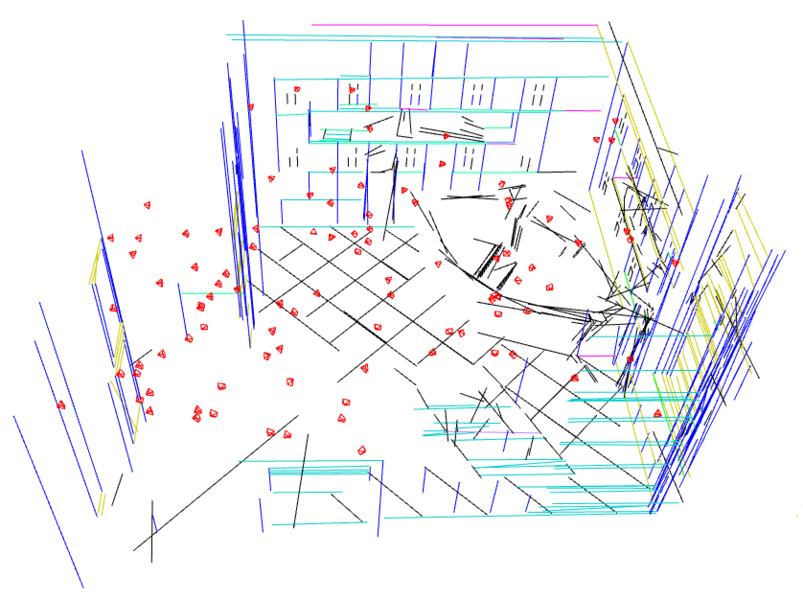}} \\
20\% & 50\% & 70\% & 100\% \\
\end{tabular}
\centering
\caption{\textbf{Incremental line reconstruction during SfM.} Parallel lines from line-VP association are colored the same. }
\label{fig::dynamic}
\end{figure}

\begin{figure}[tb]
\scriptsize
\setlength\tabcolsep{1pt} % Adjust the space between columns as needed
\begin{tabular}{cccc}
\begin{tabular}[b]{c}
\includegraphics[width=0.08\linewidth, height=0.06\linewidth]{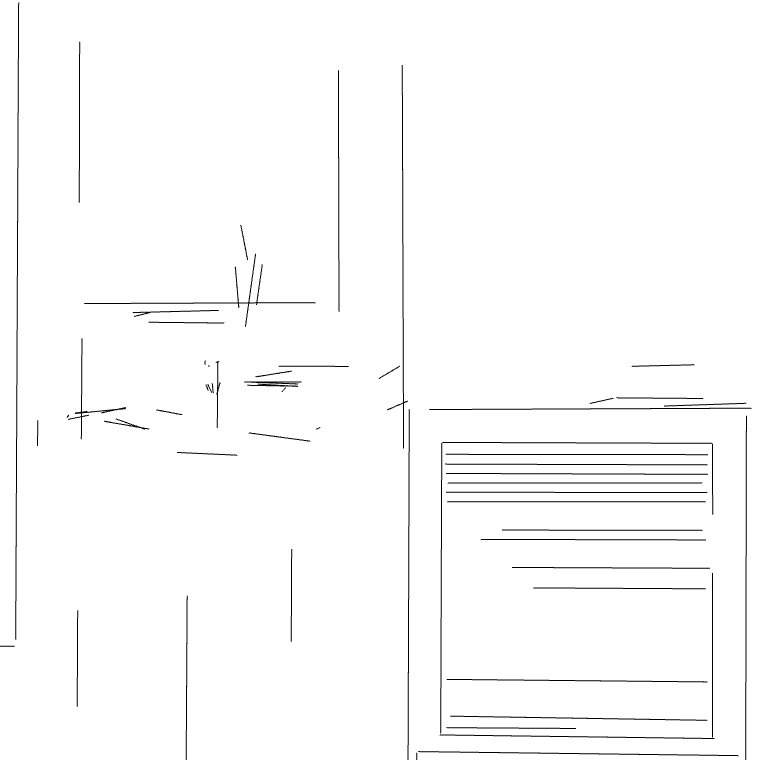} \\
\includegraphics[width=0.08\linewidth, height=0.06\linewidth]{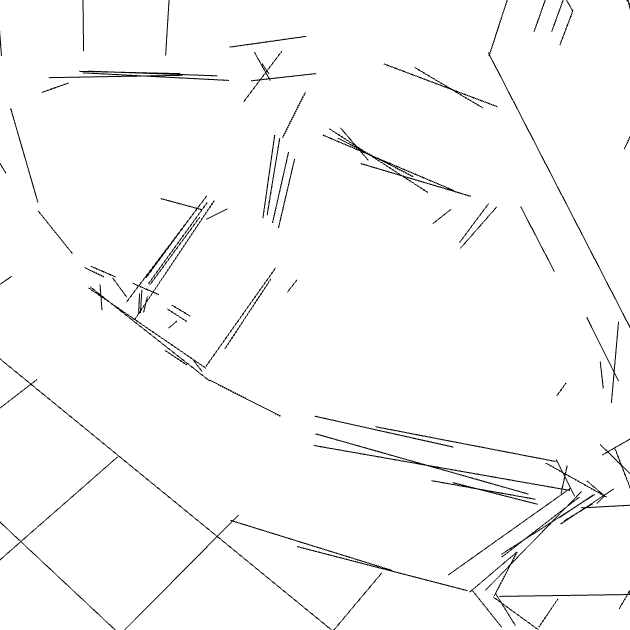} \\
\includegraphics[width=0.08\linewidth, height=0.06\linewidth]{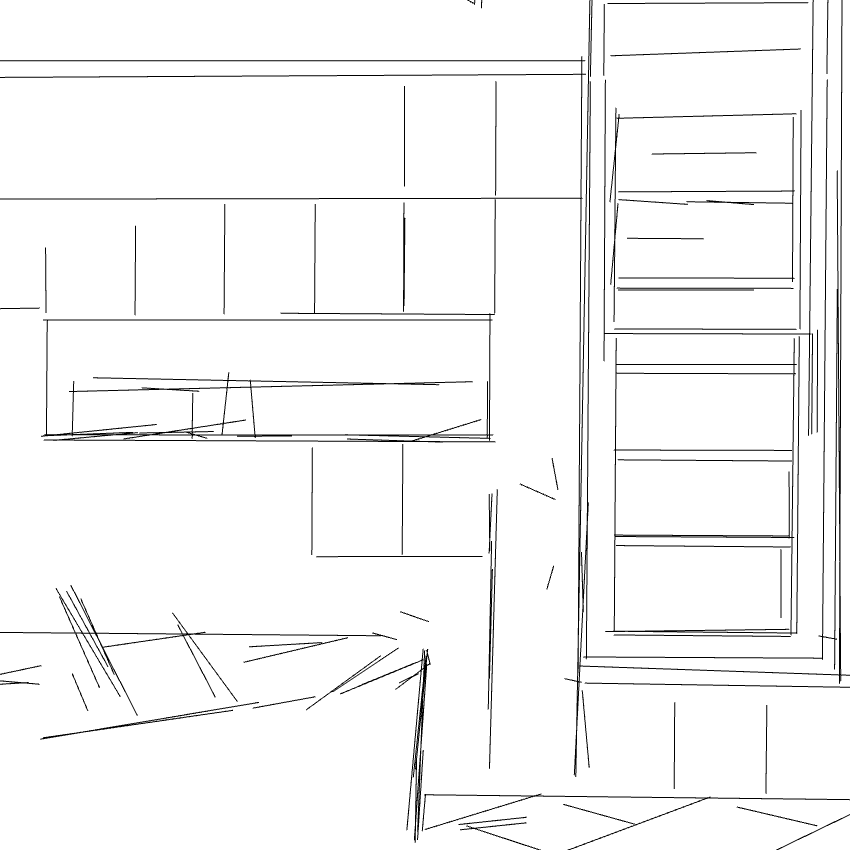} \\
\end{tabular}
&
\raisebox{-0.01\height}{\includegraphics[width=0.33\linewidth, height=0.22\linewidth]{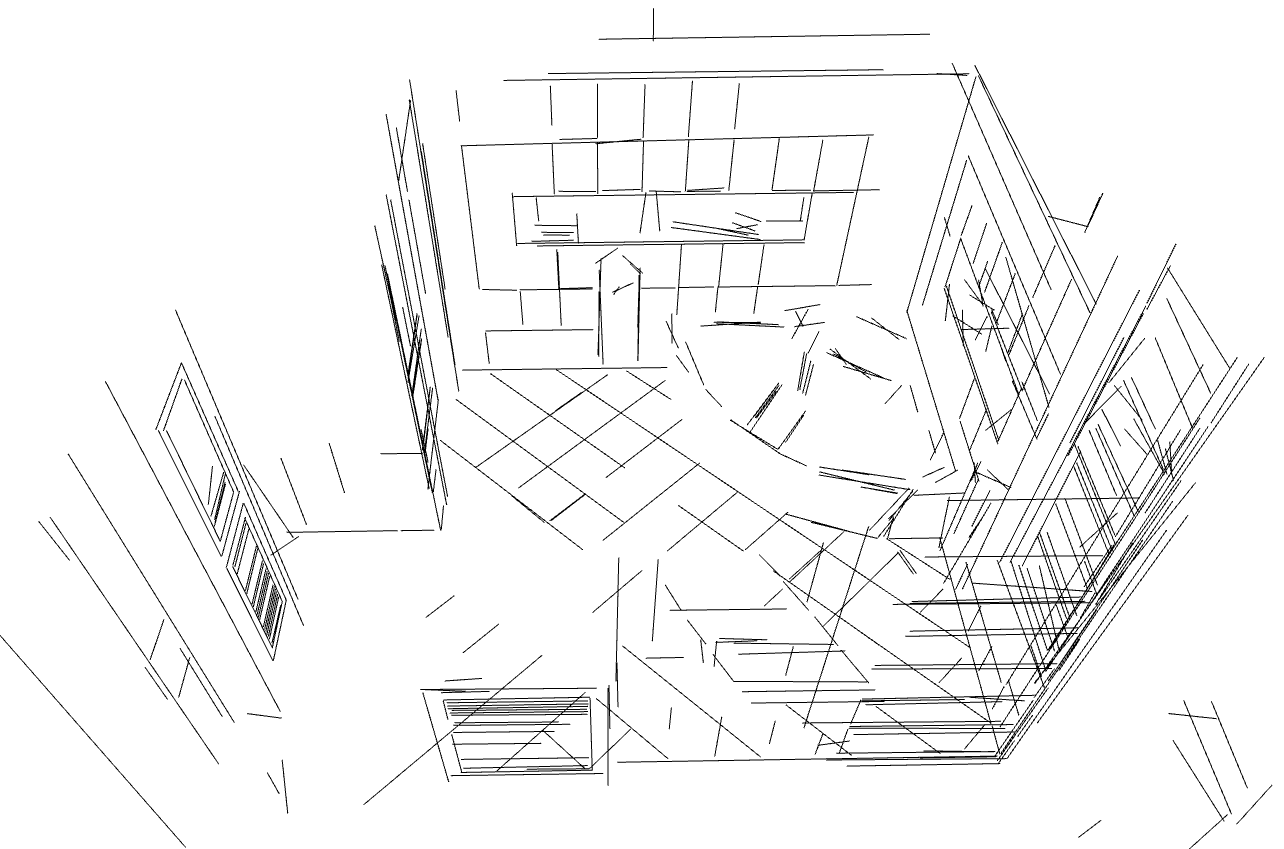}}
&
\hspace{6pt}
\begin{tabular}[b]{c}
\includegraphics[width=0.08\linewidth, height=0.06\linewidth]{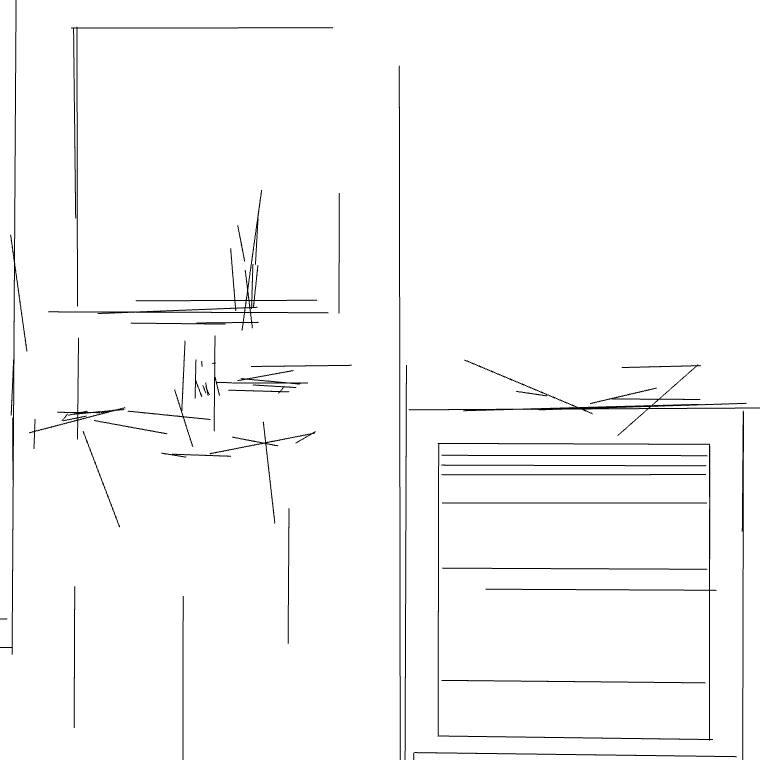} \\
\includegraphics[width=0.08\linewidth, height=0.06\linewidth]{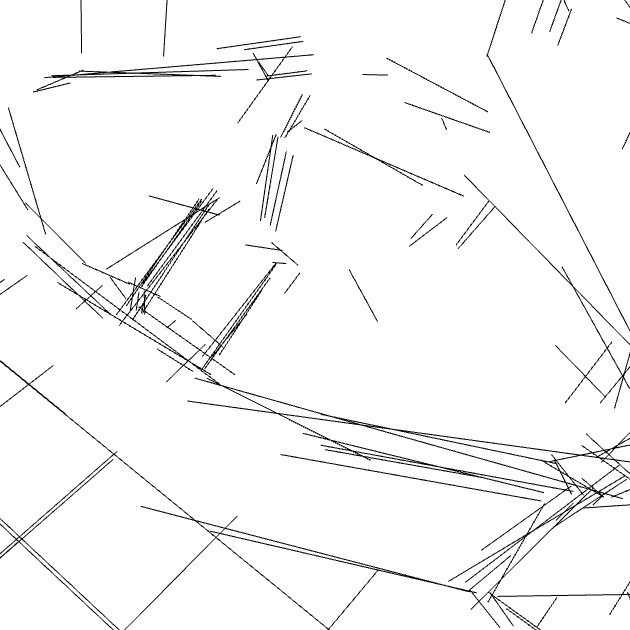} \\
\includegraphics[width=0.08\linewidth, height=0.06\linewidth]{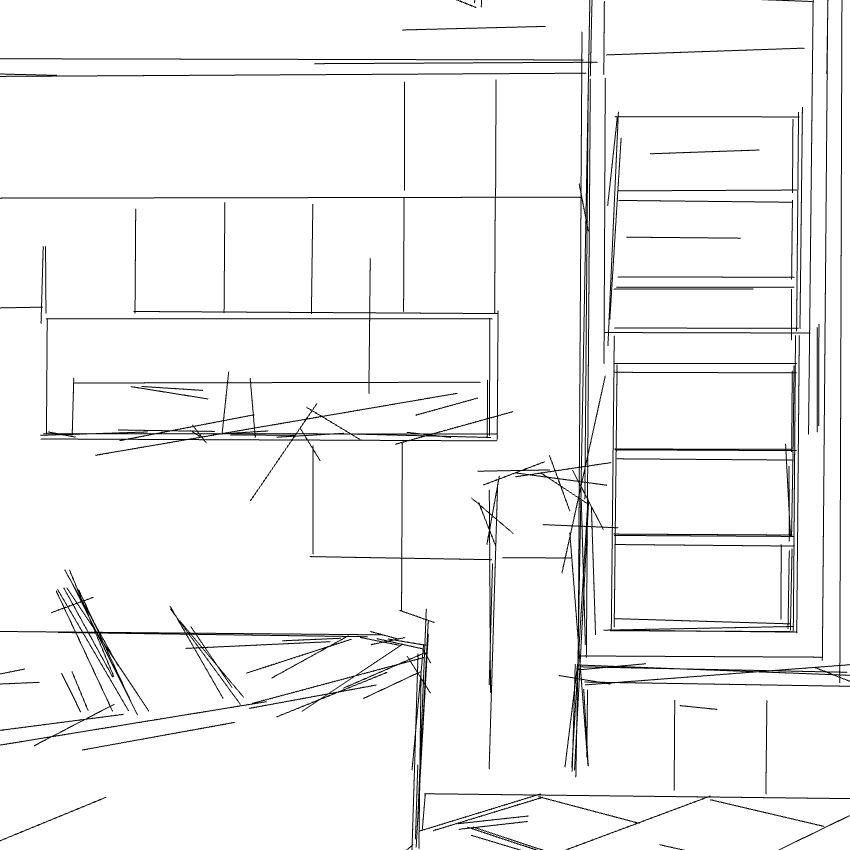} \\
\end{tabular}
&
\raisebox{-0.01\height}{\includegraphics[width=0.33\linewidth, height=0.22\linewidth]{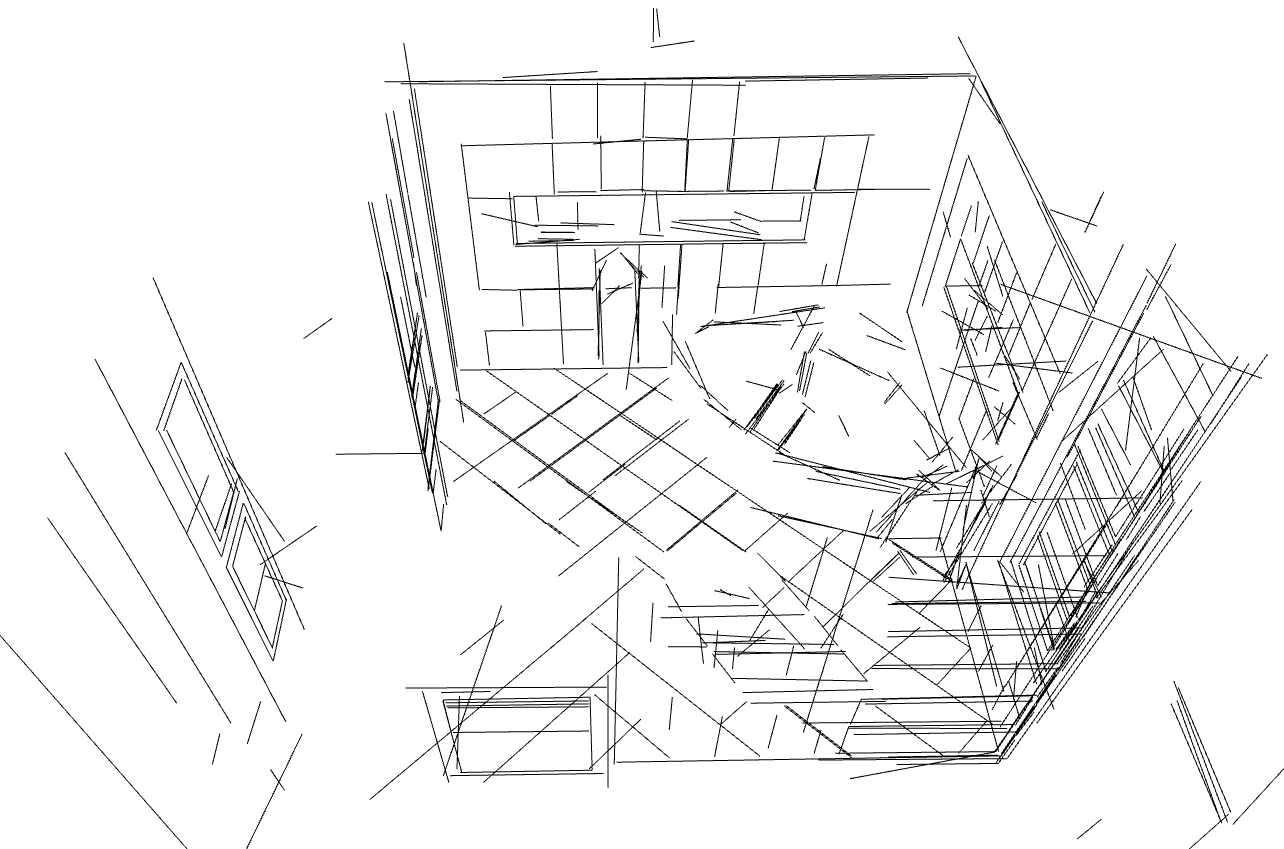}} \\
\end{tabular}
\centering
\caption{Comparison of our proposed incremental line triangulator (\textbf{Right}) with the state-of-the-art global line triangulator (\textbf{Left}) \cite{Liu_2023_LIMAP}. Without the need of getting all the posed images beforehand, our incremental triangulator achieve robust line reconstruction of comparable quality with the global counterpart. 
}
\label{fig::mapping}
\end{figure}
\begin{figure}[tb]
\scriptsize
\setlength\tabcolsep{0pt} % Adjust as needed
\begin{tabular}{ccc}
\includegraphics[width=0.27\linewidth, height=35pt]{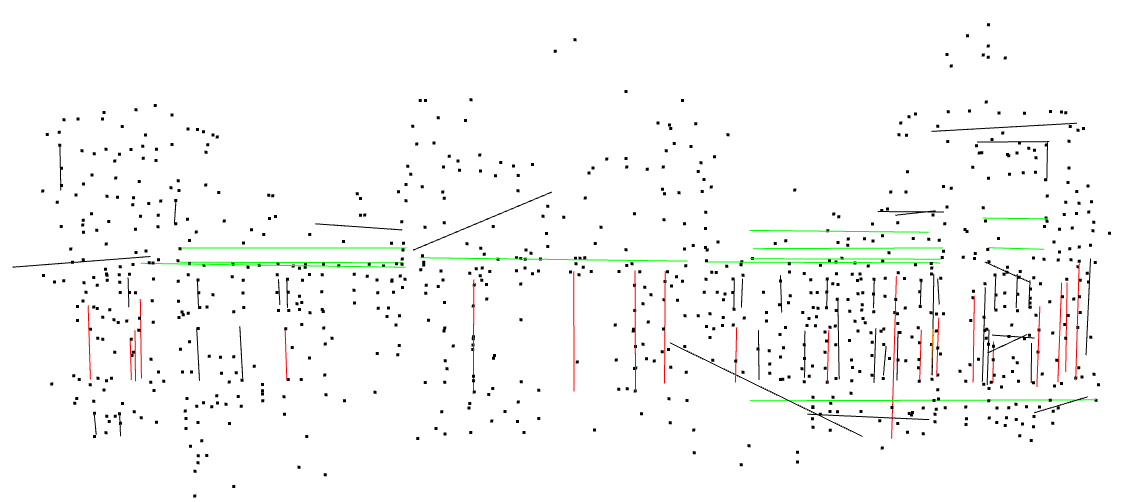} &
\includegraphics[width=0.27\linewidth, height=35pt]{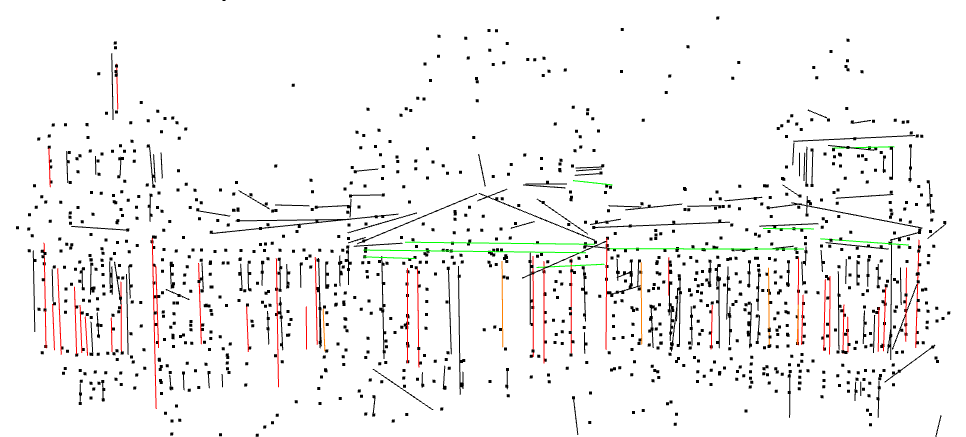} &
\includegraphics[width=0.27\linewidth, height=35pt]{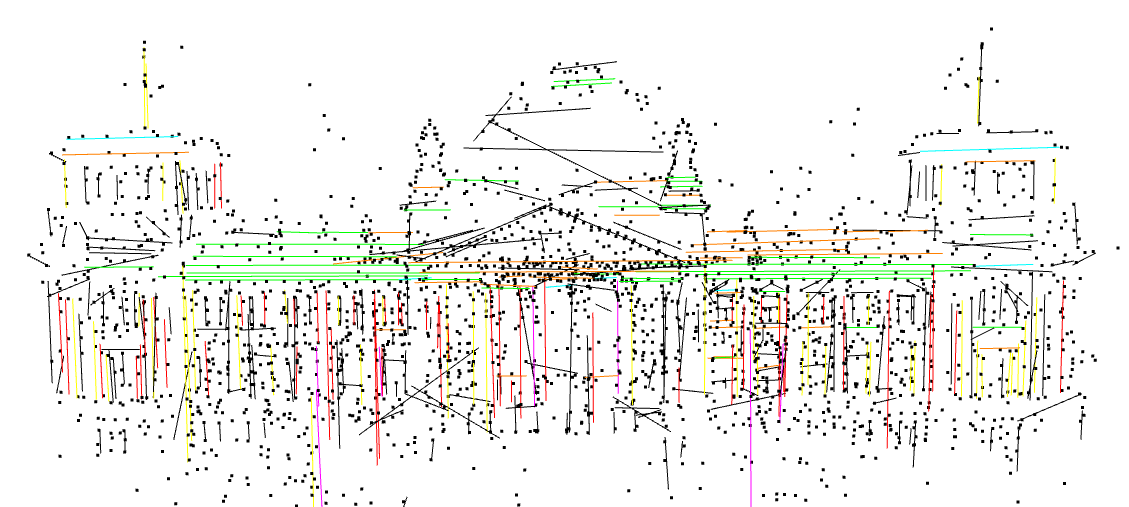} \\
% 5 Images & 10 Images & 25 Images \\
\\
\includegraphics[trim={0 20 0 45}, clip, width=0.27\linewidth, height=45pt]{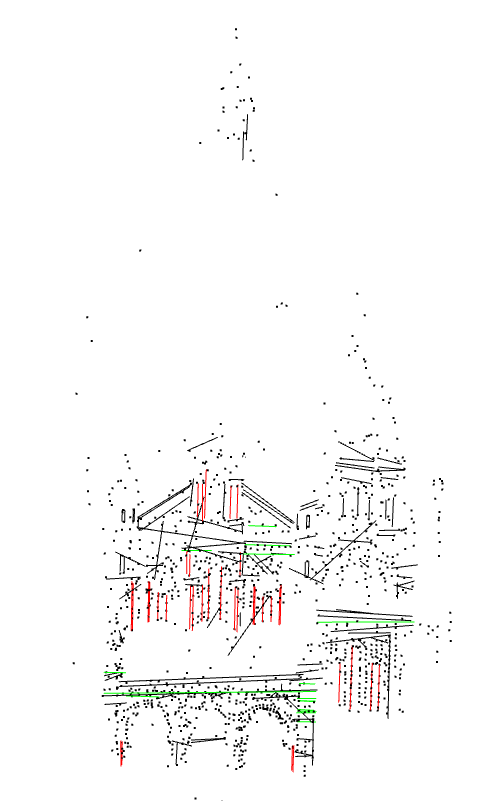} &
\includegraphics[trim={0 20 0 45}, clip, width=0.27\linewidth, height=45pt]{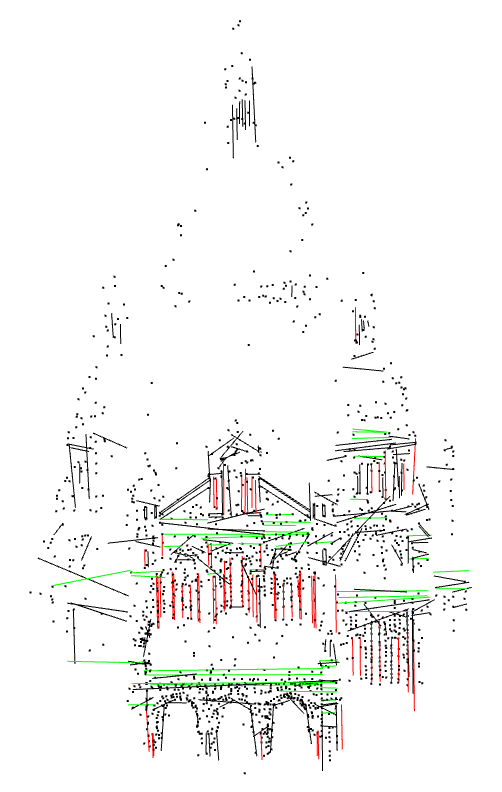} &
\includegraphics[trim={0 20 0 45}, clip, width=0.27\linewidth, height=45pt]{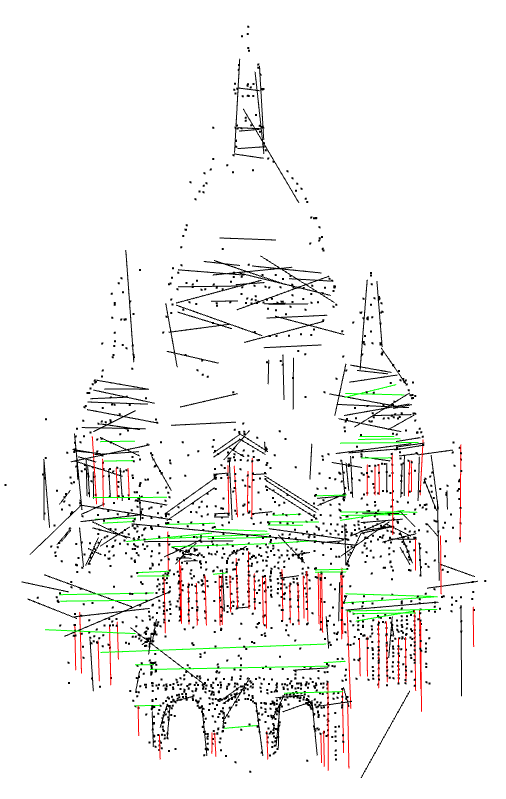} \\
% 5 Images & 10 Images & 25 Images \\
\\ 
\includegraphics[trim={0 0 0 45}, clip, width=0.27\linewidth, height=35pt]{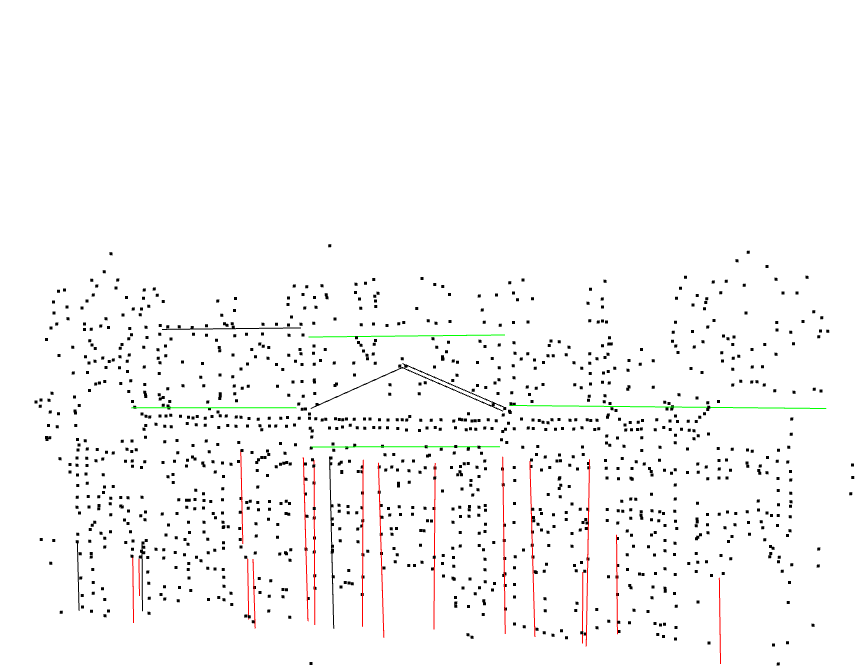} &
\includegraphics[trim={0 0 0 45}, clip, width=0.27\linewidth, height=35pt]{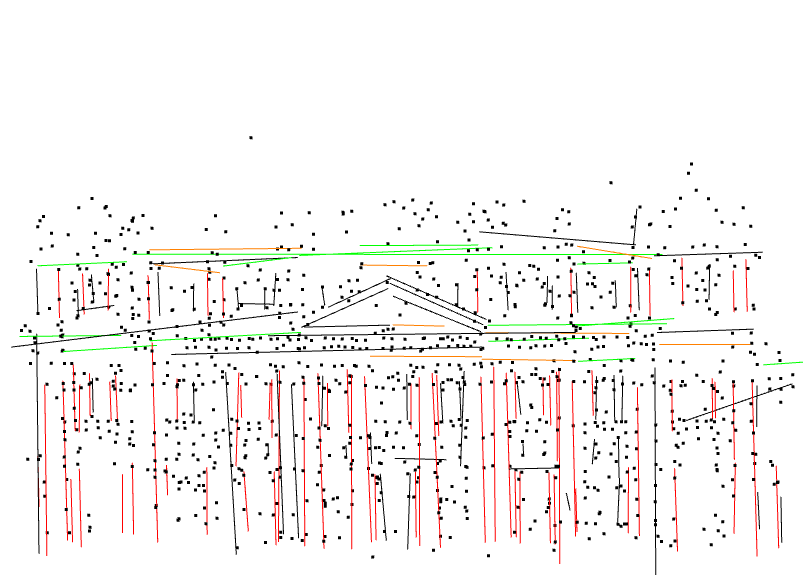} &
\includegraphics[trim={0 0 0 45}, clip, width=0.27\linewidth, height=35pt]{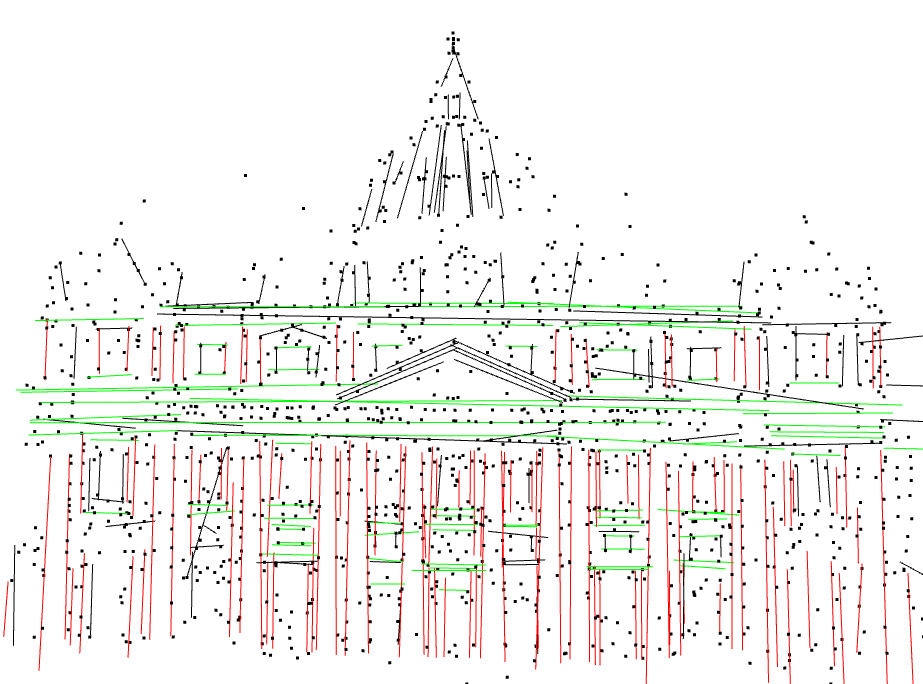} \\
5 Images & 10 Images & 25 Images \\
\end{tabular}
\centering
\caption{Our hybrid reconstruction with different small image bags from PhotoTourism \cite{snavely2006photo,jin2021image}. Our method can achieve reasonable reconstruction from as few as 5 images. }
\label{fig::photourism}
\end{figure}

We show qualitative results on the incremental process of hybrid reconstruction during SfM and the visual comparison with the global line triangulation from \cite{Liu_2023_LIMAP} in Fig. \ref{fig::dynamic} and \ref{fig::mapping}. We also show qualitative examples of our hybrid reconstruction on PhotoTourism \cite{jin2021image}. Our method is able to reconstruct reasonable 3D maps of hybrid features from as few as 5 images. 

\begin{figure}[tb]
\centering

% \setlength{\tabcolsep}{2pt} % Set the space between columns to zero

% \begin{tabular}{*{4}{c}} % Create a table with 4 centered columns
% \includegraphics[width=0.20\linewidth, height=42pt]{figs/imgs/exp_ablations/exp_zombie/10_without_zombie.png} &
% \includegraphics[width=0.20\linewidth, height=42pt]{figs/imgs/exp_ablations/exp_zombie/20_without_zombie.png} &
% \includegraphics[width=0.20\linewidth, height=42pt]{figs/imgs/exp_ablations/exp_zombie/30_without_zombie.png} &
% \includegraphics[width=0.22\linewidth, height=42pt]{figs/imgs/exp_ablations/exp_zombie/40_without_zombie.png} \\
% \includegraphics[width=0.20\linewidth, height=45pt]{figs/imgs/exp_ablations/exp_zombie/10_with_zombie.png} &
% \includegraphics[width=0.20\linewidth, height=45pt]{figs/imgs/exp_ablations/exp_zombie/20_with_zombie.png} &
% \includegraphics[width=0.20\linewidth, height=45pt]{figs/imgs/exp_ablations/exp_zombie/30_with_zombie.png} &
% \includegraphics[width=0.25\linewidth, height=45pt]{figs/imgs/exp_ablations/exp_zombie/40_with_zombie.png} \\
% \end{tabular}

\includegraphics[width=0.90\linewidth, height=130pt]{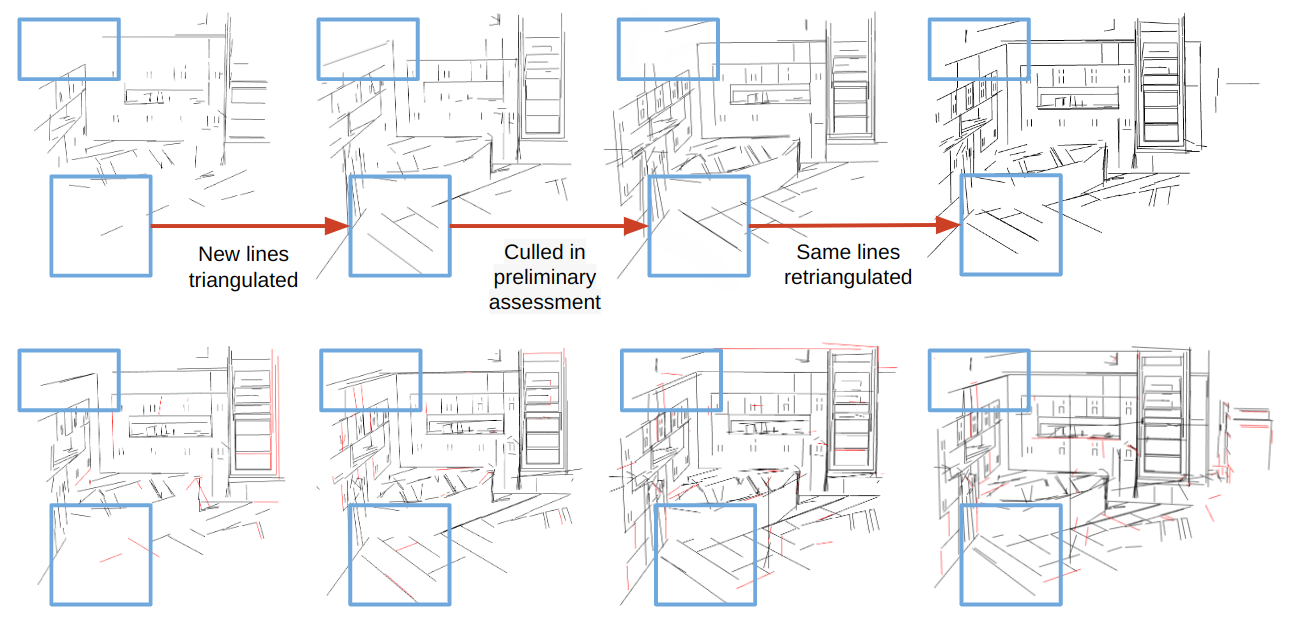}

% hard code format here
\scriptsize
\setlength{\tabcolsep}{0pt} % Set the space between columns to zero
\begin{tabular}{p{0.04\linewidth}p{0.21\linewidth}p{0.21\linewidth}p{0.21\linewidth}p{0.20\linewidth}p{0.06\linewidth}}
&
\#registered = 10 &
\#registered = 20 &
\#registered = 30 &
\#registered = 40 & \\
\end{tabular}

\caption{\textbf{Effects of caching inactive supports and two-step refinement.} With the naive strategy (first row), lines are triangulated and removed repeatedly (as highlighted in the {\color{blue} blue} boxes), which largely slows down the reconstruction process. Our proposed mechanism (second row) keeps the unreliable supports and tracks ({\color{red} red}) in the map while isolating them from the pose optimization.
% \todo{caption tba. caching inactivte supports }Incremental line reconstructions during hybrid SfM, when 10, 20, 30, 40 images are registered respectively. First row: Obtained without caching inactive supports, some lines are removed and added again. Second row: lines in red are the ones that would have been filtered without maintaining unreliable tracks.
}
\label{fig::exp_zombie}
\end{figure}
Figure \ref{fig::exp_zombie} shows an example on the effects of caching inactive supports and two-step refinement. The 3D lines are triangulated and removed repeatedly with the naive strategy, while the proposed mechanism keeps the temporarily unreliable supports and tracks while isolating them from the pose optimization.

\begin{figure}[tb]
\scriptsize
\setlength\tabcolsep{1pt}
\begin{tabular}{ccc}
{\includegraphics[trim={0 100 620 100}, clip, width=0.32\linewidth, height=50pt]{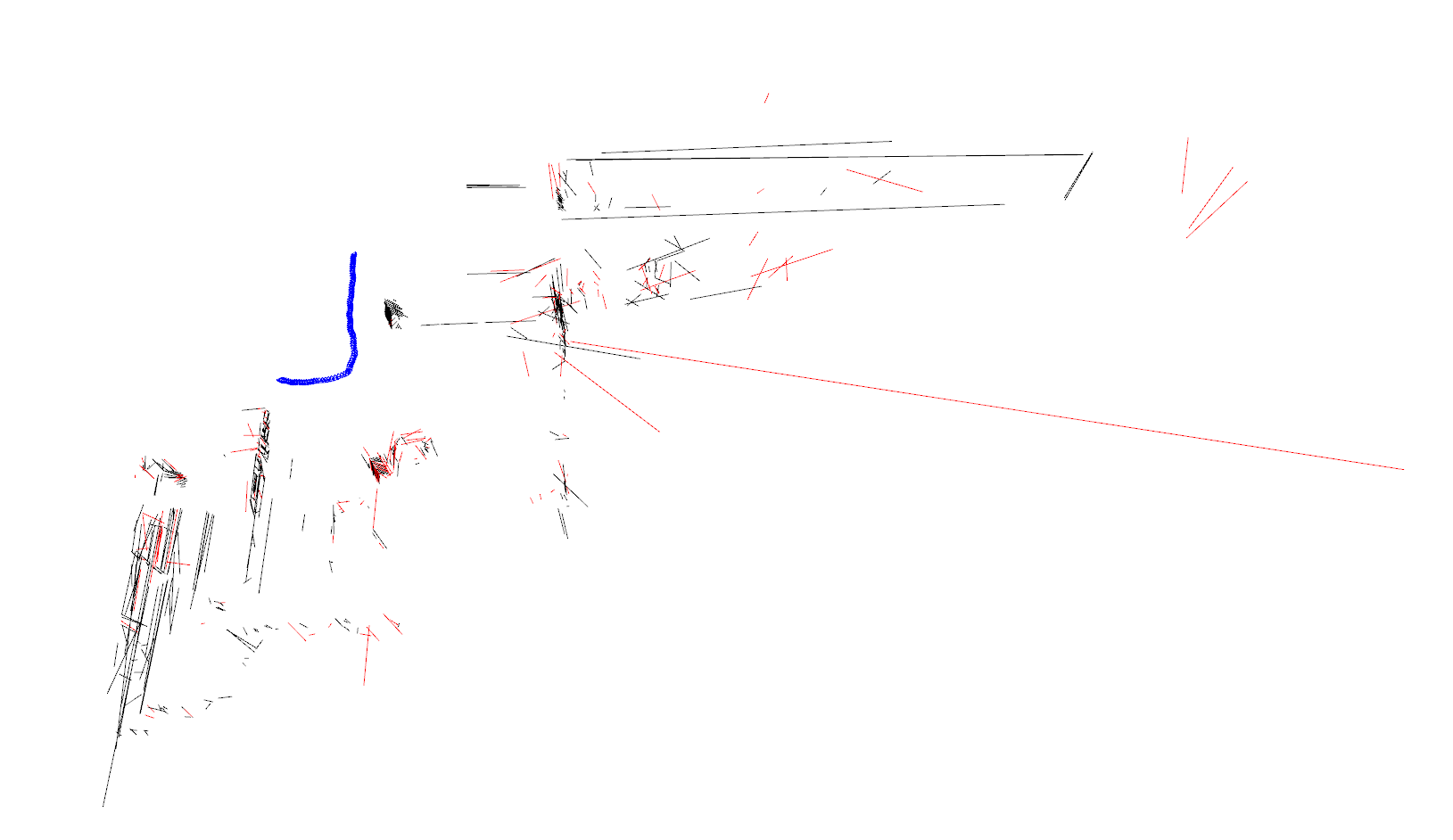}} &
{\includegraphics[trim={160 100 0 220}, clip, width=0.32\linewidth, height=50pt]{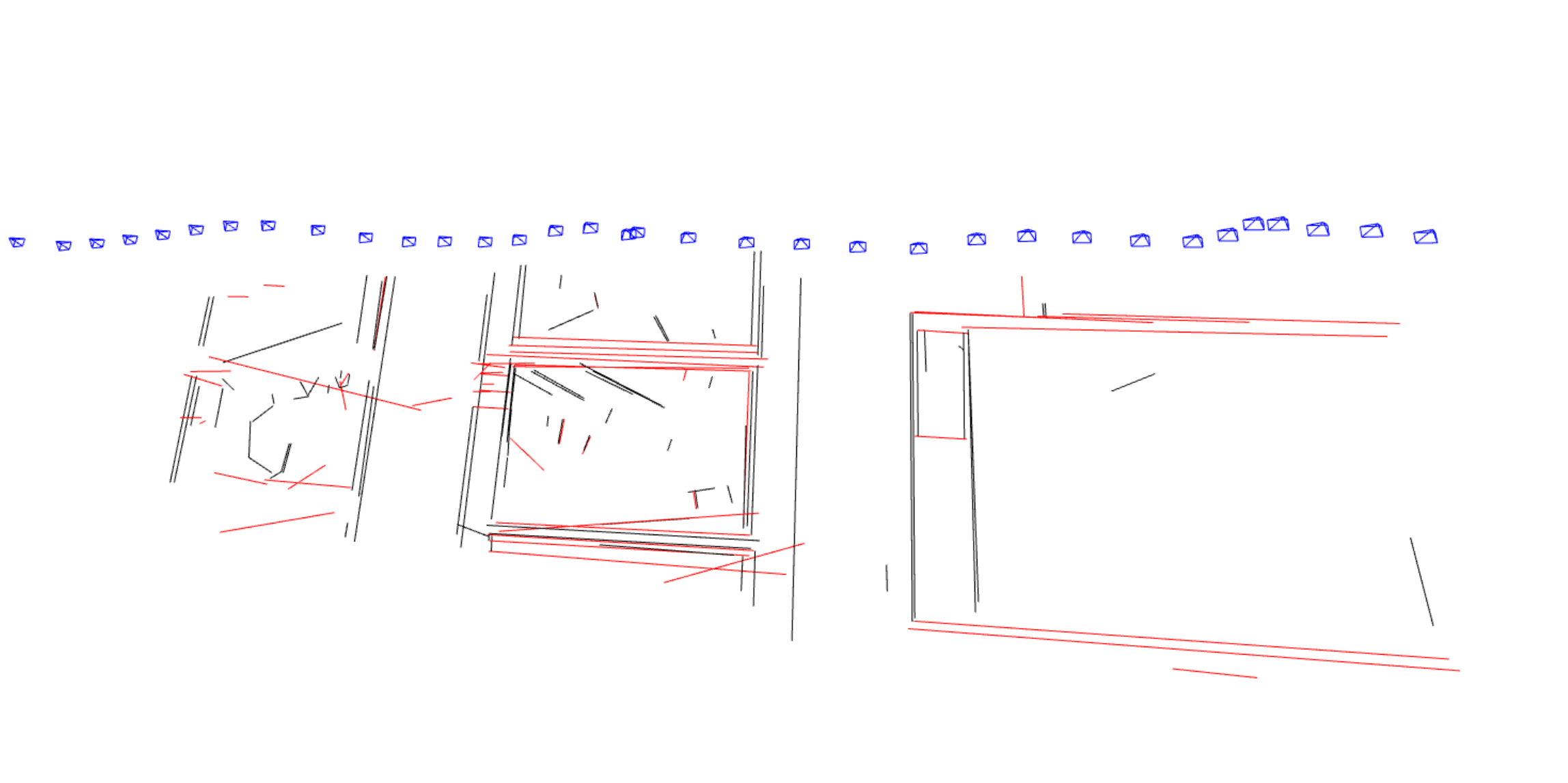}} &
{\includegraphics[trim={0 0 0 130}, clip, width=0.32\linewidth, height=50pt]{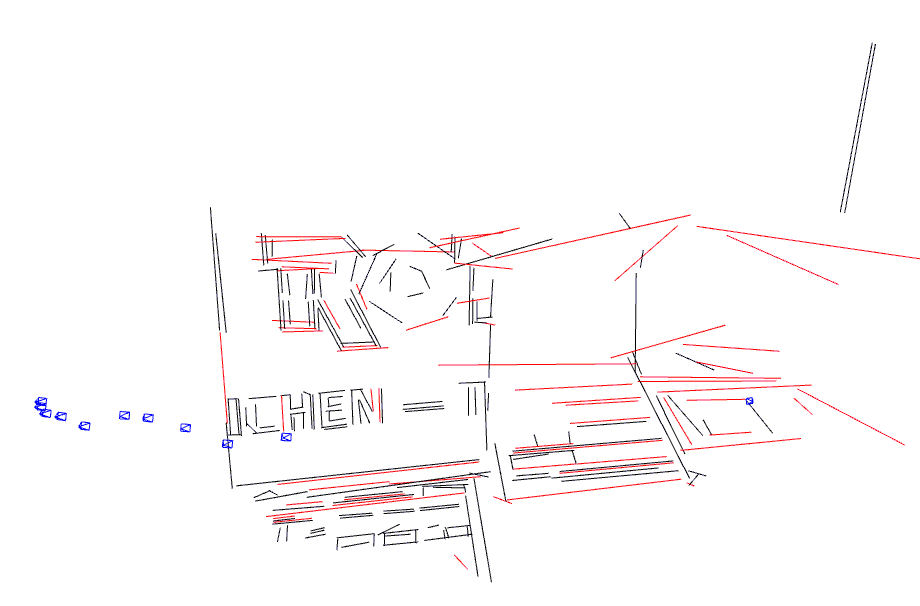}} \\
\end{tabular}
\centering
\caption{\textbf{Visualization of the scale-invariant reliability measurement based on the uncertainty propagation during SfM.} Not only can our method identify noisy lines (red) flying around space, but it can also model unstable lines from degenerate view configurations which are not necessarily short of supporting views.}
\label{fig::global_covariance}
\end{figure}

% stairs
\begin{figure}[tb]
\centering
\scriptsize
\setlength\tabcolsep{1pt}

\begin{tabular}{ccc}
% left bottom right top
{\includegraphics[trim={250, 150, 250, 50}, clip, width=0.33\linewidth]{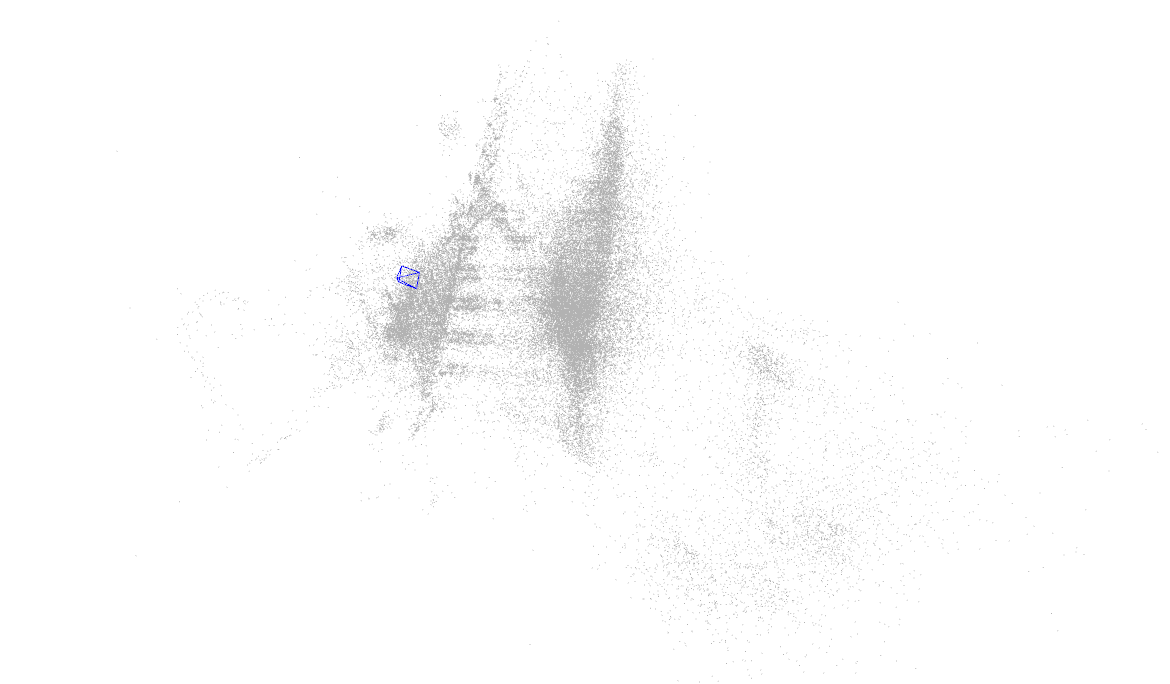}}
 &
 {\includegraphics[trim={250, 150, 250, 50}, clip, width=0.33\linewidth]{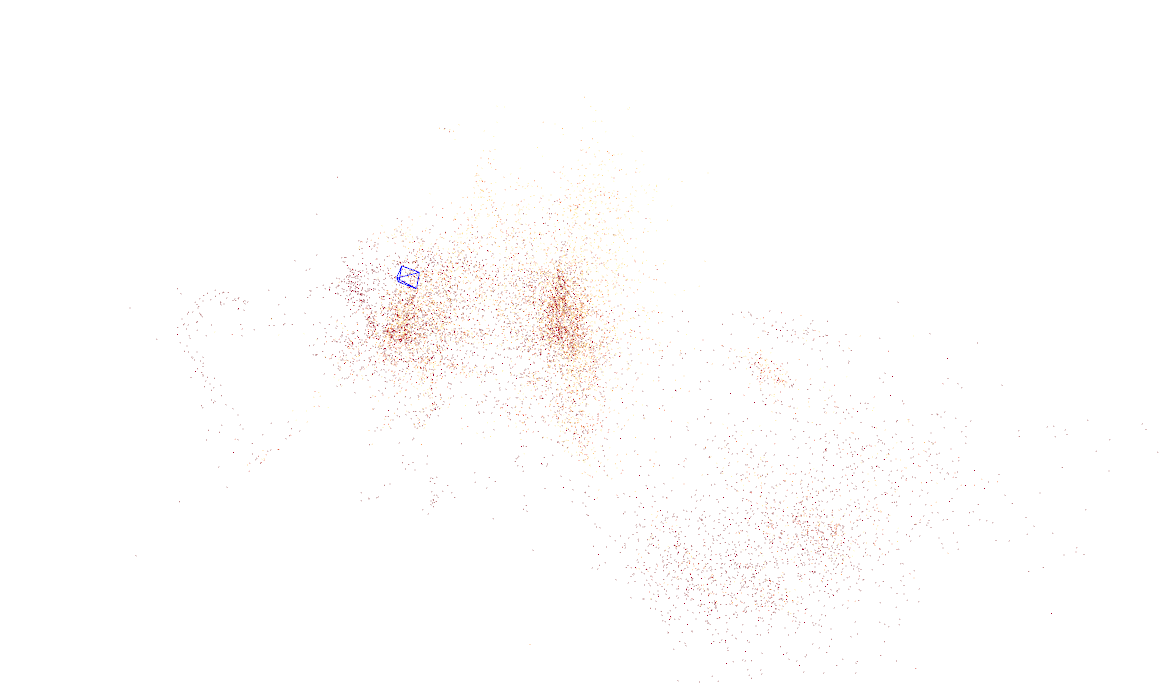}}
 &
{\includegraphics[trim={250, 150, 250, 50}, clip, width=0.33\linewidth]{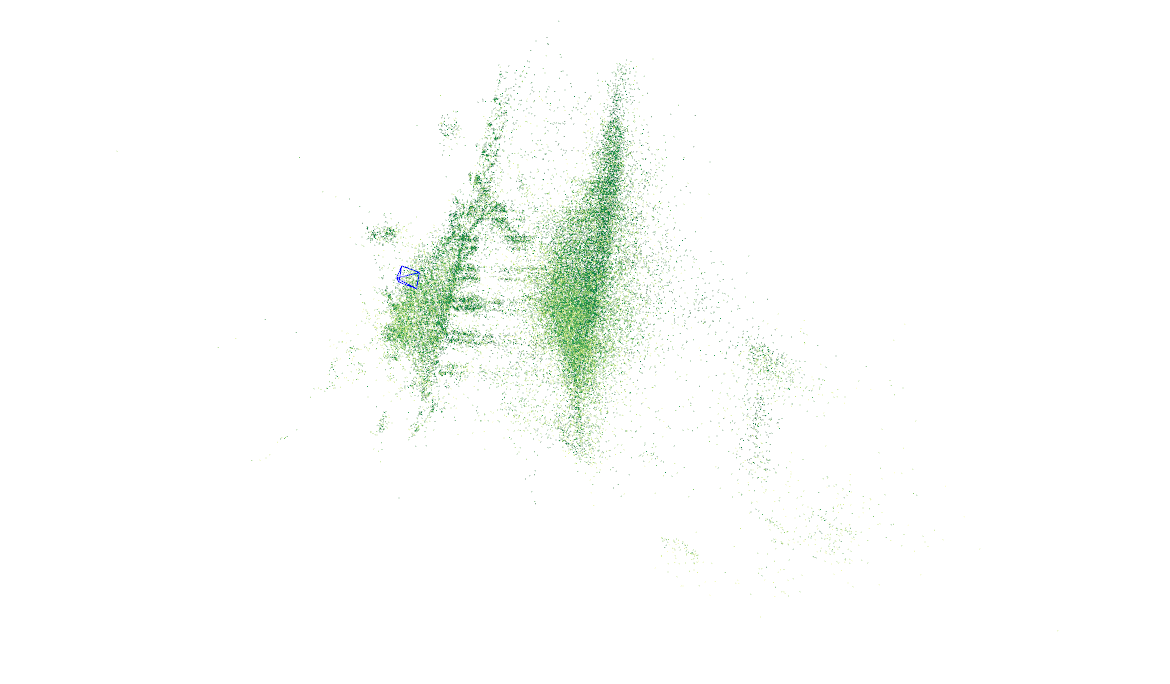}}
\\
{\includegraphics[trim={250, 150, 250, 50}, clip, width=0.33\linewidth]{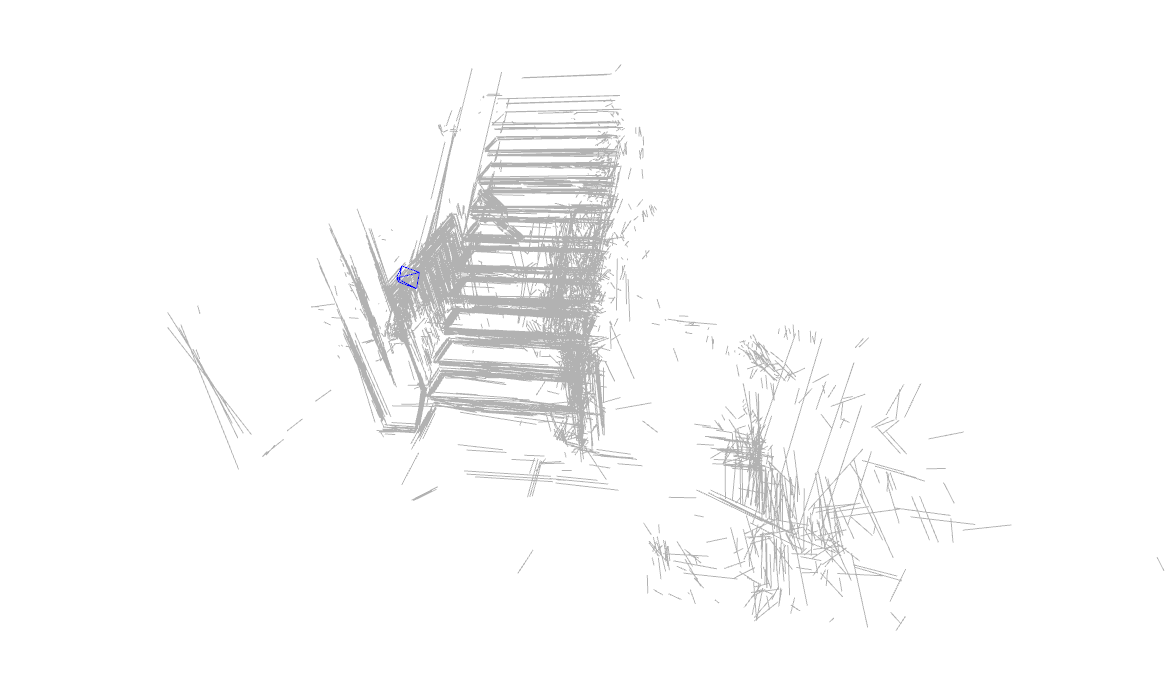}} 
&
{\includegraphics[trim={250, 150, 250, 50}, clip, width=0.33\linewidth]{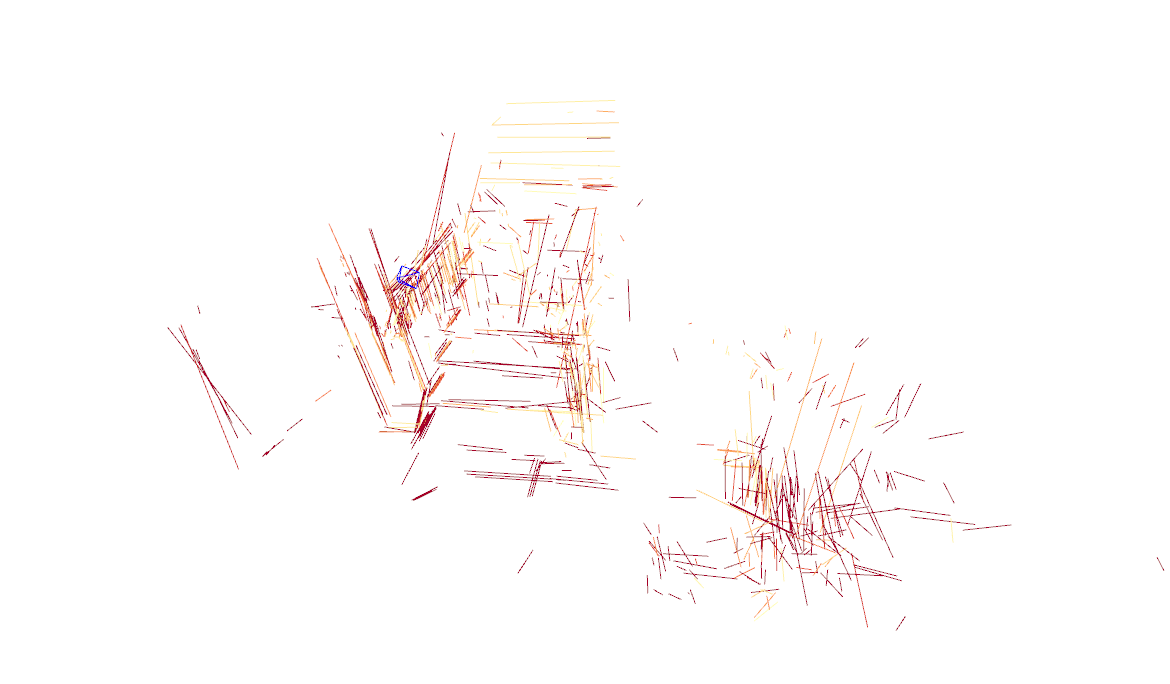}} &
{\includegraphics[trim={250, 150, 250, 50}, clip, width=0.33\linewidth]{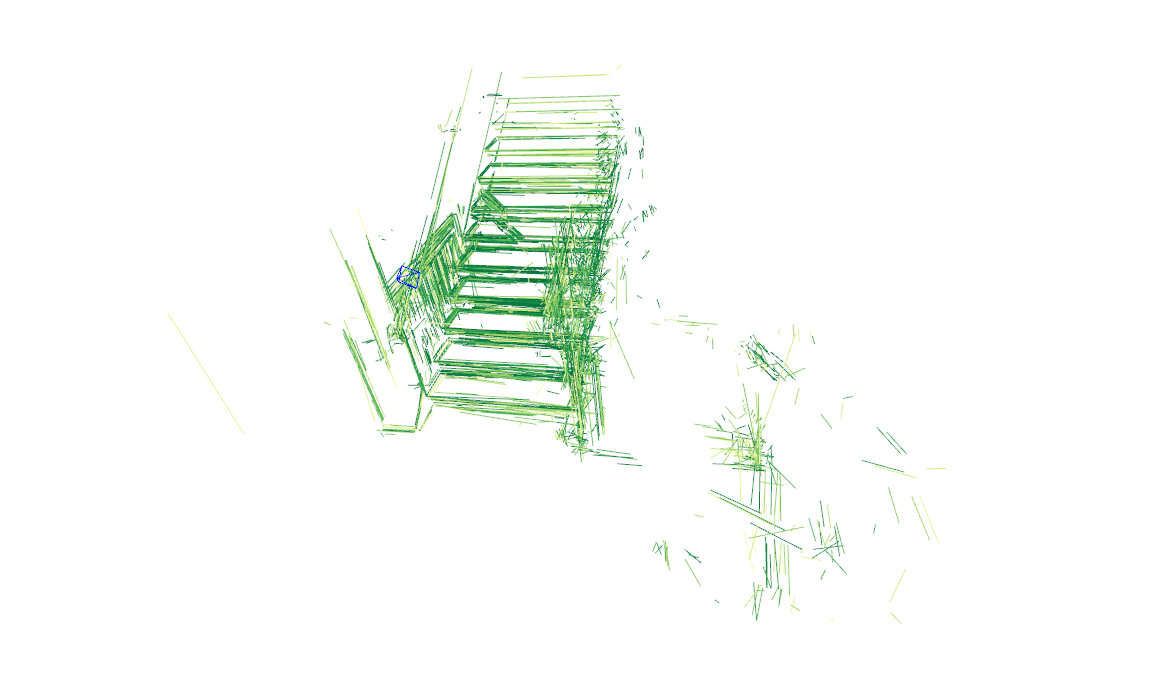}} 
\\
Full map & Filtered map w/ uncertainty & Survived map
\end{tabular}
\centering
\caption{Visualization on the reprojection uncertainty used in the uncertainty-aware localization on \textit{stairs} from 7Scenes \cite{7scenes}. We show results on both points (\textbf{first row}) and lines (\textbf{second row}).}
\label{fig::supp_localization_covariance}
\end{figure}

We further show visual examples of our propagated 3D covariance on the reconstructed 3D lines in Fig. \ref{fig::global_covariance}. With the principled uncertainty propagation our method can explicitly identify noisy lines from few views and degenerate configurations (horizontal lines cannot be reliably reconstructed with parallel horizontal motion). We also show another visual example of the effectiveness of the rejected uncertainty in Fig. \ref{fig::supp_localization_covariance}. 

\begin{figure}[tb]
\scriptsize
\begin{tabular}{cc}
{\includegraphics[width=0.40\linewidth, height=40pt]{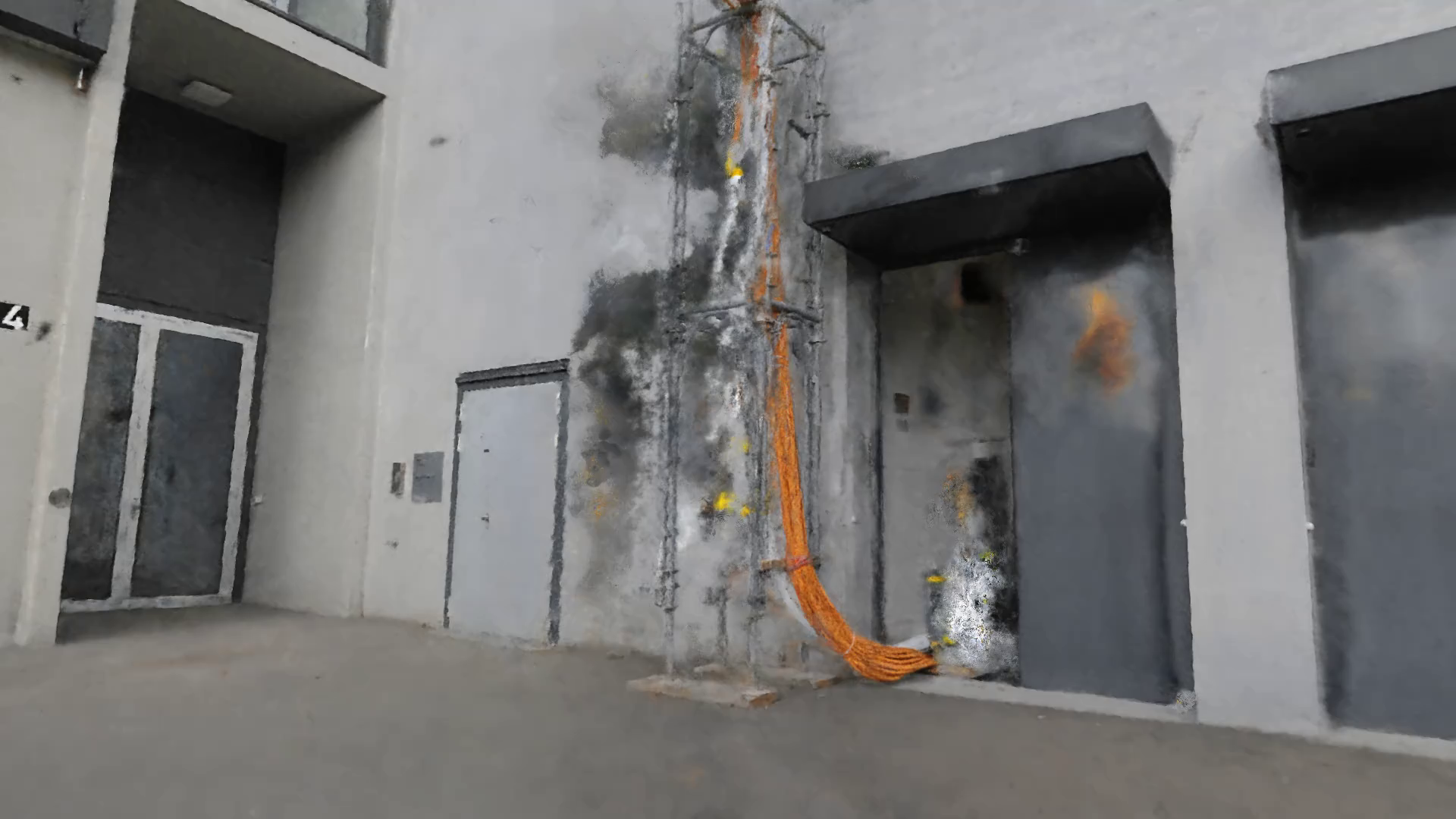}} &
{\includegraphics[width=0.40\linewidth, height=40pt]{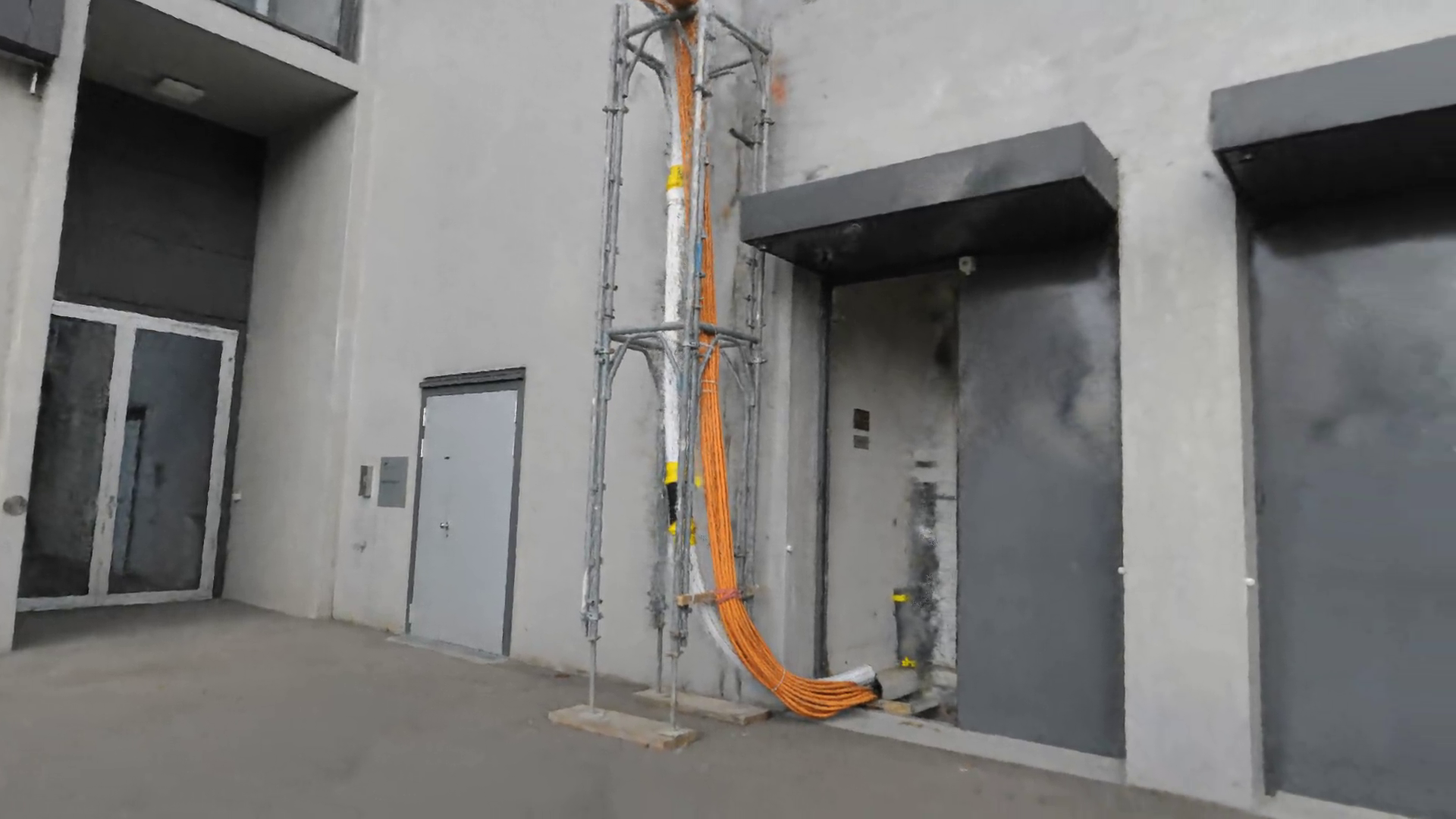}} \\ Nerfacto \cite{nerfstudio} w. COLMAP \cite{schonberger2016structure} & Nerfacto \cite{nerfstudio} w. Ours
\end{tabular}
\centering
\caption{With better accuracy and more valid registrations, our pipeline is able to improve robustness on neural rendering \cite{nerf,nerfstudio}.}
\label{fig::nerf_experiments}
\end{figure}

To further verify the improved robustness of our proposed system, we run both COLMAP \cite{schonberger2016structure} and our hybrid SfM method on the recently introduced LaMAR dataset \cite{sarlin2022lamar}. LaMAR \cite{sarlin2022lamar} is a new large-scaled dataset that is captured using AR devices in diverse environments. We use the \textit{hetrf} sensor from the HoloLens query validation data which consists of 12 sequences. Table \ref{tab::lamar_alone} shows results with SuperPoint \cite{detone2018superpoint} + SuperGlue \cite{sarlin2020superglue}. While the dataset is very challenging, we manage to get consistent improvement over COLMAP \cite{schonberger2016structure}, thanks to the inclusion of hybrid features. Nonetheless, there is still large room for future improvement on this dataset, which can further benefit from temporal modeling. This is beyond the scope of this paper and is left as the next-step future work. 

\begin{table}[tb]
\begin{center}
\scriptsize
\setlength{\tabcolsep}{10pt}
\caption{\textbf{Structure-from-Motion results on LaMAR \cite{sarlin2022lamar}}. We report the relative pose AUC for both our system (``Hybrid") and COLMAP (``Point") \cite{schonberger2016structure} on SuperPoint \cite{detone2018superpoint} + SuperGlue \cite{sarlin2020superglue}. While we achieve consistent improvement over COLMAP \cite{schonberger2016structure}, the dataset is very challenging due to low overlap and fast motion. }
\begin{tabular}{cccccccccc}
\toprule
\multirow{2}{*}{Dataset} & \multirow{2}{*}{Point Feature} & \multirow{2}{*}{Method} & \multicolumn{3}{c}{\multirow{2}{*}{AUC @ 3°/5°/10° $\uparrow$}} \\
& & & & & \\
\midrule
\multirow{2}{*}{LaMAR}
& \multirow{2}{*}{SP + SG} & Point & 2.8 & 7.9 & 14.5\\
& & Hybrid & \textbf{4.0} & \textbf{10.8} & \textbf{18.5} \\
\bottomrule
\end{tabular}
\label{tab::lamar_alone}
\end{center}
\end{table}

Finally, we apply our proposed SfM system to the widely studied application: view synthesis. Specifically, we run Nerfactos from NeRFStudio \cite{nerf, nerfstudio} on both COLMAP \cite{schonberger2016structure} and our SfM predictions. Fig. \ref{fig::nerf_experiments} shows that our method enables better view synthesis quality due to more accurate and robust camera registrations.

\clearpage
\bibliographystyle{splncs04}
\bibliography{egbib}

\begin{thebibliography}{100}
\providecommand{\url}[1]{\texttt{#1}}
\providecommand{\urlprefix}{URL }
\providecommand{\doi}[1]{https://doi.org/#1}

\bibitem{abdellali2021l2d2}
Abdellali, H., Frohlich, R., Vilagos, V., Kato, Z.: L2d2: Learnable line
  detector and descriptor. In: International Conference on 3D Vision (3DV)
  (2021)

\bibitem{agarwal2011building}
Agarwal, S., Furukawa, Y., Snavely, N., Simon, I., Curless, B., Seitz, S.M.,
  Szeliski, R.: Building rome in a day. Communications of the ACM
  \textbf{54}(10),  105--112 (2011)

\bibitem{ceres}
Agarwal, S., Mierle, K.: Ceres solver. \url{http://ceres-solver.org}

\bibitem{agarwal2010bundle}
Agarwal, S., Snavely, N., Seitz, S.M., Szeliski, R.: Bundle adjustment in the
  large. In: ECCV (2010)

\bibitem{arandjelovic2016netvlad}
Arandjelovic, R., Gronat, P., Torii, A., Pajdla, T., Sivic, J.: Netvlad: Cnn
  architecture for weakly supervised place recognition. In: CVPR (2016)

\bibitem{aastrom1998statistical}
{\AA}str{\"o}m, K., Kahl, F., Heyden, A., Berthilsson, R.: A statistical
  approach to structure and motion from image features. In: Advances in Pattern
  Recognition: Joint IAPR International Workshops SSPR'98 and SPR'98 (1998)

\bibitem{bartoli2004framework}
Bartoli, A., Coquerelle, M., Sturm, P.: A framework for pencil-of-points
  structure-from-motion. In: ECCV (2004)

\bibitem{bartoli2005structure}
Bartoli, A., Sturm, P.: Structure-from-motion using lines: Representation,
  triangulation, and bundle adjustment. Computer Vision and Image Understanding
  (CVIU)  \textbf{100}(3),  416--441 (2005)

\bibitem{bazin_cvpr_2012}
Bazin, J.C., Seo, Y., Demonceaux, C., Vasseur, P., Ikeuchi, K., Kweon, I.,
  Pollefeys, M.: Globally optimal line clustering and vanishing point
  estimation in manhattan world. In: CVPR (2012)

\bibitem{bhowmick2015divide}
Bhowmick, B., Patra, S., Chatterjee, A., Govindu, V.M., Banerjee, S.: Divide
  and conquer: Efficient large-scale structure from motion using graph
  partitioning. In: ACCV (2015)

\bibitem{brooks2001value}
Brooks, M.J., Chojnacki, W., Gawley, D., Van Den~Hengel, A.: What value
  covariance information in estimating vision parameters? In: ICCV (2001)

\bibitem{bui2024representing}
Bui, B.T., Bui, H.H., Tran, D.T., Lee, J.H.: Representing 3d sparse map points
  and lines for camera relocalization. arXiv preprint arXiv:2402.18011  (2024)

\bibitem{burnett2021radar}
Burnett, K., Yoon, D.J., Schoellig, A.P., Barfoot, T.D.: Radar odometry
  combining probabilistic estimation and unsupervised feature learning. In:
  Robotics: Science and Systems (RSS) (2021)

\bibitem{camposeco2018hybrid}
Camposeco, F., Cohen, A., Pollefeys, M., Sattler, T.: Hybrid camera pose
  estimation. In: CVPR (2018)

\bibitem{chandraker2009moving}
Chandraker, M., Lim, J., Kriegman, D.: Moving in stereo: Efficient structure
  and motion using lines. In: ICCV (2009)

\bibitem{chum2003locally}
Chum, O., Matas, J., Kittler, J.: Locally optimized ransac. In: Joint Pattern
  Recognition Symposium (2003)

\bibitem{crandall2011discrete}
Crandall, D., Owens, A., Snavely, N., Huttenlocher, D.: Discrete-continuous
  optimization for large-scale structure from motion. In: CVPR (2011)

\bibitem{factor_graphs_for_robot_perception}
Dellaert, F., Kaess, M.: Factor Graphs for Robot Perception. Foundations and
  Trends in Robotics, Vol. 6 (2017),
  \url{http://www.cs.cmu.edu/~kaess/pub/Dellaert17fnt.pdf}

\bibitem{dellaert2000structure}
Dellaert, F., Seitz, S.M., Thorpe, C.E., Thrun, S.: Structure from motion
  without correspondence. In: CVPR (2000)

\bibitem{detone2018superpoint}
DeTone, D., Malisiewicz, T., Rabinovich, A.: Superpoint: Self-supervised
  interest point detection and description. In: Computer Vision and Pattern
  Recognition Workshops (CVPRW) (2018)

\bibitem{dusmanu2021cross}
Dusmanu, M., Miksik, O., Sch\"{o}nberger, J.L., Pollefeys, M.:
  {Cross-Descriptor Visual Localization and Mapping}. In: ICCV (2021)

\bibitem{dusmanu2020multiview}
Dusmanu, M., Sch\"{o}nberger, J.L., Pollefeys, M.: {Multi-View Optimization of
  Local Feature Geometry}. In: ECCV (2020)

\bibitem{fiacco1990sensitivity}
Fiacco, A.V., Ishizuka, Y.: Sensitivity and stability analysis for nonlinear
  programming. Annals of Operations Research  \textbf{27}(1),  215--235 (1990)

\bibitem{forstner1987fast}
F{\"o}rstner, W., G{\"u}lch, E.: A fast operator for detection and precise
  location of distinct points, corners and centres of circular features. In:
  Proc. ISPRS intercommission conference on fast processing of photogrammetric
  data (1987)

\bibitem{forstner2016photogrammetric}
F{\"o}rstner, W., Wrobel, B.P.: Photogrammetric computer vision (2016)

\bibitem{frahm2010building}
Frahm, J.M., Fite-Georgel, P., Gallup, D., Johnson, T., Raguram, R., Wu, C.,
  Jen, Y.H., Dunn, E., Clipp, B., Lazebnik, S., et~al.: Building rome on a
  cloudless day. In: ECCV (2010)

\bibitem{gao2003complete}
Gao, X.S., Hou, X.R., Tang, J., Cheng, H.F.: Complete solution classification
  for the perspective-three-point problem. IEEE Trans. Pattern Analysis and
  Machine Intelligence (PAMI)  \textbf{25}(8),  930--943 (2003)

\bibitem{germain2020s2dnet}
Germain, H., Bourmaud, G., Lepetit, V.: S2dnet: Learning accurate
  correspondences for sparse-to-dense feature matching. In: ECCV (2020)

\bibitem{gomez2016robust}
Gomez-Ojeda, R., Gonzalez-Jimenez, J.: Robust stereo visual odometry through a
  probabilistic combination of points and line segments. In: ICRA (2016)

\bibitem{gomez2019pl}
Gomez-Ojeda, R., Moreno, F.A., Zuniga-No{\"e}l, D., Scaramuzza, D.,
  Gonzalez-Jimenez, J.: Pl-slam: A stereo slam system through the combination
  of points and line segments. IEEE Transactions on Robotics  \textbf{35}(3),
  734--746 (2019)

\bibitem{hartley2003multiple}
Hartley, R., Zisserman, A.: Multiple view geometry in computer vision.
  Cambridge university press (2003)

\bibitem{hartley1997triangulation}
Hartley, R.I., Sturm, P.: Triangulation. Computer Vision and Image
  Understanding (CVIU)  \textbf{68}(2),  146--157 (1997)

\bibitem{he2023detector}
He, X., Sun, J., Wang, Y., Peng, S., Huang, Q., Bao, H., Zhou, X.:
  Detector-free structure from motion. arXiv preprint arXiv:2306.15669  (2023)

\bibitem{he2018pl}
He, Y., Zhao, J., Guo, Y., He, W., Yuan, K.: Pl-vio: Tightly-coupled monocular
  visual--inertial odometry using point and line features. Sensors
  \textbf{18}(4), ~1159 (2018)

\bibitem{hofer2015line3d}
Hofer, M., Maurer, M., Bischof, H.: Line3d: Efficient 3d scene abstraction for
  the built environment. In: German Conference on Pattern Recognition (2015)

\bibitem{holynski2020reducing}
Holynski, A., Geraghty, D., Frahm, J.M., Sweeney, C., Szeliski, R.: Reducing
  drift in structure from motion using extended features. In: International
  Conference on 3D Vision (3DV) (2020)

\bibitem{huang2020tp}
Huang, S., Qin, F., Xiong, P., Ding, N., He, Y., Liu, X.: Tp-lsd: Tri-points
  based line segment detector. In: ECCV (2020)

\bibitem{pybind11}
Jakob, W., Rhinelander, J., Moldovan, D.: pybind11 -- seamless operability
  between c++11 and python. \url{https://github.com/pybind/pybind11}

\bibitem{jiang2013global}
Jiang, N., Cui, Z., Tan, P.: A global linear method for camera pose
  registration. In: ICCV (2013)

\bibitem{jin2021image}
Jin, Y., Mishkin, D., Mishchuk, A., Matas, J., Fua, P., Yi, K.M., Trulls, E.:
  Image matching across wide baselines: From paper to practice. IJCV
  \textbf{129}(2),  517--547 (2021)

\bibitem{kanatani2004geometric}
Kanatani, K.: For geometric inference from images, what kind of statistical
  model is necessary? Systems and Computers in Japan  \textbf{35}(6), ~1--9
  (2004)

\bibitem{kanazawa2003we}
Kanazawa, Y., Kanatani, K.: Do we really have to consider covariance matrices
  for image feature points? Electronics and communications in Japan (part III:
  Fundamental electronic science)  \textbf{86}(1),  1--10 (2003)

\bibitem{kendall2015posenet}
Kendall, A., Grimes, M., Cipolla, R.: {PoseNet}: A convolutional network for
  real-time {6-DoF} camera relocalization. In: ICCV (2015)

\bibitem{kerbl20233d}
Kerbl, B., Kopanas, G., Leimk{\"u}hler, T., Drettakis, G.: 3d gaussian
  splatting for real-time radiance field rendering. ACM Transactions on
  Graphics (ToG)  \textbf{42}(4),  1--14 (2023)

\bibitem{kuhn2020deepc}
Kuhn, A., Sormann, C., Rossi, M., Erdler, O., Fraundorfer, F.: Deepc-mvs: Deep
  confidence prediction for multi-view stereo reconstruction. In: International
  Conference on 3D Vision (3DV) (2020)

\bibitem{kukelova2010closed}
Kukelova, Z., Bujnak, M., Pajdla, T.: Closed-form solutions to minimal absolute
  pose problems with known vertical direction. In: ACCV (2010)

\bibitem{kushal2012visibility}
Kushal, A., Agarwal, S.: Visibility based preconditioning for bundle
  adjustment. In: CVPR (2012)

\bibitem{PoseLib}
Larsson, V.: {PoseLib - Minimal Solvers for Camera Pose Estimation}.
  \url{https://github.com/vlarsson/PoseLib},
  \url{https://github.com/vlarsson/PoseLib}

\bibitem{lebeda2012fixing}
Lebeda, K., Matas, J., Chum, O.: Fixing the locally optimized ransac--full
  experimental evaluation. In: BMVC (2012)

\bibitem{Li_2019_ICCV}
Li, H., Zhao, J., Bazin, J.C., Chen, W., Liu, Z., Liu, Y.H.: Quasi-globally
  optimal and efficient vanishing point estimation in manhattan world. In: ICCV
  (2019)

\bibitem{li2023neuralangelo}
Li, Z., M\"uller, T., Evans, A., Taylor, R.H., Unberath, M., Liu, M.Y., Lin,
  C.H.: Neuralangelo: High-fidelity neural surface reconstruction. In: CVPR
  (2023)

\bibitem{lim2022uv}
Lim, H., Jeon, J., Myung, H.: Uv-slam: Unconstrained line-based slam using
  vanishing points for structural mapping. IEEE Robotics and Automation Letters
  (RA-L)  \textbf{7}(2),  1518--1525 (2022)

\bibitem{lim2021avoiding}
Lim, H., Kim, Y., Jung, K., Hu, S., Myung, H.: Avoiding degeneracy for
  monocular visual slam with point and line features. In: ICRA (2021)

\bibitem{lindenberger2021pixel}
Lindenberger, P., Sarlin, P.E., Larsson, V., Pollefeys, M.: Pixel-perfect
  structure-from-motion with featuremetric refinement. In: ICCV (2021)

\bibitem{Liu_2023_LIMAP}
Liu, S., Yu, Y., Pautrat, R., Pollefeys, M., Larsson, V.: 3d line mapping
  revisited. In: CVPR (2023)

\bibitem{lowe2004distinctive}
Lowe, D.G.: Distinctive image features from scale-invariant keypoints. IJCV
  \textbf{60}(2),  91--110 (2004)

\bibitem{lu2007fast}
Lu, F., Hartley, R.: A fast optimal algorithm for l 2 triangulation. In: ACCV
  (2007)

\bibitem{marzorati2007integration}
Marzorati, D., Matteucci, M., Migliore, D., Sorrenti, D.G.: Integration of 3d
  lines and points in 6dof visual slam by uncertain projective geometry. In:
  EMCR (2007)

\bibitem{mateus2021incremental}
Mateus, A., Tahri, O., Aguiar, A.P., Lima, P.U., Miraldo, P.: On incremental
  structure from motion using lines. IEEE Transactions on Robotics
  \textbf{38}(1),  391--406 (2021)

\bibitem{meidow2009reasoning}
Meidow, J., Beder, C., F{\"o}rstner, W.: Reasoning with uncertain points,
  straight lines, and straight line segments in 2d. ISPRS Journal of
  Photogrammetry and Remote Sensing  \textbf{64}(2),  125--139 (2009)

\bibitem{micusik2017structure}
Micusik, B., Wildenauer, H.: Structure from motion with line segments under
  relaxed endpoint constraints. IJCV  \textbf{124}(1),  65--79 (2017)

\bibitem{nerf}
Mildenhall, B., Srinivasan, P.P., Tancik, M., Barron, J.T., Ramamoorthi, R.,
  Ng, R.: Nerf: Representing scenes as neural radiance fields for view
  synthesis. In: ECCV (2020)

\bibitem{moulon2016openmvg}
Moulon, P., Monasse, P., Perrot, R., Marlet, R.: Open{MVG}: Open multiple view
  geometry. In: International Workshop on Reproducible Research in Pattern
  Recognition (2016)

\bibitem{muhle2023learning}
Muhle, D., Koestler, L., Jatavallabhula, K.M., Cremers, D.: Learning
  correspondence uncertainty via differentiable nonlinear least squares. In:
  CVPR (2023)

\bibitem{mur2015orb}
Mur-Artal, R., Montiel, J.M.M., Tardos, J.D.: Orb-slam: a versatile and
  accurate monocular slam system. IEEE Transactions on Robotics
  \textbf{31}(5),  1147--1163 (2015)

\bibitem{nister2005preemptive}
Nist{\'e}r, D.: Preemptive ransac for live structure and motion estimation.
  Machine Vision and Applications  \textbf{16}(5),  321--329 (2005)

\bibitem{nurutdinova2015towards}
Nurutdinova, I., Fitzgibbon, A.: Towards pointless structure from motion: 3d
  reconstruction and camera parameters from general 3d curves. In: ICCV (2015)

\bibitem{pautrat2021sold2}
Pautrat, R., Lin, J.T., Larsson, V., Oswald, M.R., Pollefeys, M.: Sold2:
  Self-supervised occlusion-aware line description and detection. In: CVPR
  (2021)

\bibitem{pautrat2023vanishing}
Pautrat, R., Liu, S., Hruby, P., Pollefeys, M., Barath, D.: Vanishing point
  estimation in uncalibrated images with prior gravity direction. In: ICCV
  (2023)

\bibitem{pautrat2023gluestick}
Pautrat, R., Su{\'a}rez, I., Yu, Y., Pollefeys, M., Larsson, V.: Gluestick:
  Robust image matching by sticking points and lines together. In: ICCV (2023)

\bibitem{Pautrat_2023_DeepLSD}
Pautrat, R., Barath, D., Larsson, V., Oswald, M.R., Pollefeys, M.: Deeplsd:
  Line segment detection and refinement with deep image gradients. In: CVPR
  (2023)

\bibitem{persson2018lambda}
Persson, M., Nordberg, K.: Lambda twist: An accurate fast robust perspective
  three point (p3p) solver. In: ECCV (2018)

\bibitem{poggi2016learning}
Poggi, M., Mattoccia, S.: Learning from scratch a confidence measure. In: BMVC
  (2016)

\bibitem{pumarola2017pl}
Pumarola, A., Vakhitov, A., Agudo, A., Sanfeliu, A., Moreno-Noguer, F.:
  Pl-slam: Real-time monocular visual slam with points and lines. In: ICRA
  (2017)

\bibitem{qian2004structure}
Qian, G., Chellappa, R.: Structure from motion using sequential monte carlo
  methods. IJCV  \textbf{59},  5--31 (2004)

\bibitem{Qian2022ARO}
Qian, Y., Elder, J.H.: A reliable online method for joint estimation of focal
  length and camera rotation. In: ECCV (2022)

\bibitem{roberts:2021}
Roberts, M., Ramapuram, J., Ranjan, A., Kumar, A., Bautista, M.A., Paczan, N.,
  Webb, R., Susskind, J.M.: {Hypersim}: {A} photorealistic synthetic dataset
  for holistic indoor scene understanding. In: ICCV (2021)

\bibitem{hloc}
Sarlin, P.E.: Visual localization made easy with hloc.
  \url{https://github.com/cvg/Hierarchical-Localization/}

\bibitem{sarlin2020superglue}
Sarlin, P.E., DeTone, D., Malisiewicz, T., Rabinovich, A.: Superglue: Learning
  feature matching with graph neural networks. In: CVPR (2020)

\bibitem{sarlin2022lamar}
Sarlin, P.E., Dusmanu, M., Schönberger, J.L., Speciale, P., Gruber, L.,
  Larsson, V., Miksik, O., Pollefeys, M.: {LaMAR: Benchmarking Localization and
  Mapping for Augmented Reality}. In: ECCV (2022)

\bibitem{schindler2006line}
Schindler, G., Krishnamurthy, P., Dellaert, F.: Line-based structure from
  motion for urban environments. In: International Symposium on 3D Data
  Processing, Visualization, and Transmission (3DPVT) (2006)

\bibitem{schonberger2016structure}
Schonberger, J.L., Frahm, J.M.: Structure-from-motion revisited. In: CVPR
  (2016)

\bibitem{schops2014semi}
Sch{\"o}ps, T., Engel, J., Cremers, D.: Semi-dense visual odometry for ar on a
  smartphone. In: International Symposium on Mixed and Augmented Reality
  (ISMAR) (2014)

\bibitem{schops2017multi}
Schops, T., Schonberger, J.L., Galliani, S., Sattler, T., Schindler, K.,
  Pollefeys, M., Geiger, A.: A multi-view stereo benchmark with high-resolution
  images and multi-camera videos. In: CVPR (2017)

\bibitem{seber2003nonlinear}
Seber, G.A., Wild, C.J.: Nonlinear regression. New Jersey: John Wiley \& Sons
  \textbf{62}(63), ~1238 (2003)

\bibitem{seki2016patch}
Seki, A., Pollefeys, M.: Patch based confidence prediction for dense disparity
  map. In: BMVC (2016)

\bibitem{7scenes}
Shotton, J., Glocker, B., Zach, C., Izadi, S., Criminisi, A., Fitzgibbon, A.:
  Scene coordinate regression forests for camera relocalization in {RGB-D}
  images. In: CVPR (2013)

\bibitem{shu2023structure}
Shu, F., Wang, J., Pagani, A., Stricker, D.: Structure plp-slam: Efficient
  sparse mapping and localization using point, line and plane for monocular,
  rgb-d and stereo cameras. In: ICRA (2023)

\bibitem{sinha2012multi}
Sinha, S.N., Steedly, D., Szeliski, R.: A multi-stage linear approach to
  structure from motion. In: Trends and Topics in Computer Vision: ECCV 2010
  Workshops, Heraklion, Crete, Greece, September 10-11, 2010, Revised Selected
  Papers, Part II 11 (2012)

\bibitem{snavely2006photo}
Snavely, N., Seitz, S.M., Szeliski, R.: Photo tourism: exploring photo
  collections in 3d. In: ACM SIGGRAPH (2006)

\bibitem{steedly2003spectral}
Steedly, D., Essa, I.A., Dellaert, F.: Spectral partitioning for structure from
  motion. In: ICCV (2003)

\bibitem{steele2005feature}
Steele, R.M., Jaynes, C.: Feature uncertainty arising from covariant image
  noise. In: CVPR (2005)

\bibitem{sturm2012benchmark}
Sturm, J., Engelhard, N., Endres, F., Burgard, W., Cremers, D.: A benchmark for
  the evaluation of rgb-d slam systems. In: IROS (2012)

\bibitem{theia-manual}
Sweeney, C.: Theia multiview geometry library: Tutorial \& reference.
  \url{http://theia-sfm.org}

\bibitem{sweeney2015optimizing}
Sweeney, C., Sattler, T., Hollerer, T., Turk, M., Pollefeys, M.: Optimizing the
  viewing graph for structure-from-motion. In: ICCV (2015)

\bibitem{nerfstudio}
Tancik, M., Weber, E., Ng, E., Li, R., Yi, B., Kerr, J., Wang, T.,
  Kristoffersen, A., Austin, J., Salahi, K., Ahuja, A., McAllister, D.,
  Kanazawa, A.: Nerfstudio: A modular framework for neural radiance field
  development. In: ACM SIGGRAPH 2023 Conference Proceedings (2023)

\bibitem{tang2018ba}
Tang, C., Tan, P.: Ba-net: Dense bundle adjustment network. In: International
  Conference on Learning Representations (ICLR) (2019)

\bibitem{taylor1995structure}
Taylor, C.J., Kriegman, D.J.: Structure and motion from line segments in
  multiple images. IEEE Trans. Pattern Analysis and Machine Intelligence (PAMI)
   \textbf{17}(11),  1021--1032 (1995)

\bibitem{toldo2008robust}
Toldo, R., Fusiello, A.: Robust multiple structures estimation with j-linkage.
  In: ECCV (2008)

\bibitem{triggs2000bundle}
Triggs, B., McLauchlan, P.F., Hartley, R.I., Fitzgibbon, A.W.: Bundle
  adjustment—a modern synthesis. In: Vision Algorithms: Theory and Practice:
  International Workshop on Vision Algorithms Corfu, Greece, September 21--22,
  1999 Proceedings (2000)

\bibitem{von2008lsd}
Von~Gioi, R.G., Jakubowicz, J., Morel, J.M., Randall, G.: Lsd: A fast line
  segment detector with a false detection control. IEEE Trans. Pattern Analysis
  and Machine Intelligence (PAMI)  \textbf{32}(4),  722--732 (2008)

\bibitem{wang2023visual}
Wang, J., Karaev, N., Rupprecht, C., Novotny, D.: Visual geometry grounded deep
  structure from motion. In: CVPR (2024)

\bibitem{wang2023posediffusion}
Wang, J., Rupprecht, C., Novotny, D.: Posediffusion: Solving pose estimation
  via diffusion-aided bundle adjustment. In: ICCV (2023)

\bibitem{wang2023dust3r}
Wang, S., Leroy, V., Cabon, Y., Chidlovskii, B., Revaud, J.: Dust3r: Geometric
  3d vision made easy. In: CVPR (2024)

\bibitem{wei2019real}
Wei, X., Huang, J., Ma, X.: Real-time monocular visual slam by combining points
  and lines. In: IEEE International Conference on Multimedia and Expo (ICME)
  (2019)

\bibitem{wilson2014robust}
Wilson, K., Snavely, N.: Robust global translations with 1dsfm. In: ECCV (2014)

\bibitem{wu2011visualsfm}
Wu, C.: Visualsfm: A visual structure from motion system. http://www. cs.
  washington. edu/homes/ccwu/vsfm  (2011)

\bibitem{wu2013towards}
Wu, C.: Towards linear-time incremental structure from motion. In:
  International Conference on 3D Vision (3DV) (2013)

\bibitem{xiao2023level}
Xiao, Y., Xue, N., Wu, T., Xia, G.S.: Level-s2fm: Structure from motion on
  neural level set of implicit surfaces. In: CVPR (2023)

\bibitem{xue2020holistically}
Xue, N., Wu, T., Bai, S., Wang, F., Xia, G.S., Zhang, L., Torr, P.H.:
  Holistically-attracted wireframe parsing. In: CVPR (2020)

\bibitem{yan2023plpf}
Yan, J., Zheng, Y., Yang, J., Mihaylova, L., Yuan, W., Gu, F.: Plpf-vslam: An
  indoor visual slam with adaptive fusion of point-line-plane features. Journal
  of Field Robotics  (2023)

\bibitem{yu2022monosdf}
Yu, Z., Peng, S., Niemeyer, M., Sattler, T., Geiger, A.: Monosdf: Exploring
  monocular geometric cues for neural implicit surface reconstruction. In:
  NeurIPS (2022)

\bibitem{zeisl2009estimation}
Zeisl, B., Georgel, P.F., Schweiger, F., Steinbach, E.G., Navab, N., Munich,
  G.: Estimation of location uncertainty for scale invariant features points.
  In: BMVC (2009)

\bibitem{zhang2018uncertainty}
Zhang, H., Grie{\ss}bach, D., Wohlfeil, J., B{\"o}rner, A.: Uncertainty model
  for template feature matching. In: Image and Video Technology: 8th
  Pacific-Rim Symposium, PSIVT 2017, Wuhan, China, November 20-24, 2017,
  Revised Selected Papers 8 (2018)

\bibitem{zhang2024cameras}
Zhang, J.Y., Lin, A., Kumar, M., Yang, T.H., Ramanan, D., Tulsiani, S.: Cameras
  as rays: Pose estimation via ray diffusion. In: International Conference on
  Learning Representations (ICLR) (2024)

\bibitem{zhang2015}
Zhang, L., Lu, H., Hu, X., Koch, R.: Vanishing point estimation and line
  classification in a manhattan world with a unifying camera model. IJCV
  \textbf{117} (2015)

\bibitem{zhao2021confidence}
Zhao, W., Liu, S., Wei, Y., Guo, H., Liu, Y.J.: A confidence-based iterative
  solver of depths and surface normals for deep multi-view stereo. In: ICCV
  (2021)

\bibitem{zhou2018stable}
Zhou, L., Ye, J., Kaess, M.: A stable algebraic camera pose estimation for
  minimal configurations of 2d/3d point and line correspondences. In: ACCV
  (2018)

\bibitem{zuo2017robust}
Zuo, X., Xie, X., Liu, Y., Huang, G.: Robust visual slam with point and line
  features. In: IROS (2017)

\end{thebibliography}
\end{document}